\documentclass[journal, twoside]{IEEEtran}
\usepackage{amsmath,amsfonts}
\usepackage{algorithmic}
\usepackage{algorithm}
\usepackage{array}
\usepackage{textcomp}
\usepackage{stfloats}
\usepackage{url}
\usepackage{verbatim}
\usepackage{graphicx}
\usepackage{afterpage}
\usepackage{amssymb}
\usepackage{bbm}
\usepackage{bm}
\usepackage{enumerate}

\usepackage{multirow}
\usepackage{placeins}
\usepackage{tabularx}
\usepackage{threeparttable}

\usepackage[caption=false, font=footnotesize]{subfig}

\usepackage{booktabs}
\usepackage{xcolor}

\usepackage[backend=biber, style=ieee, citestyle=ieee-comp, url=true, maxbibnames=99, sorting=none]{biblatex}

\nocite{*}

\usepackage[colorlinks]{hyperref}
\usepackage{orcidlink}

\begin{document}

\title{Dual-Thresholded Heatmap-Guided Proposal Clustering and Negative Certainty Supervision with Enhanced Base Network for Weakly Supervised Object Detection}

\author{
Yuelin Guo$^{\orcidlink{0009-0009-3448-529X}}$, Haoyu He$^{\orcidlink{0000-0002-1505-0284}}$, Zhiyuan Chen$^{\orcidlink{0009-0008-0390-0262}}$, Zitong Huang$^{\orcidlink{0009-0001-1931-8217}}$,\\
Renhao Lu$^{\orcidlink{0000-0002-4467-1215}}$, Lu Shi$^{\orcidlink{0009-0009-9945-9914}}$, Zejun Wang$^{\orcidlink{0009-0009-0034-5459}}$, Weizhe Zhang$^{\orcidlink{0000-0003-4783-876X}}$,~\IEEEmembership{Senior Member,~IEEE}

\thanks{Yuelin Guo, and Zhiyuan Chen are with the Institute of Cyberspace Security, Harbin Institute of Technology, Shenzhen, Shenzhen 518055, China (e-mail: gyl2565309278@gmail.com; xeesoxeechen@gmail.com).}
\thanks{Haoyu He is with the Faculty of Information Technology, Monash University, Victoria 3800, Australia (e-mail: Charles.haoyu.he@gmail.com).}
\thanks{Zitong Huang is with the Center on Machine Learning Research, Harbin Institute of Technology, Harbin 150001, China (e-mail: zitonghuang@outlook.com).}
\thanks{Renhao Lu is with the Department of New Networks, Peng Cheng Laboratory, Shenzhen 518066, China (e-mail: lurh100@pcl.ac.cn).}
\thanks{Lu Shi is with the Institute of Cyberspace Security, Harbin Institute of Technology, Shenzhen, Shenzhen 518055, China, and the Department of New Networks, Peng Cheng Laboratory, Shenzhen 518066, China (e-mail: mathis.lu.stone@gmail.com).}
\thanks{Zejun Wang is with the School of Cyberspace Science, Harbin Institute of Technology, Harbin 150001, China (e-mail: zejunwang@stu.hit.edu.cn).}
\thanks{Weizhe Zhang is with the Institute of Cyberspace Security, Harbin Institute of Technology, Shenzhen, Shenzhen 518055, China, the Department of New Networks, Peng Cheng Laboratory, Shenzhen 518066, China, and the School of Cyberspace Science, Harbin Institute of Technology, Harbin 150001, China (e-mail: wzzhang@hit.edu.cn).}
}



\maketitle

\begin{abstract}
Weakly supervised object detection (WSOD) has attracted significant attention in recent years, as it does not require box-level annotations. State-of-the-art methods generally adopt a multi-module network, which employs WSDDN as the multiple instance detection network module and uses multiple instance refinement modules to refine performance. However, these approaches suffer from three key limitations. First, existing methods tend to generate pseudo GT boxes that either focus only on discriminative parts, failing to capture the whole object, or cover the entire object but fail to distinguish between adjacent intra-class instances. Second, the foundational WSDDN architecture lacks a crucial background class representation for each proposal and exhibits a large semantic gap between its branches. Third, prior methods discard ignored proposals during optimization, leading to slow convergence. To address these challenges, we propose the Dual-thresholded heAtmap-guided proposal clustering and Negative Certainty supervision with Enhanced base network (DANCE) method for WSOD. Specifically, we first devise a heatmap-guided proposal selector (HGPS) algorithm, which utilizes dual thresholds on heatmaps to pre-select proposals, enabling pseudo GT boxes to both capture the full object extent and distinguish between adjacent intra-class instances. We then construct a weakly supervised basic detection network (WSBDN), which augments each proposal with a background class representation and uses heatmaps for pre-supervision to bridge the semantic gap between matrices. At last, we introduce a negative certainty supervision (NCS) loss on ignored proposals to accelerate convergence. Extensive experiments on the challenging PASCAL VOC and MS COCO datasets demonstrate the effectiveness and superiority of our method. Our code is publicly available at \url{https://github.com/gyl2565309278/DANCE}.
\end{abstract}

\begin{IEEEkeywords}
Weakly supervised object detection, heatmap-guided proposal selector, weakly supervised basic detection network, negative certainty supervision.
\end{IEEEkeywords}

\section{Introduction} \label{sec:1}

\IEEEPARstart{O}{bject} detection is an important and fundamental problem in computer vision, which aims at identifying the categories and determining the locations of objects within a given image. Driven by the advancements in Convolutional Neural Networks (CNNs) \cite{Krizhevsky2012AlexNet, Simonyan2015VGG, He2016ResNet, Lin2017FPN} and Transformers \cite{Vaswani2017Transformer, Dosovitskiy2020ViT, Liu2021Swin, Fan2021MViT}, as well as the emergence of large-scale datasets with bounding box annotations \cite{Everingham2010VOC, Lin2014COCO, Russakovsky2015ImageNet}, the field of fully supervised object detection (FSOD) \cite{Girshick2014RCNN, Girshick2015FRCNN, Ren2015FRRCNN, Liu2016SSD, Redmon2016YOLO, Redmon2017YOLOv2, Carion2020DETR, Zhu2021Deformable-DETR, Meng2021Conditional-DETR, Liu2022DAB-DETR, Li2022DN-DETR} has witnessed rapid progress in recent years. However, annotating precise bounding-box-level labels is a labor-intensive and time-consuming process. This difficulty has given rise to the weakly supervised object detection (WSOD) task, which aims at predicting precise bounding boxes under only image-level supervision.

\begin{figure}[!t]
\centering
\vspace{-0.8em}
\subfloat[]{
\includegraphics[width=0.3\columnwidth]{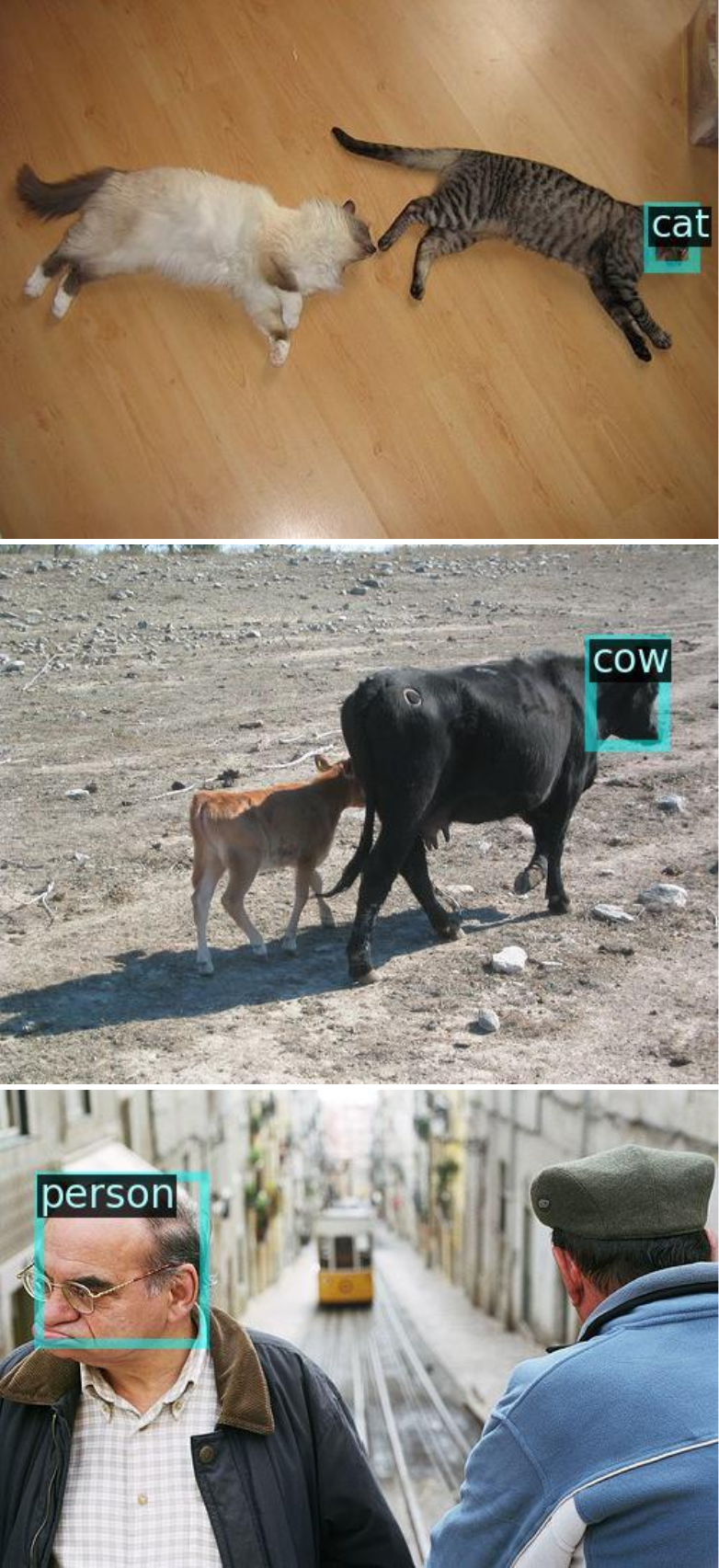}
\label{fig:challenges-a}
}
\hspace{-9pt}
\subfloat[]{
\includegraphics[width=0.3\columnwidth]{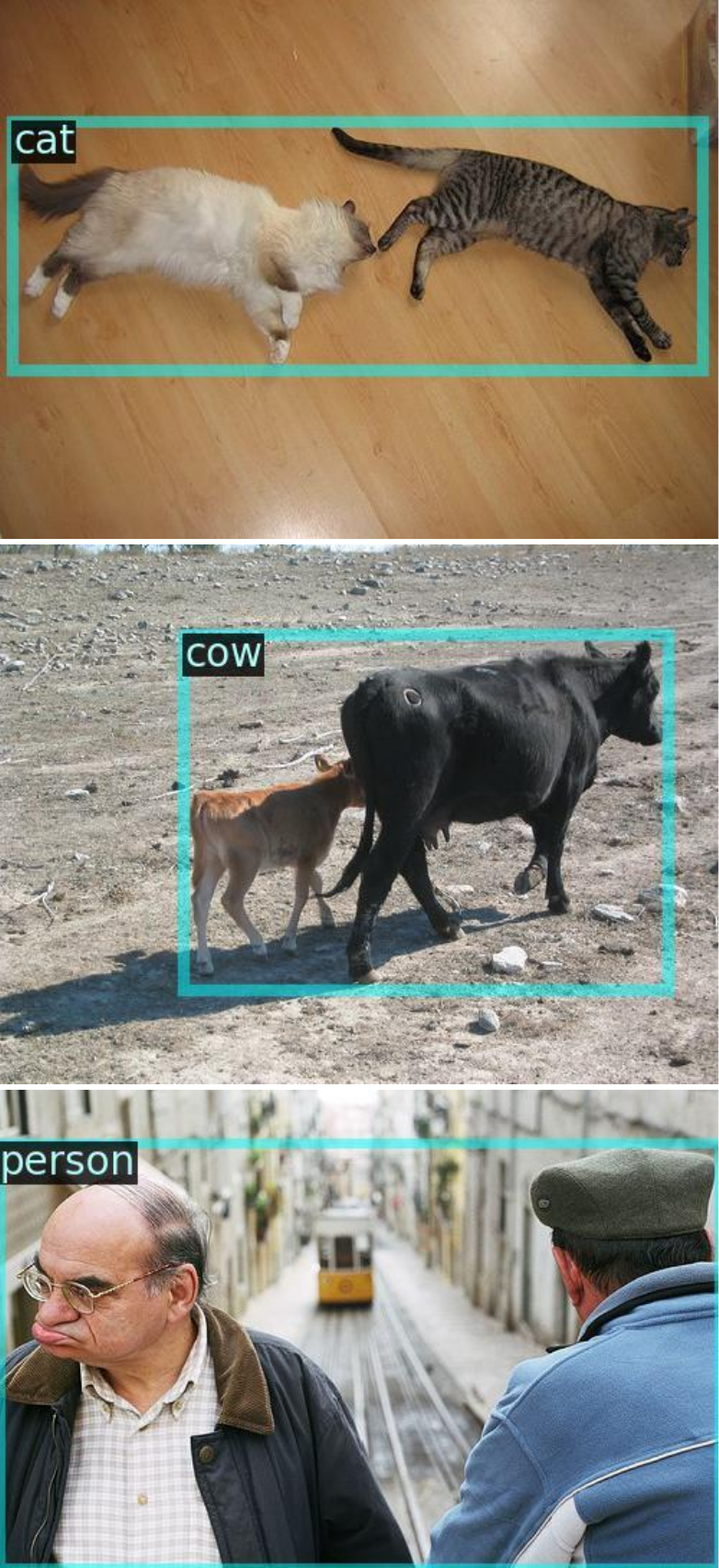}
\label{fig:challenges-b}
}
\hspace{-9pt}
\subfloat[]{
\includegraphics[width=0.3\columnwidth]{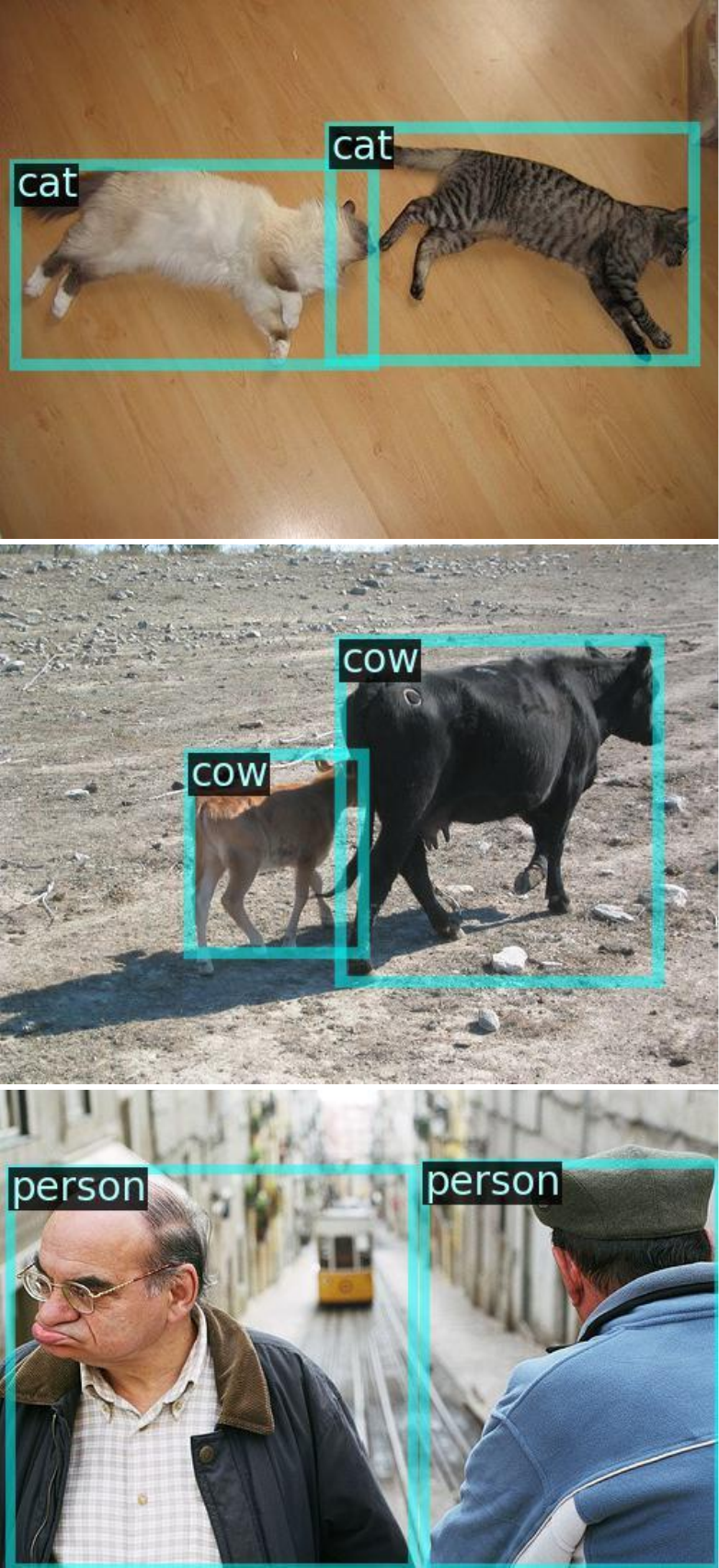}
\label{fig:challenges-c}
}
\vspace{-0.3em}
\caption{Common challenges: (a) High-scoring-proposal methods \cite{Tang2017OICR, Tang2018PCL} capture only discriminative parts and miss some of the instances. (b) Thresholding-heatmap methods \cite{Zhang2018ZLDN, Chen2020SLV} merging adjacent intra-class instances. Ideal condition: (c) Our DANCE, which employs dual thresholds on heatmaps to select proposals, generates one tight box per instance.}
\label{fig:challenges}
\vspace{-1.5em}
\end{figure}

Due to the lack of ground-truth bounding boxes, the mainstream paradigm for WSOD \cite{Bilen2016WSDDN, Tang2017OICR, Tang2018PCL, Ren2020MIST, Huang2020CASD, Yin2021IM-CFB, Yin2023CBL} typically employs a two-stage learning process. In the first stage, region proposals are pre-extracted using methods such as Selective Search (SS) \cite{Uijlings2013SS}, Edge Boxes (EB) \cite{Zitnick2014EB}, Multiscale Combinatorial Grouping (MCG) \cite{Arbelaez2014MCG}, or the Segment Anything Model (SAM) \cite{Kirillov2023SAM} to exploit rich contextual prior knowledge from each image. In the second stage, a backbone network is utilized to extract features for each proposal. These features are then used to classify the proposals under the Multiple Instance Learning (MIL) constraint: a positive bag (the image) must contain at least one positive instance (a proposal), whereas a negative bag consists solely of negative instances (no proposal). Furthermore, multiple Instance Refinement (IR) modules are often trained in cascade. In this process, pseudo ground-truth (GT) boxes selected from the current module are used to supervise the next module.

It is evident that the quality of pseudo GT boxes fundamentally determines the performance of WSOD, whose upper bound is FSOD, as the pseudo boxes are perfectly equal to the ground-truth annotations. Therefore, a vast body of research is dedicated to devising various strategies for obtaining more accurate pseudo GT boxes. Among these, the most common and widespread approaches \cite{Tang2017OICR, Tang2018PCL, Wu2022BUAA-PAL, Yin2023CBL} leverage the classification score matrix, selecting high-scoring proposals as pseudo GT boxes. However, these methods suffer from the problem that high-scoring proposals always tend to concentrate on the discriminative part of an object, as illustrated in Figure \ref{fig:challenges}\subref{fig:challenges-a}. Therefore, another line of approaches \cite{Diba2017WCCN, Zhang2018ZLDN, Chen2020SLV, Liao2022SPE} circumvent the issue by setting a low threshold on heatmaps to ensure the generated pseudo GT boxes encompass the full extent of an object. Nevertheless, these thresholding-based methods inevitably fail to distinguish among adjacent intra-class instances, as shown in Figure \ref{fig:challenges}\subref{fig:challenges-b}.

These phenomena lead us to the following summary: Relying solely on the classification score matrix leads to the discriminative part of objects. Meanwhile, relying solely on thresholding heatmaps results in an inability to distinguish among adjacent intra-class objects. Either of these shortcomings ultimately degrades the final performance of WSOD.

In light of these deficiencies in existing methods, a natural yet compelling motivation presents itself: Why not integrate both strengths while mitigating their weaknesses to obtain more accurate pseudo GT boxes? Leveraging heatmaps to perceive objects' approximate location and outline to prevent pseudo GT boxes from collapsing into just the discriminative parts. Introducing proposals and their classification scores to ensure the pseudo GT boxes are not solely reliant on connected threshold regions of the heatmaps, thereby enabling differentiation of adjacent or overlapping intra-class objects. Therefore, we propose a \textbf{Heatmap-Guided Proposal Selector (HGPS)} algorithm, applying dual thresholds on heatmaps for proposal selection and obtaining pseudo GT boxes as presented in Figure \ref{fig:challenges}\subref{fig:challenges-c}.

\begin{table}[!t]
\caption{The performance gap between the two matrices in WSDDN}
\label{tab:WSDDN}
\centering
\newcolumntype{L}{>{\raggedright\arraybackslash}p{40pt}}
\newcolumntype{M}{>{\centering\arraybackslash}p{31pt}}
\begin{tabular}{L | M M | M M}
\toprule[0.15em]
\multirow{2}{*}{Method} & \multicolumn{2}{c |}{$\bm{s}^{(0)}$} & \multicolumn{2}{c}{$\bm{ws}^{(0)}$} \\
\cmidrule[0.05em]{2-5}
                        & \multicolumn{1}{M |}{mAP} & \multicolumn{1}{M |}{mCorLoc} & \multicolumn{1}{M |}{mAP} & \multicolumn{1}{M}{mCorLoc} \\
\midrule[0.05em]
WSDDN & \multicolumn{1}{M |}{5.0} & \multicolumn{1}{M |}{24.2} & \multicolumn{1}{M |}{34.0} & \multicolumn{1}{M}{57.5} \\
\bottomrule[0.15em]
\end{tabular}
\end{table}

Additionally, we observe that nearly all existing methods rely on WSDDN \cite{Bilen2016WSDDN}, a foundational WSOD model, as their Multiple Instance Detection Network (MIDN) module. However, its design and application differ significantly from general object detectors. First, for a dataset with $C$ categories, a general detector assigns $C+1$ scores to each proposal, where the final score represents the confidence that the proposal belongs to the background category. However, the score vector computed by WSDDN has only a shape of $C$, lacking a semantic representation for the background. Second, the classification scores of a general detector are obtained directly through class-wise softmax. In contrast, WSDDN's detection scores are derived from the Hadamard product of both class-wise softmax and proposal-wise softmax branches. For this reason, we separately evaluated the performance of the class-wise softmax matrix, which is semantically aligned with general detectors, and the final Hadamard product matrix. The results are presented in Table \ref{tab:WSDDN}, where $\bm{s}^{(0)}$ and $\bm{ws}^{(0)}$ denote the class-wise softmax matrix and the final Hadamard product matrix respectively. Although $\bm{ws}^{(0)}$ achieves 34.0\% mAP and 57.5\% mCorLoc, $\bm{s}^{(0)}$ yields only 5.0\% mAP and 24.2\% mCorLoc, revealing significant gap. To put it another way, the two matrices are semantically misaligned to a critical degree. The performance of $\bm{s}^{(0)}$ is so poor that it can be seen as catastrophically degraded, acting as an internal bottleneck fundamentally limiting the performance of the final $\bm{ws}^{(0)}$. Therefore, we propose a \textbf{Weakly Supervised Basic Detection Network (WSBDN)} --- a new foundational WSOD model. It reverts the feature representation for each proposal from $C$ back to $C+1$, and we thus define a novel ``box-level image label'' correspondingly. Furthermore, we introduce an additional supervision signal to the class-wise softmax branch, which narrows the gap between the two matrices and, in turn, boosts the overall performance.

Finally, we notice that many WSOD approaches \cite{Tang2017OICR, Tang2018PCL, Ren2020MIST, Huang2020CASD} tend to discard proposals with minimal IoU overlap with pseudo GT boxes during training, not applying any loss on them, which results in a slow convergence rate. However, although the IoU thresholding prevents us from determining the exact categories of these proposals, we can still be certain of the categories they do not belong to --- namely, the categories absent from the image. Therefore, we propose a \textbf{Negative Certainty Supervision (NCS)} loss on these ignored proposals. By constraining them with absolutely reliable negative information, we reduce the optimization difficulty and thus accelerate network convergence.

Accordingly, the contributions of our \textbf{Dual-thresholded heAtmap-guided proposal clustering and Negative Certainty supervision with Enhanced base network (DANCE)} framework can be summarized as follows:
\begin{itemize}
    \item [1)] We devise HGPS, an algorithm that generates pseudo GT boxes by using heatmaps to provide rough locations and leveraging proposals to bound each instance precisely. It ensures the pseudo GT boxes not only extend beyond the discriminative part of each object, but also distinguish between adjacent intra-class objects.
    \item [2)] We construct WSBDN, a superior foundational MIDN module as a replacement of WSDDN. It aligns the semantic representation in WSOD with that in FSOD, and narrows the performance gap between the two inner matrices, which can significantly boost the baseline's performance.
    \item [3)] We introduce NCS, a loss that leverages negative certainty information as a supervision signal on ignored proposals. It enables these proposals to converge in the definitely correct direction, thereby facilitating faster convergence across the entire network.
\end{itemize}

We conducted extensive experiments on the challenging PASCAL VOC \cite{Everingham2010VOC} and COCO \cite{Lin2014COCO} datasets, and the experimental results show that our proposed DANCE method performs favorably against the SOTA methods.

\section{Related Work}

The WSOD task has been applied to numerous specific scenarios and has spawned various sub-tasks since its inception, such as remote sensing image detection \cite{Qian2025RSI1, Qian2025RSI2, Qian2025RSI3}, cross-supervised object detection \cite{Xu2023CST}, and cross-domain weakly supervised object detection \cite{Xu2022H2FA-R-CNN, Tang2023DETR-GA}. Therefore, the accuracy of algorithms within the WSOD domain itself is of crucial importance.

\subsection{Pseudo Ground-Truth Boxes Generation}

Since WSDDN \cite{Bilen2016WSDDN} and OICR \cite{Tang2017OICR} were proposed, nearly all mainstream WSOD works have employed WSDDN as the foundational MIDN module and leverage a multi-module refinement strategy as pioneered by OICR. In this paradigm, the quality of pseudo GT boxes is the most critical determinant of a model's final performance, attracting nearly all researchers' attention in this field.

Among them, the most widespread approach is to utilize the classification score matrix. The underlying assumption is that proposal scores obtained during training possess a certain degree of confidence, which can thus serve as prior knowledge to guide the selection of pseudo GT boxes. The most representative methods in this category include OICR \cite{Tang2017OICR}, PCL \cite{Tang2018PCL}, W2F \cite{Zhang2018W2F}, MIST \cite{Ren2020MIST}, CBL \cite{Yin2023CBL}, MCC-MCT \cite{Wu2024MCC-MCT}, and ICBC \cite{Yin2025ICBC}. These methods employ various techniques --- such as graph construction, merging, non-maximum suppression (NMS) \cite{Neubeck2006NMS} filtering, masking, differencing, cyclical supervision, and so on --- to select high-scoring proposals as pseudo GT boxes. However, these methods are all plagued by two common problems: 1) If an image has multiple objects of the same class, the pseudo GT boxes may miss some instances, resulting in incomplete supervision. 2) When the initial high-scoring proposals are themselves inaccurate (e.g., focusing only on the discriminative part of an object), these methods lack an additional mechanism to correct them, causing the model to converge in the wrong direction.

Note that it will be highly beneficial to performance if finer-grained information can be obtained in weakly supervised tasks to, in turn, guide the higher-level frame. In the field of object detection, a box-level task, the underlying information is at the pixel level. This naturally leads to the insight of using category-specific heatmaps to provide additional support for WSOD, aiding in generating more accurate pseudo GT boxes. The most straightforward approaches, such as WCCN \cite{Diba2017WCCN}, ZLDN \cite{Zhang2018ZLDN}, SLV \cite{Chen2020SLV}, SPE \cite{Liao2022SPE}, et al, are to directly apply a threshold to the heatmap and use the resulting tightest bounding box of each connected region as pseudo GTs. However, this series of methods has an unavoidable drawback: if the threshold is set too high, the resulting bounding boxes will fail to capture the entire object; conversely, if it is set too low, the boxes will merge adjacent intra-class instances.

Recognizing that directly using boxes from either a high or a low threshold has inherent flaws and thus limits the quality of pseudo GT boxes, we instead treat them solely as a coarse positional prior to guide the generation of higher-quality pseudo GT boxes. Considering that the large pool of region proposals contains high-quality candidates, we still select a part of the proposals as pseudo GT boxes, thereby raising the quality's upper bound. Moreover, the localization knowledge provided by the threshold boxes prevents the selected proposals from concentrating solely on the discriminative part of objects. As a result, all the aforementioned issues are elegantly resolved. We name our pseudo GT boxes generation algorithm as \textbf{Heatmap-Guided Proposal Selector (HGPS)}.

Additionally, another group of heatmap-based methods indirectly guides the scores or loss of a proposal, with representative works including TS\textsuperscript{2}C \cite{Wei2018TS2C}, OAIL \cite{Kosugi2019OAIL}, WSOD\textsuperscript{2} \cite{Zeng2019WSOD2}, SDCN \cite{Li2019SDCN}, et al. These methods evaluate the difference between the proportion of high-activation points within a proposal and that in its surrounding context region, which is then used to suppress the scores or down-weight the loss of proposals focusing only on the discriminative part. This, in turn, allows proposals containing complete object information to achieve higher scores. However, these methods share two fatal limitations: 1) They lack precise control over the scores, meaning the selected proposals may still contain incomplete objects. 2) The selection strategy still relies on an OICR-style approach, i.e., selecting only the single top-scoring proposal as the pseudo GT box. This results in an incomplete set of pseudo instances when multiple objects of the same category are present in an image.

Some other methods, such as TPEE \cite{Yang2019TPEE} and CASD \cite{Huang2020CASD}, utilizes heatmaps to enhance the feature maps. However, these approaches offer limited gains for WSOD, as the feature representation from modern backbones are already quite powerful. BUAA-PAL \cite{Wu2022BUAA-PAL} even points out that the performance boost of CASD is not primarily derived from its feature alignment design. Instead, the improvement stems from its strategy of using multiple data-augmented versions of the same image in each training iteration. It processes a data volume several times larger than that of other methods for the same number of epochs, which leads to its most main performance gains.

Recently, another type of method leverages the proposal features to refine pseudo GT boxes. IM-CFB \cite{Yin2021IM-CFB}, NDI-WSOD \cite{Wang2022NDI-WSOD}, OD-WSCL \cite{Seo2022OD-WSCL}, NPGC \cite{Zhang2023NPGC} et al optimize the selection results by using feature similarity to construct class-specific feature pools. However, this class of methods generally suffers from two problems: 1) The positive feature pool is often constructed based on the features of the top-scoring proposal and its variations. When this top-scoring proposal is itself inaccurate, the feature pool inherits the same bias. 2) The negative feature pool is built by ``Negative Deterministic Information''. However, no additional constraints are imposed to rectify proposals with significant classification errors. This leads to a sub-optimal classification capability and slow convergence of the entire network.

Therefore, we propose \textbf{Negative Certainty Supervision (NCS)}, utilizing negative certainty labels to supervise ignored proposals on their negative certainty categories, which is denoted as the ``classification-ignored loss'' and integrated into the overall loss function. By guiding these proposals to converge in the absolutely correct direction, we ensure that every proposal is supervised, thereby alleviating the optimization difficulty and accelerating the convergence rate of the entire network.

\subsection{Multiple Instance Detection Networks}

Since WSDDN was proposed, it has been adopted as the cornerstone MIDN module in nearly all subsequent methods. Naturally, a significant body of work has also been dedicated to proposing a superior MIDN structure to boost overall model performance by enhancing this foundational component. ContextLocNet \cite{Kantorov2016ContextLocNet} and PSLR \cite{Zhang2020PSLR} alleviate the discriminative part issue by incorporating contextual features while suppressing internal ones. WS-JDS \cite{Shen2019WS-JDS} and CSC \cite{Shen2019CSC} apply heatmaps to the proposal-wise and class-wise softmax branches, respectively, to provide WSDDN with pixel-level information. SCS \cite{Liu2019SCS} and D-MIL \cite{Gao2022D-MIL} employ parallel WSDDN streams, while C-MIDN \cite{Gao2019C-MIDN} and P-MIDN+MGSC \cite{Xu2021P-MIDN+MGSC} opt for a cascaded design, both aiming to enrich the pseudo GT boxes generated by the MIDN module. CPNet \cite{Li2022CPNet} replaces the fully-connected layers in WSDDN's two streams with Cross-Attention \cite{Vaswani2017Transformer} structures. However, the works mentioned above still have a shared limitation: they fail to break free from the confines of the WSDDN framework. Consequently, they all inherit the inherent weaknesses of WSDDN, which indirectly restricts their potential performance. 

Admittedly, works such as CST \cite{Xu2023CST} have re-introduced the background class representation to WSDDN, while H\textsuperscript{2}FA-R-CNN \cite{Xu2022H2FA-R-CNN} and DETR-GA \cite{Tang2023DETR-GA} have departed from the WSOD paradigm to directly adopt the FSOD paradigm. However, these methods belong to the fields of cross-supervised object detection or cross-domain weakly supervised object detection and, strictly speaking, do not fall within the scope of the standard WSOD domain. Since their training data partially contains box-level annotations, the concept of a background class is inherent in their problem definition. Therefore, it is natural for these works to add back the background representation or diverge from the WSOD paradigm. In contrast, we have improved the WSDDN paradigm under the standard WSOD setting. Our \textbf{Weakly Supervised Basic Detection Network (WSBDN)} first augments each proposal with a background class representation and then leverages heatmaps to bridge the semantic gap between the matrices. This enhancement boosts the base module itself, which, in turn, drives the performance improvement of the overall model.

\section{Method}

\subsection{Preliminary Knowledge}  \label{sec:3-1}

Given an image $\bm{I} \in \mathbb{R}^{H \times W \times 3}$, many existing works \cite{Bilen2016WSDDN, Tang2017OICR, Tang2018PCL, Ren2020MIST, Huang2020CASD} denote its image-level label as $\bm{y} = {\left[y_1, y_2, \cdots, y_C\right]}^{T} \in \mathbb{R}^{C}$, where $H$ and $W$ mean hight and width of the image, $C$ denotes the total number of object categories in the dataset, and $y_c = 1$ or $0$ indicates the presence or absence of at least one object of category $c$ in the image. The corresponding region proposals pre-generated for image $\bm{I}$ is defined as $\bm{\mathcal{P}} = \left\{\bm{P}_1, \bm{P}_2, \cdots, \bm{P}_R\right\}$, where $R$ denotes the number of proposals.

Due to the lack of instance-level annotations in WSOD, these works combine MIL with a CNN model as MIDN to accomplish the detection task. They utilize a two-stream weakly supervised deep detection network (WSDDN) \cite{Bilen2016WSDDN} as their MIDN module. The feature $\bm{f}_r \in \mathbb{R}^{D}$ of each proposal $\bm{P}_r$ is first extracted through a CNN backbone, followed by an RoI Pooling layer \cite{Girshick2015FRCNN} and two FC layers. Then these proposal features are fed into the dual branches in WSDDN (i.e., classification and detection branch). Each branch employs a FC layer to map the feature $\bm{f}_r$ from $\mathbb{R}^{D}$ to $\mathbb{R}^{C}$, and thus the matrices in each of the two parallel branches for all proposals are denoted as $\bm{\varphi}^{\mathrm{cls}}, \bm{\varphi}^{\mathrm{det}} \in \mathbb{R}^{R \times C}$. The first matrix $\bm{\varphi}^{\mathrm{cls}}$ then undergoes a softmax operation along the class dimension to produce: ${\left[\mathrm{Softmax}_{\mathrm{cls}} \left(\bm{\varphi}^{\mathrm{cls}}\right)\right]}_{r, c} = \frac{\exp \left(\varphi_{r, c}^{\mathrm{cls}}\right)}{\sum_{c'=1}^{C} \exp \left(\varphi_{r, c'}^{\mathrm{cls}}\right)}$, while the second matrix is softmax-operated along its proposal dimension: ${\left[\mathrm{Softmax}_{\mathrm{pro}} \left(\bm{\varphi}^{\mathrm{det}}\right)\right]}_{r, c} = \frac{\exp \left(\varphi_{r, c}^{\mathrm{det}}\right)}{\sum_{r'=1}^{R} \exp \left(\varphi_{r', c}^{\mathrm{det}}\right)}$. After that, the proposal score matrix is generated by Hadamard product $\bm{\varphi}^{\left(0\right)} = \mathrm{Softmax}_\mathrm{cls} \left(\bm{\varphi}^{\mathrm{cls}}\right) \odot \mathrm{Softmax}_\mathrm{pro} \left(\bm{\varphi}^{\mathrm{det}}\right)$. Finally, the image-level class prediction scores are generated through the summation over the proposal's dimension: $\varphi_c^{\mathrm{img}} = \sum_{r=1}^{R} \varphi_{r, c}^{\left(0\right)}$. In this way, the MIDN module is trained by the binary cross-entropy loss function:
\begin{equation}
\resizebox{0.9\columnwidth}{!}{$\displaystyle
    \mathcal{L}_{\mathrm{WSDDN}} = -\sum_{c=1}^{C} \left[y_c \log \varphi_c^{\mathrm{img}} + \left(1 - y_c\right) \log \left(1 - \varphi_c^{\mathrm{img}}\right)\right].
$}
\end{equation}

Subsequently, OICR \cite{Tang2017OICR} adds multiple cascaded IR modules after WSDDN and utilizes the pseudo GT boxes selected from each module to supervise the next. This paradigm became the dominant paradigm for WSOD. Specifically, each IR module contains a classification branch, which is parallel to the classification and detection branches as mentioned above in WSDDN. However, different from the two branches, the FC layer in the IR module maps each proposal's feature $\bm{f}_r$ from $\mathbb{R}^{D}$ to $\mathbb{R}^{C+1}$, followed by a class-wise softmax operation to yield a predicted score vector, where $\left(C+1\right)$ denotes $C$ different foreground classes and a background class. The score vectors of all proposals constitute a score matrix, which is recorded as $\bm{\varphi}^{(k)} \in \mathbb{R}^{R \times \left(C+1\right)}$ in the $k$-th IR module. During training, the top-scoring proposal $r_c^{(k-1)}=\arg\max_{r \in \left\{1, 2, \cdots, R\right\}} \varphi_{r, c}^{(k-1)}$ is selected as the pseudo GT box for each class $c$ with label $y_c=1$. Then the box-level label $y_{r}^{(k)}$ is assigned to each proposal $\bm{P}_r$ via Intersection-over-Union (IoU) threshold, where $y_r^{(k)} = c$ indicates that the proposal $\bm{P}_r$ belongs to class $c$. In this way, the $k$-th IR module is trained by the cross-entropy loss function:
\begin{equation}
    \mathcal{L}_{\mathrm{OICR}}^{(k)} = -\frac{1}{R} \sum_{r=1}^{R} w_r^{(k)} \sum_{c=1}^{C+1} \mathbbm{1}\left[y_r^{(k)} = c\right] \log \varphi_{r, c}^{(k)},
\end{equation}
where $w_r^{(k)}$ is the confidence score from the $(k-1)$-th module on the class of the pseudo GT box that best matches $\bm{P}_r$.

Finally, the weakly supervised detector is trained end-to-end by combining the loss functions of WSDDN and OICR:
\begin{equation}
    \mathcal{L}_{\mathrm{total}} = \mathcal{L}_{\mathrm{WSDDN}} + \sum_{k=1}^{K} \mathcal{L}_{\mathrm{OICR}}^{\left(k\right)},
\end{equation}
where $K$ represents the total number of IR modules.

\subsection{Overview of Proposed Model}

\begin{figure*}[!t]
\centering
\includegraphics[width=\textwidth]{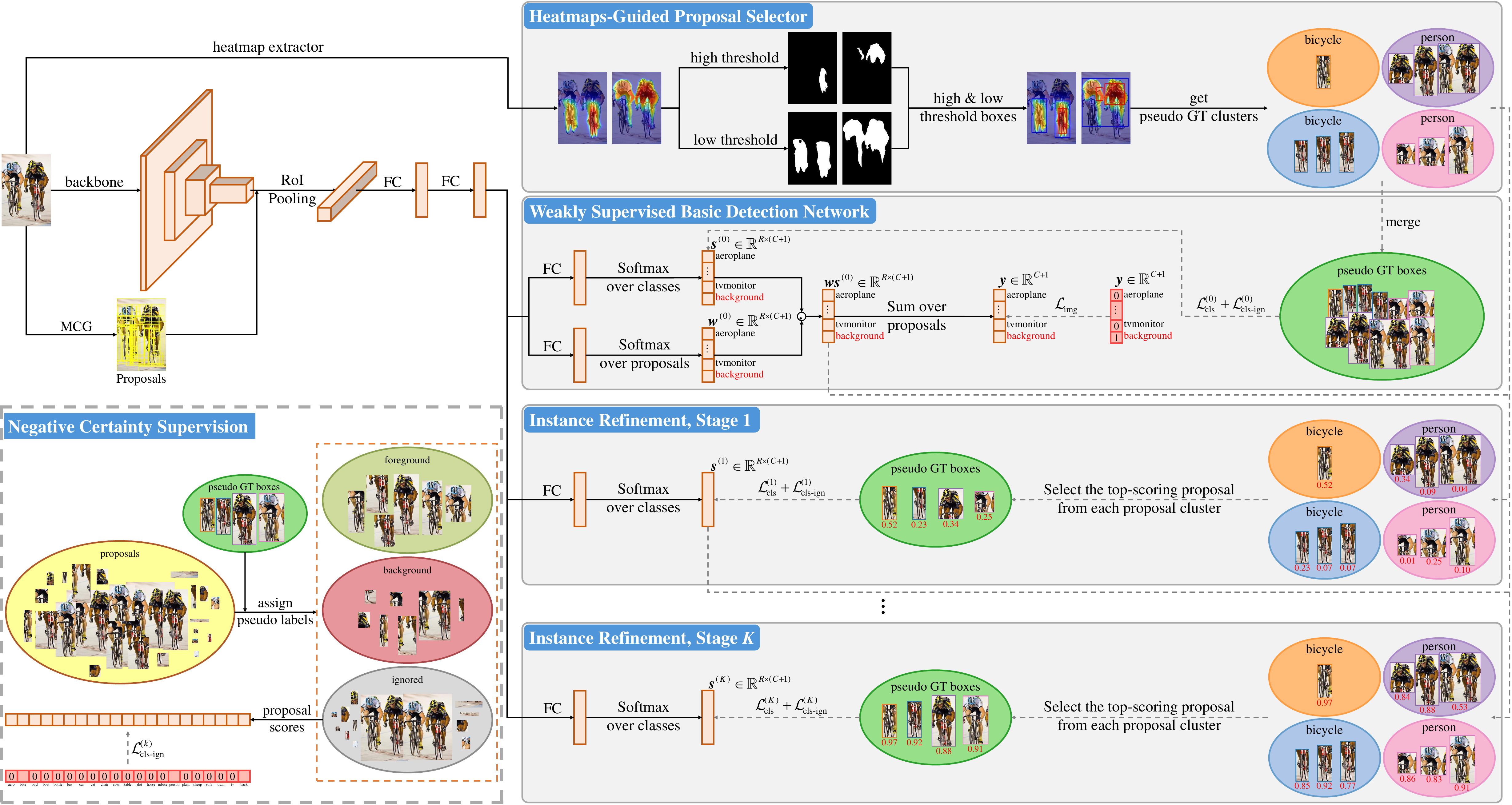}
\caption{The overview architecture of \textbf{DANCE}. \textbf{HGPS}: Given an image, category-specific heatmaps are first obtained through the heatmap extractor. Dual thresholds are then applied to generate tight bounding boxes, where proposals falling between the high and scaled low boxes are assigned to corresponding clusters as a pseudo-GT-box candidate set. During training, the top-scoring proposal within each cluster is selected to form pseudo GT boxes for adjacent modules' supervision. \textbf{WSBDN}: It extends WSDDN by incorporating a background class shape into each proposal's feature representation, and leverages merged pseudo GT boxes obtained via HGPS to impose initial constraints on its class-wise softmax branch. \textbf{NCS}: For proposals ignored after IoU-based partitioning, their scores for categories confirmed to be absent in the image are trained toward zero. All modules are used during training, while only the backbone and the $K$-th IR module are used during inference.}
\label{fig:pipeline}
\end{figure*}

The entire DANCE framework is shown in Figure \ref{fig:pipeline}. In HGPS, we apply a high and a low threshold to each category-specific heatmap. Then, the proposals situated between each high-threshold box and the corresponding low-threshold box are grouped together to form a cluster. For training, pseudo GT boxes are dynamically generated by selecting the top-scoring proposal from each cluster, based on the class-specific scores from the preceding module. In WSBDN, all proposals from all clusters are directly treated as pseudo GT boxes to supervise the class-wise softmax branch. In NCS, we leverage negative certainty information to formulate supervision on ignored proposals further, ensuring a consistent optimization trajectory for each proposal.

\subsection{Heatmap-Guided Proposal Selector} \label{sec:3-3}

In this section, we elaborate on our proposed HGPS algorithm, which includes two main steps: The first step is to construct pseudo GT clusters by heatmaps and proposals. The second step is to select pseudo GT boxes from clusters by combining proposal scores. Because our operations are performed independently for each category, we will omit the category subscript $c$ in the subsequent notation for simplicity and clarity. Detailed motivation is stated in Appendix \ref{app:B}.

\subsubsection{Construction of Pseudo GT Clusters} \label{sec:3-3-1}

\begin{figure*}[t]
\centering
\vspace{-0.8em}
\subfloat[]{
\includegraphics[width=0.162\textwidth]{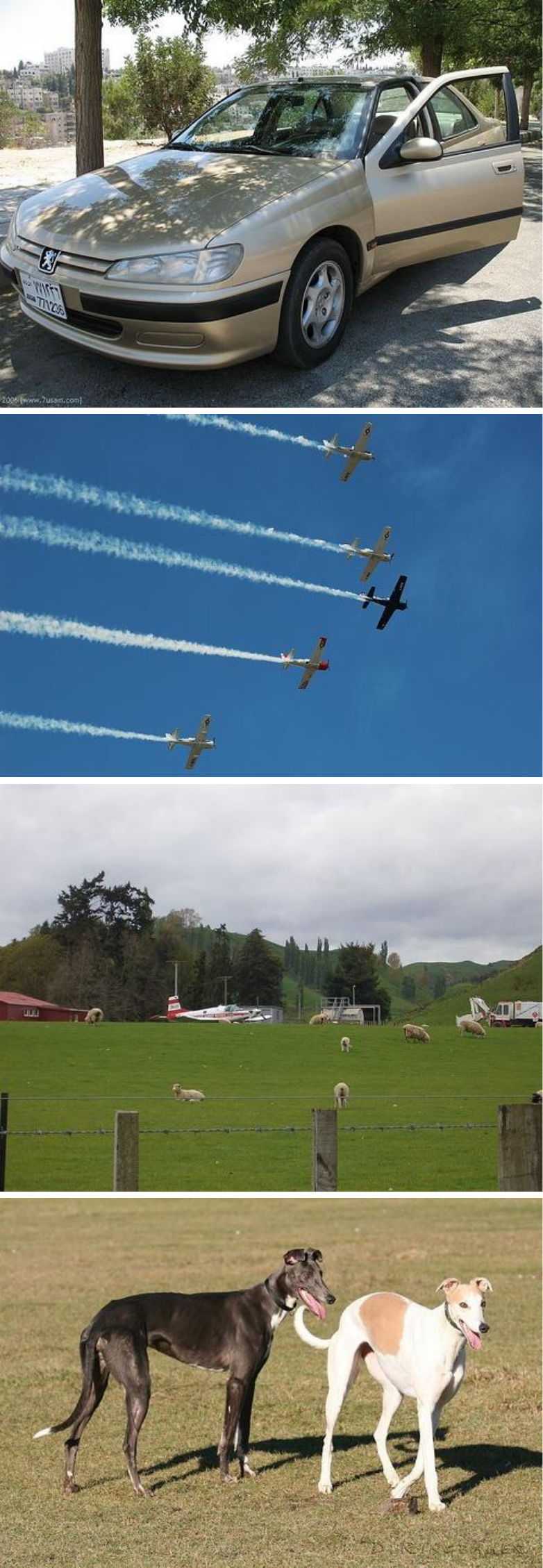}
\label{fig:hgps-vis-a}
}
\hspace{-9pt}
\subfloat[]{
\includegraphics[width=0.162\textwidth]{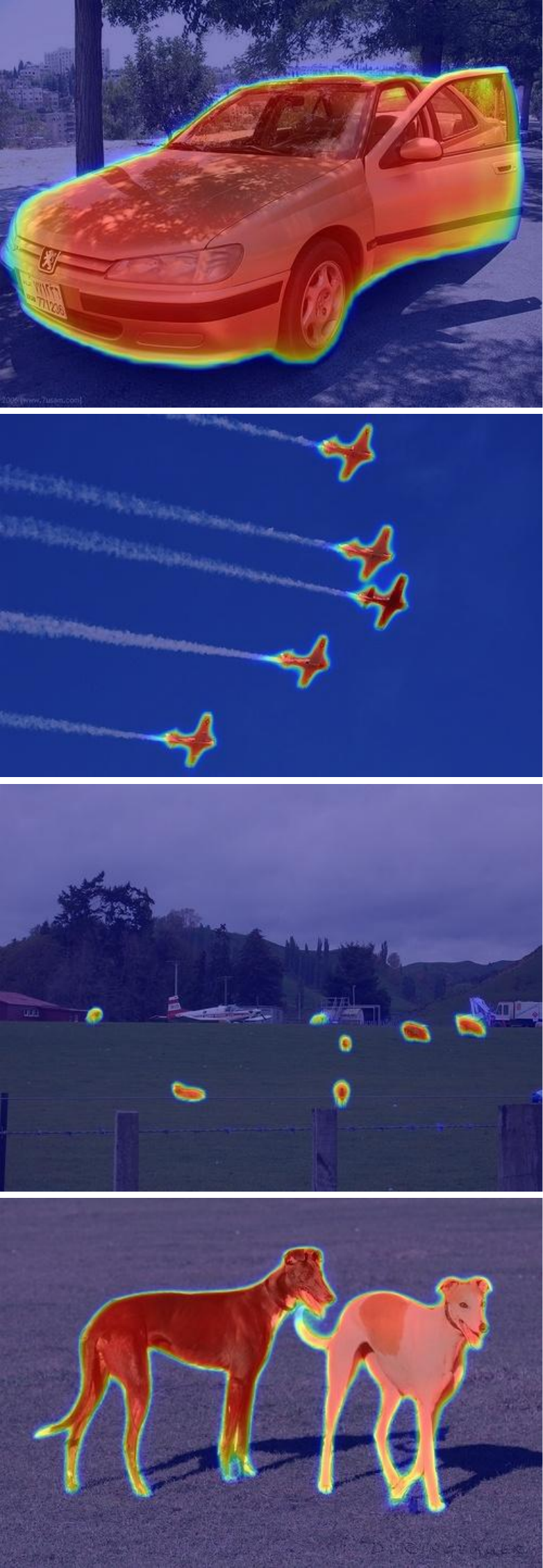}
\label{fig:hgps-vis-b}
}
\hspace{-9pt}
\subfloat[]{
\includegraphics[width=0.162\textwidth]{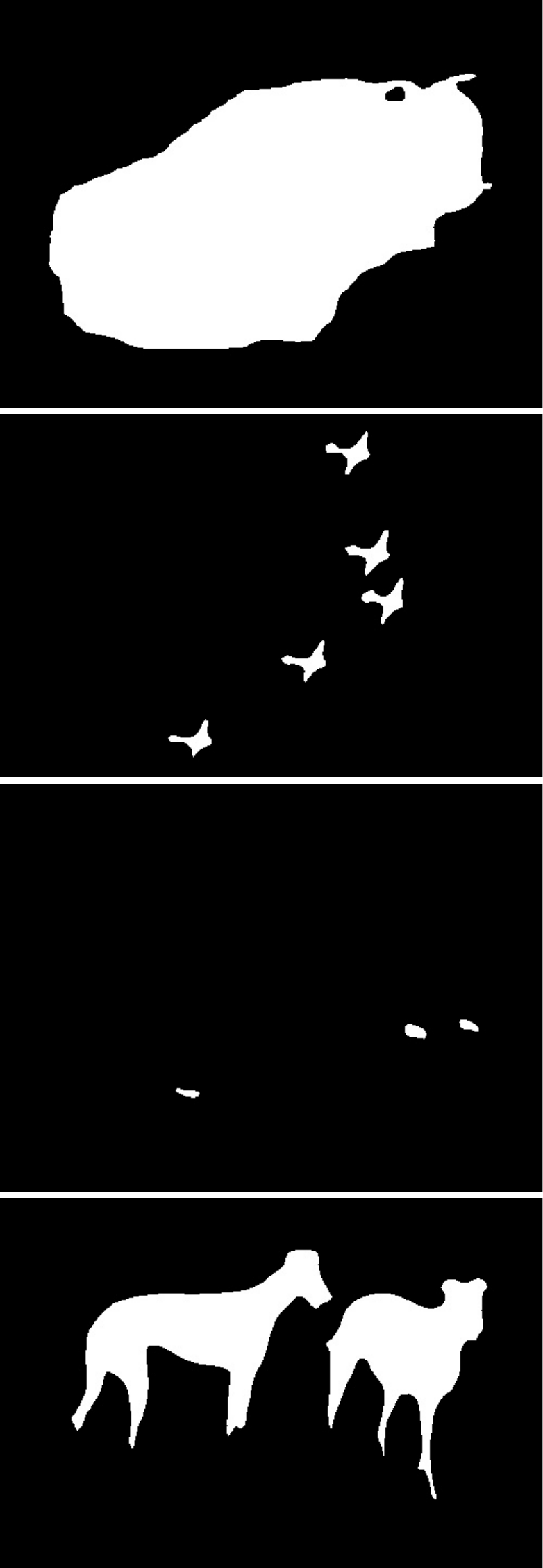}
\label{fig:hgps-vis-c}
}
\hspace{-9pt}
\subfloat[]{
\includegraphics[width=0.162\textwidth]{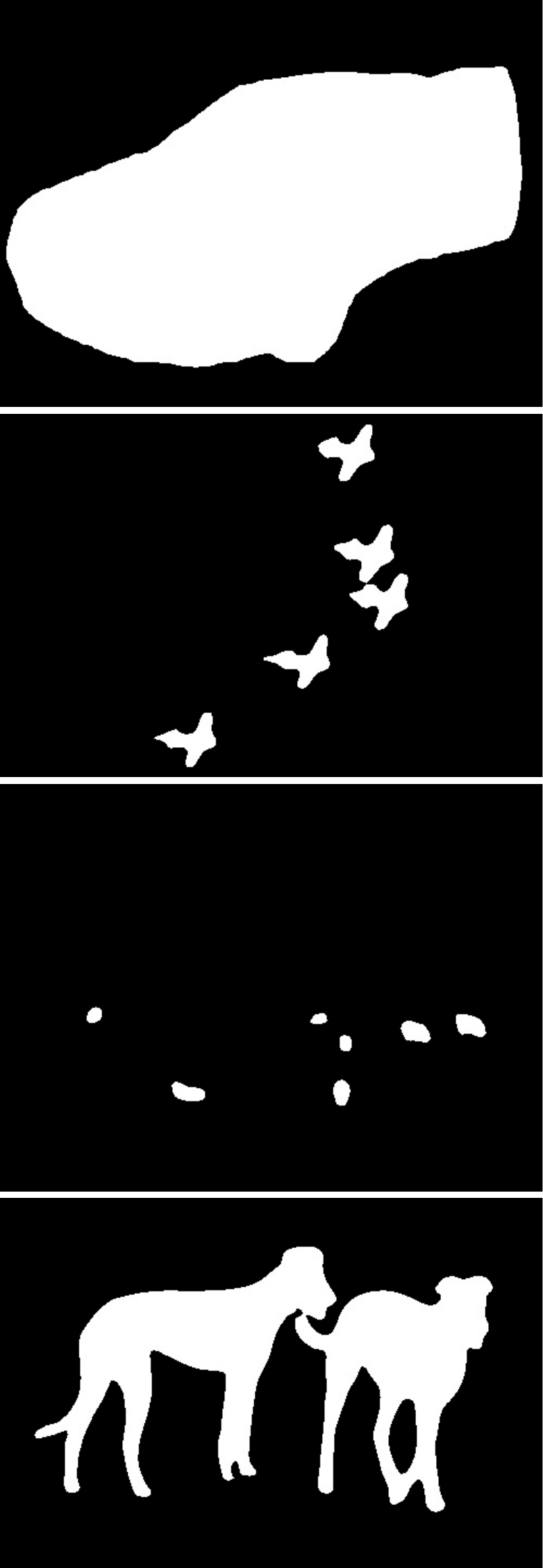}
\label{fig:hgps-vis-d}
}
\hspace{-9pt}
\subfloat[]{
\includegraphics[width=0.162\textwidth]{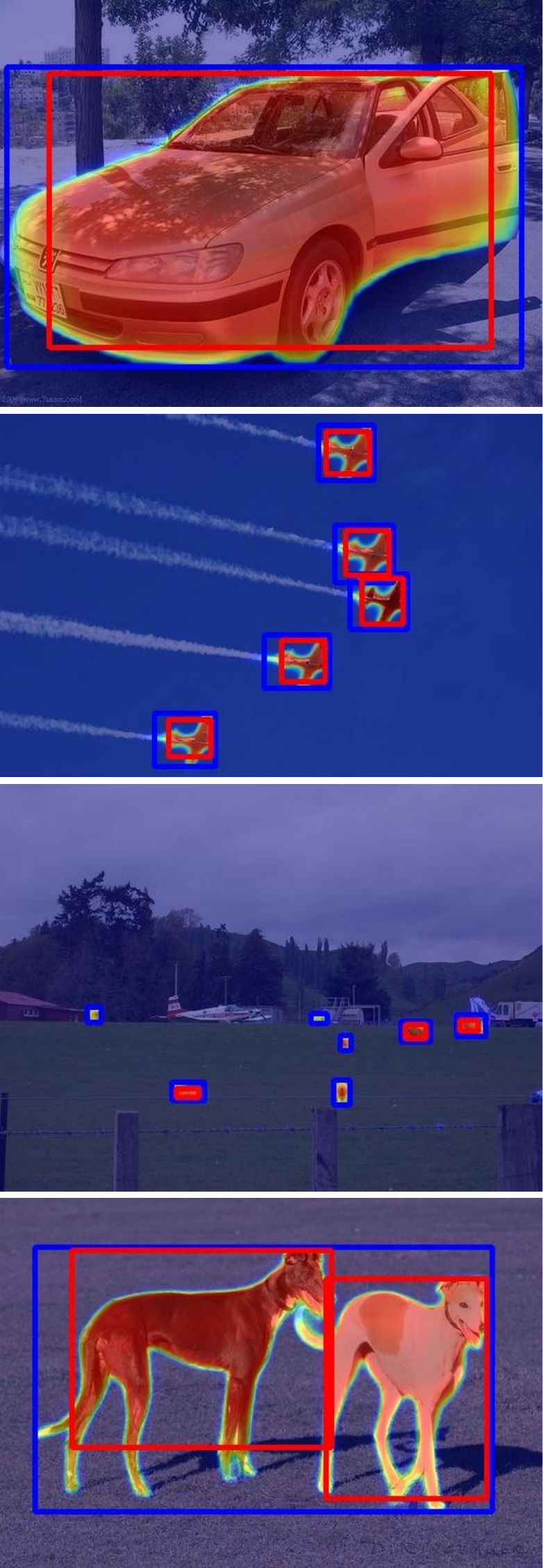}
\label{fig:hgps-vis-e}
}
\hspace{-9pt}
\subfloat[]{
\includegraphics[width=0.162\textwidth]{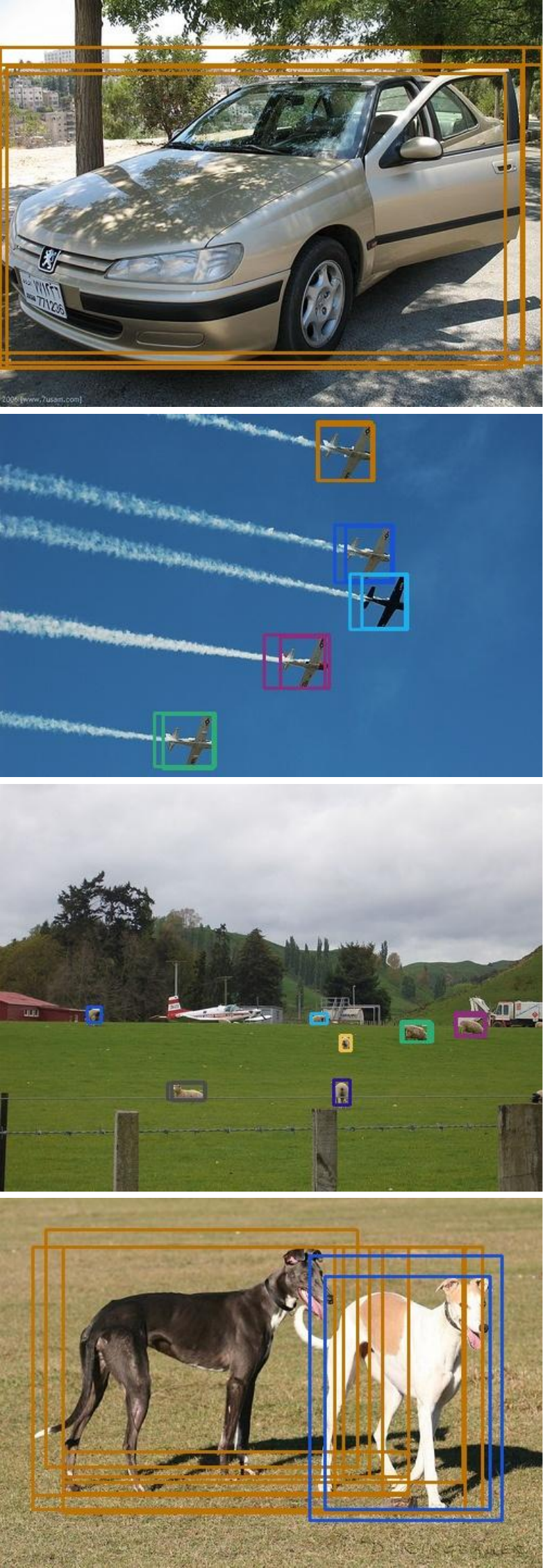}
\label{fig:hgps-vis-f}
}
\caption{Visualizations of our proposed HGPS algorithm. (a) original image. (b) category-specific heatmap. (c) high threshold mask. (d) low threshold mask. (e) high \& low threshold boxes. (f) pseudo GT clusters. In (e), the first two rows reveal $M_n = 1$, the third row reveals both $M_n = 0$ and $M_n = 1$, and the fourth row reveals $M_n \geq 2$ as described in \ref{sec:3-3-1}.}
\label{fig:hgps-vis}
\end{figure*}

Given an image $\bm{I} \in \mathbb{R}^{H \times W \times 3}$, we first employ S2C \cite{Kweon2024S2C} to distill the rich semantic knowledge from the large segmentation model SAM \cite{Kirillov2023SAM} into the CAM \cite{Zhou2016CAM} network, thereby obtaining highly precise class activation maps $\bm{A} \in \mathbb{R}^{H' \times W' \times C}$. For each category $c$ present in $\bm{I}$ (i.e., where the class label $y_c = 1$), we upsample the corresponding map $\bm{A}_c$ via interpolation to original shapes $H \times W$, followed by min-max normalization, which yields category-specific heatmap $\tilde{\bm{A}}_c \in \mathbb{R}^{H \times W}$ satisfying $\min\limits_{h, w} \tilde{\bm{A}}_{h, w, c} = 0$ and $\max\limits_{h, w} \tilde{\bm{A}}_{h, w, c} = 1$. The visualizations of some heatmap examples are shown in Figure \ref{fig:hgps-vis}\subref{fig:hgps-vis-b}.

Next, we apply high and low thresholds to the heatmaps to obtain binary masks, as shown in Figure \ref{fig:hgps-vis}\subref{fig:hgps-vis-c} and \subref{fig:hgps-vis-d}, respectively. We observe that the high-threshold region, while able to distinguish between adjacent intra-class instances, does not cover the complete object information. Conversely, the low-threshold region merges adjacent same-class objects into a single entity but encompasses more complete boundary information. Therefore, we first extract the tightest bounding box for each high- and low-threshold region, respectively, as illustrated in Figure \ref{fig:hgps-vis}\subref{fig:hgps-vis-e}. Then, we filter for proposals that lie between each high-threshold box and its corresponding low-threshold box to form proposal clusters. This process pre-selects proposals that perfectly bound a single object, serving them as the candidate set for our pseudo GT boxes.

\begin{figure}[!t]
\centering
\includegraphics[width=0.85\columnwidth]{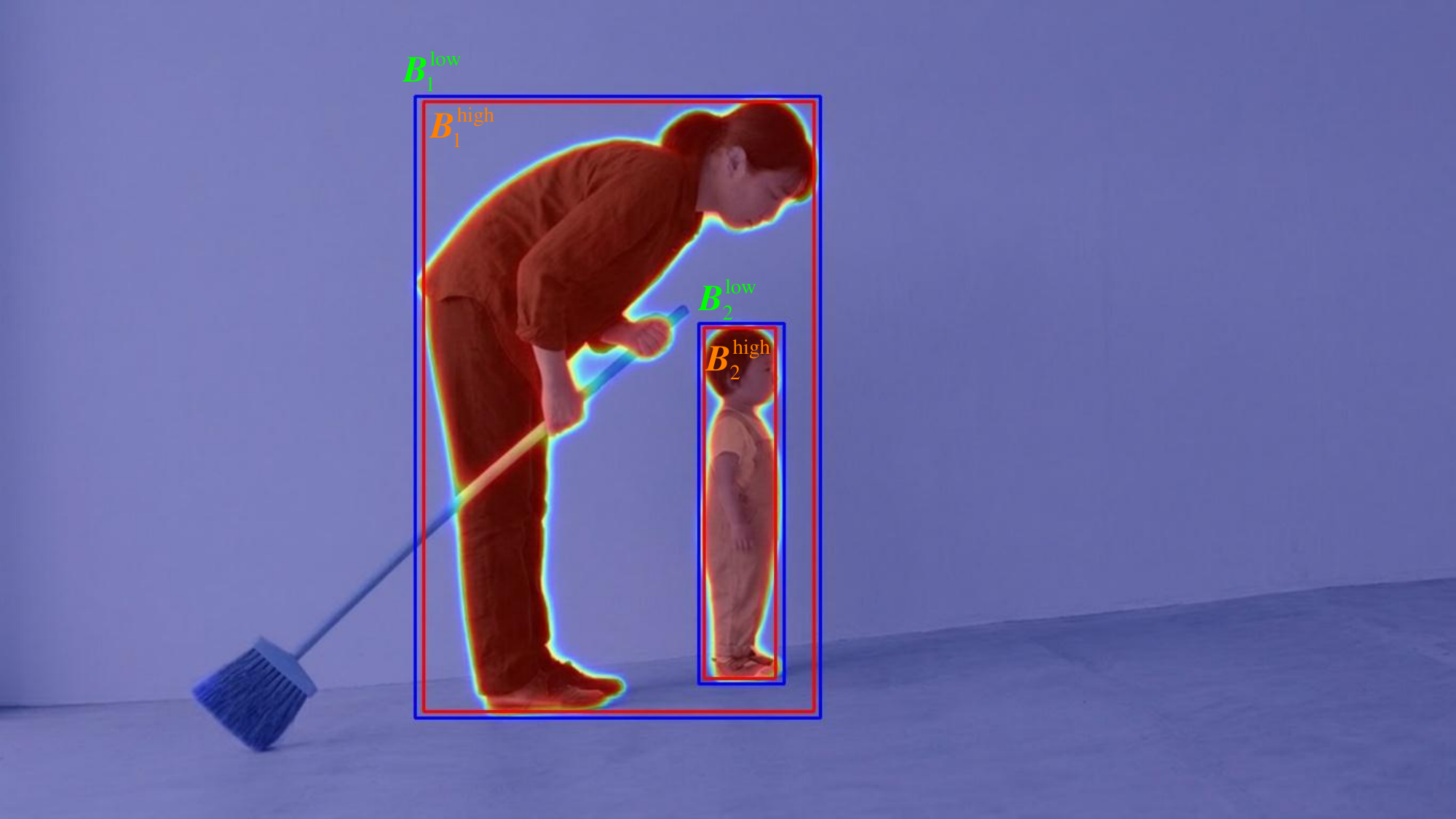}
\caption{Subordinate relationship between tightest bounding boxes. Although $\bm{B}_2^{\mathrm{high}}$ is fully contained within $\bm{B}_1^{\mathrm{low}}$, it is not ``subordinate'' to $\bm{B}_1^{\mathrm{low}}$. This is because their respective masks do not satisfy the containment relationship. In this case, $\bm{B}_2^{\mathrm{high}}$ is subordinate to $\bm{B}_2^{\mathrm{low}}$, while $\bm{B}_1^{\mathrm{high}}$ is subordinate to $\bm{B}_1^{\mathrm{low}}$.}
\label{fig:subordinate}
\end{figure}

Here, we define the tightest bounding box of each thresholded region as a ``threshold box''. Since our heatmaps are normalized, there must exist at least one high-threshold box and one low-threshold box. Because the values of points within a high-threshold connected region are all greater than the low threshold, each high-threshold connected region is necessarily contained within at least one low-threshold connected region. Furthermore, as these regions are connected, each high-threshold connected region must be contained within exactly one low-threshold connected region. Therefore, we define a ``subordinate'' relationship between high- and low-threshold boxes: if a high-threshold connected region is wholly contained within a low-threshold connected region, we say that this high-threshold box is subordinate to that low-threshold box. As shown in Figure \ref{fig:subordinate}, our ``subordinate'' relationship does not directly rely on the containment between boxes, because box containment is only a necessary but insufficient condition for region containment.

Specifically, With the low threshold $\tau^{\mathrm{low}}$, we define $\bm{\mathcal{B}}^{\mathrm{low}} = \left\{\bm{B}_1^{\mathrm{low}}, \bm{B}_2^{\mathrm{low}}, \cdots, \bm{B}_N^{\mathrm{low}}\right\}$ as the set of low-threshold boxes, where $N$ is the number of low-threshold boxes for class $c$. Correspondingly, With the high threshold $\tau^{\mathrm{high}}$, let $M$ be the total number of high-threshold boxes, and $\bm{\mathcal{B}}_n^{\mathrm{high}} = \left\{\bm{B}_{n, 1}^{\mathrm{high}}, \bm{B}_{n, 2}^{\mathrm{high}}, \cdots, \bm{B}_{n, M_n}^{\mathrm{high}}\right\}$ as the set of high-threshold boxes, where $M_n$ is the number of high-threshold boxes subordinate to low-threshold box $\bm{B}_n^{\mathrm{low}}$ and $\sum_{n = 1}^{N} M_n = M$.

According to the number $M_n$ of high-threshold boxes subordinate to the low-threshold box $\bm{B}_n^{\mathrm{low}}$, we build pseudo GT clusters under the following three cases:

\begin{itemize}
    \item $M_n = 0$. It indicates that no high-threshold box is subordinate to the low-threshold box $\bm{B}_n^{\mathrm{low}}$. We directly assign $\left\{\bm{B}_n^{\mathrm{low}}\right\}$ as an independent singleton cluster.
    \item $M_n = 1$. It indicates exactly one high-threshold box $\bm{B}_{n, 1}^{\mathrm{high}}$ is subordinate to the low-threshold box $\bm{B}_n^{\mathrm{low}}$. We first obtain $\bm{B}_n^{r, \mathrm{low}}$ by enlarging the low-threshold box $\bm{B}_n^{\mathrm{low}}$ with a factor of $r$. We then group all proposals, which are spatially located between $\bm{B}_{n, 1}^{\mathrm{high}}$ and $\bm{B}_n^{r, \mathrm{low}}$, together with the original low-threshold box $\bm{B}_n^{\mathrm{low}}$ into a set to form an independent cluster.
    \item $M_n \geq 2$. We build an individual cluster for each high-threshold box $\bm{B}_{n, m_n}^{\mathrm{high}}$, where $m_n = 1, 2, \cdots, M_n$. We first enlarge $\bm{B}_n^{\mathrm{low}}$ and $\bm{B}_{n, 1}^{\mathrm{high}}, \bm{B}_{n, 2}^{\mathrm{high}}, \cdots, \bm{B}_{n, M_n}^{\mathrm{high}}$ by factor $r$ to obtain $\bm{B}_n^{r, \mathrm{low}}$ and $\bm{B}_{n, 1}^{r, \mathrm{high}}, \bm{B}_{n, 2}^{r, \mathrm{high}}, \cdots, \bm{B}_{n, M_n}^{r, \mathrm{high}}$. Then, for each high-threshold box $\bm{B}_{n, m_n}^{\mathrm{high}}$, we group all proposals spatially located between $\bm{B}_{n, m_n}^{\mathrm{high}}$ and $\bm{B}_n^{r, \mathrm{low}}$, together with the enlarged high-threshold box $\bm{B}_{n, m_n}^{r, \mathrm{high}}$ into a set to form an independent cluster. However, if a proposal contains multiple high-threshold boxes, we will only assign it to the cluster with which the scaled high-threshold box has the maximum IoU value. This one-to-one assignment ensures no box is duplicated across clusters, which in turn guarantees that the pseudo GT boxes selected from each cluster are different.
\end{itemize}

\begin{algorithm}[!t]
\caption{Construction of Pseudo GT Clusters}
\label{alg:pseudo-GT-clusters}
\footnotesize
\textbf{Input}: Image $\bm{I} \in \mathbb{R}^{H \times W \times 3}$; Image-level labels $\bm{y} = [y_1, y_2, \cdots,$ ${y_C]}^{T} \in \mathbb{R}^{C}$; Region proposals $\bm{\mathcal{P}} = \{\bm{P}_1, \bm{P}_2, \dots, \bm{P}_R\}$; S2C module\\
\textbf{Parameter}: High threshold $\tau^{\mathrm{high}}$; Low threshold $\tau^{\mathrm{low}}$; Rescaling factor $r$ for boxes\\
\textbf{Output}: Pseudo GT cluster lists $\mathbb{P}$
\begin{algorithmic}[1]
\STATE Class activation maps $\bm{A} = \mathrm{S2C} \left(\bm{I}\right)$.
\STATE Category-specific heatmaps $\tilde{\bm{A}} = \mathrm{Normalize}\left(\bm{A}\right)$.
\FOR{each category $c=1$ to $C$}
    \STATE Let pseudo GT cluster list $\mathbb{P}_c = \varnothing$.
    \IF{$y_c = 1$}
        \STATE Let $\xi=1$.
        \STATE Using $\tau^{\mathrm{low}}$ on $\tilde{\bm{A}}_c$ to get $N$ low-threshold boxes $\bm{\mathcal{B}}^{\mathrm{low}} = \left\{\bm{B}_1^{\mathrm{low}}, \bm{B}_2^{\mathrm{low}}, \cdots, \bm{B}_N^{\mathrm{low}}\right\}$ and enlarge them by the factor of $r$ to get $\bm{\mathcal{B}}^{r, \mathrm{low}} = \left\{\bm{B}_1^{r, \mathrm{low}}, \bm{B}_2^{r, \mathrm{low}}, \cdots, \bm{B}_N^{r, \mathrm{low}}\right\}$.
        \STATE Using $\tau^{\mathrm{high}}$ on $\tilde{\bm{A}}_c$ to get $M_n$ high-threshold boxes $\bm{\mathcal{B}}_n^{\mathrm{high}} = \left\{\bm{B}_{n, 1}^{\mathrm{high}}, \bm{B}_{n, 2}^{\mathrm{high}}, \cdots, \bm{B}_{n, M_n}^{\mathrm{high}}\right\}$ corresponding to each low-threshold box $\bm{B}_n^{\mathrm{low}}$ and enlarge them by the factor of $r$ to get $\bm{\mathcal{B}}_n^{r, \mathrm{high}} = \left\{\bm{B}_{n, 1}^{r, \mathrm{high}}, \bm{B}_{n, 2}^{r, \mathrm{high}}, \cdots, \bm{B}_{n, M_n}^{r, \mathrm{high}}\right\}$.
        \FOR{$n=1$ to $N$}
            \IF{$M_n = 0$}
                \STATE Build pseudo GT cluster $\bm{\mathcal{P}}_\xi = \left\{\bm{B}_n^{\mathrm{low}}\right\}$.
                \STATE $\mathbb{P}_c.\mathrm{add}\left(\bm{\mathcal{P}}_\xi\right)$.
                \STATE $\xi=\xi+1$.
            \ELSIF{$M_n = 1$}
                \STATE Find the $\hat{M}_{n, 1}$ proposals $\bm{P}_{r_1^{n, 1}}$, $\bm{P}_{r_2^{n, 1}}$, $\cdots$, $\bm{P}_{r_{\hat{M}_{n, 1}}^{n, 1}}$ spatially located between $\bm{B}_{n, 1}^{\mathrm{high}}$ and $\bm{B}_n^{r, \mathrm{low}}$.
                \STATE Build pseudo GT cluster $\bm{\mathcal{P}}_\xi = \Bigg\{\bm{B}_n^{\mathrm{low}}, \bm{P}_{r_1^{n, 1}}, \bm{P}_{r_2^{n, 1}}, \cdots$, $\bm{P}_{r_{\hat{M}_{n, 1}}^{n, 1}}\Bigg\}$.
                \STATE $\mathbb{P}_c.\mathrm{add}\left(\bm{\mathcal{P}}_\xi\right)$.
                \STATE $\xi=\xi+1$.
            \ELSE
                \FOR{$m_n = 1$ to $M_n$}
                    \STATE Find the $\hat{M}_{n, m_n}$ proposals $\bm{P}_{r_1^{n, m_n}}$, $\bm{P}_{r_2^{n, m_n}}$, $\cdots$, $\bm{P}_{r_{\hat{M}_{n, m_n}}^{n, m_n}}$ spatially located between $\bm{B}_{n, m_n}^{\mathrm{high}}$ and $\bm{B}_n^{r, \mathrm{low}}$.
                    \STATE Build pseudo GT cluster $\bm{\mathcal{P}}_{\xi+m_n-1} = \bigg\{\bm{B}_{n, m_n}^{r, \mathrm{high}}$, $\bm{P}_{r_1^{n, m_n}}, \bm{P}_{r_2^{n, m_n}}, \cdots, \bm{P}_{r_{\hat{M}_{n, m_n}}^{n, m_n}}\bigg\}$.
                \ENDFOR
                \STATE Remain the proposal $\bm{P}_r$ only in the cluster having the maximum IoU with $\bm{B}_{n, m_n}^{r, \mathrm{high}}$ if it appears in multiple clusters.
                \FOR{$m_n = 1$ to $M_n$}
                    \STATE $\mathbb{P}_c.\mathrm{add}\left(\bm{\mathcal{P}}_{\xi+m_n-1}\right)$.
                \ENDFOR
                \STATE $\xi=\xi+M_n$.
            \ENDIF
        \ENDFOR
    \ENDIF
\ENDFOR
\STATE \textbf{return} Pseudo GT cluster lists $\mathbb{P}$.
\end{algorithmic}
\end{algorithm}

Obviously, $\bm{B}_n^{\mathrm{low}}$ and $\bm{B}_{n, m_n}^{r, \mathrm{high}}$ must be spatially located between $\bm{B}_{n, m_n}^{\mathrm{high}}$ and $\bm{B}_n^{r, \mathrm{low}}$, so we add them into their respective set to ensure there is at least one bounding box in each cluster. The whole construction process is summarized in Algorithm \ref{alg:pseudo-GT-clusters}, resulting in the pseudo GT cluster lists $\mathbb{P}$.

\subsubsection{Selecting Pseudo GT Boxes from Pseudo GT Clusters} \label{sec:3-3-2}

\begin{algorithm}[!t]
\caption{Get Pseudo GT boxes for each IR module}
\label{alg:pseudo-GT-boxes-k}
\textbf{Input}: Image-level labels $\bm{y} = {[y_1, y_2, \cdots, y_C]}^{T} \in \mathbb{R}^{C}$; Pseudo GT cluster lists $\mathbb{P}$; Score matrix $\bm{s}^{\left(k-1\right)} \in \mathbb{R}^{R \times \left(C+1\right)}$ if $k > 1$ else $\bm{ws}^{(0)} \in \mathbb{R}^{R \times \left(C+1\right)}$ for all proposals output from the $\left(k-1\right)$-th module\\
\textbf{Output}: Pseudo GT boxes $\bm{\mathcal{T}}^{(k)}$ ($k \geq 1$) for supervising the $k$-th IR module
\begin{algorithmic}[1]
\STATE Let $\bm{\mathcal{T}}^{(k)} = \varnothing$.
\STATE Let $j = 0$.
\FOR{each category $c=1$ to $C$}
    \IF{$y_c=1$}
        \FOR{each pseudo GT cluster $\bm{\mathcal{P}}_\xi = \bigg\{\bm{P}_{r_1^\xi}, \bm{P}_{r_2^\xi}, \cdots$, $\bm{P}_{r_{N_\xi}^\xi}\bigg\}$ in $\mathbb{P}_c$}
            \IF{$k = 1$}
                \STATE Let $n_\xi^* = \arg \max_{n_\xi \in \left\{1, 2, \cdots, N_\xi\right\}} ws_{r_{n_\xi}^\xi, c}^{\left(0\right)}$.
            \ELSE
                \STATE Let $n_\xi^* = \arg \max_{n_\xi \in \left\{1, 2, \cdots, N_\xi\right\}} s_{r_{n_\xi}^\xi, c}^{\left(k-1\right)}$.
            \ENDIF
            \STATE Let $\bm{T}_j^{(k)} = \bm{P}_{r_{n_\xi^*}^\xi}$, $s_j^{(k)} = s_{r_{n_\xi^*}^\xi, c}^{\left(k-1\right)}$, $\hat{y}_j^{(k)} = c$.
            \STATE $\bm{\mathcal{T}}^{(k)}.\mathrm{add}\left(\left(\bm{T}_j^{(k)}, s_j^{(k)}, \hat{y}_j^{(k)}\right)\right)$.
            \STATE $j = j + 1$.
        \ENDFOR
    \ENDIF
\ENDFOR
\STATE \textbf{return} Pseudo GT boxes $\bm{\mathcal{T}}^{(k)}$.
\end{algorithmic}
\end{algorithm}

We leverage the pseudo GT cluster lists $\mathbb{P}$ in conjunction with the score matrix from the $(k-1)$-th IR module to generate pseudo GT boxes. For each cluster, we select the top-scoring box in its corresponding category. Specifically, for each pseudo GT cluster $\bm{\mathcal{P}}_\xi = \left\{\bm{P}_{r_1^\xi}, \bm{P}_{r_2^\xi}, \cdots, \bm{P}_{r_{N_\xi}^\xi}\right\}$ within list $\mathbb{P}_c$ (where $y_c=1$), we first obtain the score vector $\bm{s}_{r_1^\xi}^{\left(k-1\right)}, \bm{s}_{r_2^\xi}^{\left(k-1\right)}, \dots, \bm{s}_{r_{N_\xi}^\xi}^{\left(k-1\right)}$ of each box from the $(k-1)$-th IR module. Subsequently, we identify the proposal with the highest score on category $c$ in each cluster of which the index $r_{n_\xi^*}^\xi$ is determined by $n_\xi^* = \arg \max_{n_\xi \in \left\{1, 2, \cdots, N_\xi\right\}} s_{r_{n_\xi}^\xi, c}^{\left(k-1\right)}$. The corresponding proposal $\bm{P}_{r_{n_\xi^*}^\xi}$ is then selected as one of the pseudo GT boxes. The detailed process is summarized in Algorithm \ref{alg:pseudo-GT-boxes-k}, obtaining pseudo GT boxes $\bm{\mathcal{T}}^{(k)}$.

Then, each proposal is assigned a pseudo GT box, governed by the IoU thresholds $0 < \tau_2^{\mathrm{IoU}} < \tau_1^{\mathrm{IoU}} < 1$, to supervise the $k$-th IR module:
\begin{equation} \label{equ:pseudo-labels}
    y_r^{(k)} = \begin{cases}
        \hat{y}_j^{(k)}, & t_{r, j}^{(k)} \geq \tau_1^{\mathrm{IoU}} \\
        C+1, & \tau_2^{\mathrm{IoU}} \leq t_{r, j}^{(k)} < \tau_1^{\mathrm{IoU}} \\
        -1, & t_{r, j}^{(k)} < \tau_2^{\mathrm{IoU}}
    \end{cases}
\end{equation}
where $t_{r, j}^{(k)} = \max_{\left(\bm{T}_j^{(k)}, s_j^{(k)}, \hat{y}_j^{(k)}\right) \in \bm{\mathcal{T}}^{(k)}} \mathrm{IoU} \left(\bm{P}_r, \bm{T}_j^{(k)}\right)$ and $y_r^{(k)} = -1$ means the proposal ignored during training. We also weight the cross-entropy loss for each proposal like prior works \cite{Tang2017OICR, Tang2018PCL}. We set the weight $w_r^{(k)}$ to the confidence score $s_j^{(k)}$ of the pseudo GT box that has the maximum IoU with it. Thus, the final training objective for the classification loss in the $k$-th IR module is formulated as:
\begin{equation}
    \mathcal{L}_{\mathrm{cls}}^{(k)} = -\frac{1}{R_{\mathrm{cls}}^{(k)}} \sum_{r=1}^{R} w_r^{(k)} \sum_{c=1}^{C+1} \mathbbm{1}\left[y_r^{(k)} = c\right] \log s_{r, c}^{(k)},
\end{equation}
where $R_{\mathrm{cls}}^{(k)}$ denotes the number of proposals satisfying $y_r^{(k)} \neq -1$.

\subsection{Weakly Supervised Basic Detection Network}

In this section, we introduce the design of the WSBDN module. It is worth noting that the score vector computed by WSDDN \cite{Bilen2016WSDDN} for each proposal omits the background class dimension. This is clearly inconsistent with reality and inevitably confounds the scores of foreground categories, inherently limiting the absolute correctness of the predictions at the architectural level. Furthermore, relying solely on an image-level loss at the end imposes merely an implicit constraint on $\bm{s}^{(0)}$. Due to this lack of explicit supervision, $\bm{s}^{(0)}$ can converge in arbitrary directions; as long as it couples with a complementary $\bm{w}^{(0)}$, the final image-level scores can still appear correct. This phenomenon severely compromises the intrinsic accuracy of $\bm{s}^{(0)}$, making it highly noisy. Therefore, we reintroduce the background class representation into the model and modify the label formulation accordingly. Additionally, we utilize heatmaps obtained from preprocessing to derive initial pseudo GT boxes with a certain level of confidence. These boxes are applied to directly supervise on $\bm{s}^{(0)}$, thereby boosting the intrinsic performance of the baseline model. Detailed motivation is stated in Appendix \ref{app:C}.

We first align the label dimensions of WSOD and FSOD, which enables the model's output for each proposal to change from $\mathbb{R}^C$ to $\mathbb{R}^{C+1}$. Specifically, we push the label definition down one level: we posit that $y_c=1$ or $0$ indicates whether there are any proposals representing class $c$ among all proposals in an image. Given the evident fact that every image must contain proposals that represent the background, it naturally follows the assumption that we assign $y_{C+1}=1$ for all images. We call this newly defined label $\bm{y} \in \mathbb{R}^{C+1}$ the ``box-level image label''.

Under this setting, we employ two parallel branches to compute a classification vector $\bm{\varphi}_r^{\mathrm{cls}} \in \mathbb{R}^{C+1}$ and a weighting vector $\bm{\varphi}_r^{\mathrm{wgt}} \in \mathbb{R}^{C+1}$, respectively, for each proposal $\bm{P}_r$. These vectors are then processed by a class-wise softmax and a proposal-wise softmax, respectively, yielding a score matrix $\bm{s}^{(0)} = \mathrm{Softmax}_{\mathrm{cls}} \left(\bm{\varphi}^{\mathrm{cls}}\right) \in \mathbb{R}^{R \times (C+1)}$ and a weight matrix $\bm{w}^{(0)} = \mathrm{Softmax}_{\mathrm{pro}} \left(\bm{\varphi}^{\mathrm{wgt}}\right) \in \mathbb{R}^{R \times (C+1)}$, where $s_{r, c}^{\left(0\right)} = \frac{\exp \left(\varphi_{r, c}^{\mathrm{cls}}\right)}{\sum_{c'=1}^{C+1} \exp \left(\varphi_{r, c'}^{\mathrm{cls}}\right)}$ and $w_{r, c}^{\left(0\right)} = \frac{\exp \left(\varphi_{r, c}^{\mathrm{wgt}}\right)}{\sum_{r'=1}^{R} \exp \left(\varphi_{r', c}^{\mathrm{wgt}}\right)}$. Subsequently, we compute the final weighted score matrix $\bm{ws}^{(0)} \in \mathbb{R}^{R \times (C+1)}$ through element-wise product: $\bm{ws}^{(0)} = \bm{s}^{(0)} \odot \bm{w}^{(0)}$. This matrix is then summed over the proposal dimension to yield the final box-level image prediction score $\bm{s}^{\mathrm{img}} \in \mathbb{R}^{C+1}$, where $s_c^{\mathrm{img}} = \sum_{r=1}^{R} ws_{r, c}^{(0)}$. At last, the image loss is built by a binary cross-entropy loss function:
\begin{equation}
\resizebox{0.88\columnwidth}{!}{$\displaystyle
    \mathcal{L}_{\mathrm{img}} = -\sum_{c=1}^{C+1} \left[y_c \log s_c^{\mathrm{img}} + \left(1 - y_c\right) \log \left(1 - s_c^{\mathrm{img}}\right)\right].
$}
\end{equation}

As present in Table \ref{tab:WSDDN}, there exists a significant semantic misalignment between $\bm{s}^{(0)}$ and $\bm{ws}^{(0)}$. A mere 5.0\% mAP indicates that $\bm{s}^{(0)}$ completely fails to learn correct classification information. From this perspective, the poor performance of $\bm{s}^{(0)}$ also acts as a bottleneck, constraining the potential of the final detection matrix $\bm{ws}^{(0)}$.

\begin{algorithm}[!t]
\caption{Get Pseudo GT boxes for WSBDN module.}
\label{alg:pseudo-GT-boxes-0}
\textbf{Input}: Image-level labels $\bm{y} = {[y_1, y_2, \cdots, y_C]}^{T} \in \mathbb{R}^{C}$; Pseudo GT cluster lists $\mathbb{P}$\\
\textbf{Output}: Pseudo GT boxes $\bm{\mathcal{T}}^{(0)}$ for supervising the WSBDN module
\begin{algorithmic}[1]
\STATE Let $\bm{\mathcal{T}}^{(0)} = \varnothing$.
\STATE Let $j = 0$.
\FOR{each category $c=1$ to $C$}
    \IF{$y_c=1$}
        \FOR{each pseudo GT cluster $\bm{\mathcal{P}}_{c, i} = \left\{\bm{P}_{r_1^{c, i}}, \bm{P}_{r_2^{c, i}}, \cdots, \bm{P}_{r_{N_{c, i}}^{c, i}}\right\}$ in $\mathbb{P}_c$}
            \FOR{each proposal $\bm{P}_{r_{n_{c, i}}^{c, i}}$ in $\bm{\mathcal{P}}_{c, i}$}
                \STATE Let $\bm{T}_j^{(0)} = \bm{P}_{r_{n_{c, i}}^{c, i}}$, $\hat{y}_j^{(0)} = c$.
                \STATE $\bm{\mathcal{T}}^{(0)}.\mathrm{add}\left(\left(\bm{T}_j^{(0)}, \hat{y}_j^{(0)}\right)\right)$.
                \STATE $j = j + 1$.
            \ENDFOR
        \ENDFOR
    \ENDIF
\ENDFOR
\STATE \textbf{return} Pseudo GT boxes $\bm{\mathcal{T}}^{(0)}$.
\end{algorithmic}
\end{algorithm}

Therefore, in order to make $\bm{s}^{(0)}$ no longer noisy, we supervise it through the pseudo GT cluster lists $\mathbb{P}$ pre-obtained from Algorithm \ref{alg:pseudo-GT-clusters}. Since the proposals within the clusters are generally of high quality, we directly treat them all as pseudo GT boxes. The generation of initial pseudo GT boxes $\mathcal{T}^{(0)}$ is summarized in Algorithm \ref{alg:pseudo-GT-boxes-0}.

Because WSBDN is the base module and thus no preceding scores can serve as weight, we do not apply any weight to each proposal's loss in this initial stage. The classification loss for the class-wise softmax branch of WSBDN is formulated as a standard, unweighted cross-entropy loss:
\begin{equation}
    \mathcal{L}_{\mathrm{cls}}^{(0)} = -\frac{1}{R_{\mathrm{cls}}^{(0)}} \sum_{r=1}^{R} \sum_{c=1}^{C+1} \mathbbm{1}\left[y_r^{(0)} = c\right] \log s_{r, c}^{(0)}.
\end{equation}

\subsection{Negative Certainty Supervision}

Beyond the HGPS algorithm and the WSBDN module discussed above, we leverage negative certainty information for supervising the proposals that were designated as ``ignored'' in Equation \ref{equ:pseudo-labels}. The core insight is that, while the specific class of an ignored proposal is ambiguous, we can be certain about the classes it does not belong to. For example, if an image labeled only with ``person'' and ``horse'', the true class of an ignored proposal --- be ``person'', ``horse'', or ``background'' --- remains unknown, but it is definitively that the true class must not be ``cat'', ``aeroplane'' or any other categories. However, entirely omitting these proposals from the supervision process leaves them unconstrained, allowing them to produce high confidence scores on these impossible categories. Such a phenomenon hinders network convergence; it implicitly steers the network towards completely erroneous directions, drastically complicating the optimization process. To capitalize on this, we enforce the scores of these ignored proposals to converge towards 0 for all classes $c'$ that are absent from the image (i.e., where the box-level image label $y_{c'} = 0$), which is formulated as a binary cross-entropy loss:
\begin{equation}
\resizebox{0.9\columnwidth}{!}{$\displaystyle
    \mathcal{L}_{\mathrm{cls-ign}}^{(k)} = -\frac{1}{R_{\mathrm{cls-ign}}^{(k)}} \sum_{r=1}^{R} \sum_{c'=1}^{C+1} \mathbbm{1}\left[y_{c'} = 0\right] \log \left(1 - s_{r, c'}^{(k)}\right),
$}
\end{equation}
where $R_{\mathrm{cls-ign}}^{(k)} = R - R_{\mathrm{cls}}^{(k)}$ denotes the number of proposals satisfying $y_r^{(k)} = -1$. In this way, we mitigate the highly non-convex nature and accelerate the convergence rate.

\subsection{The Overall Training Objectives}

With all the aforementioned loss functions, we define the total loss for the base MIDN module WSBDN as follows:
\begin{equation}
    \mathcal{L}_{\mathrm{WSBDN}} = \mathcal{L}_{\mathrm{img}} + \mathcal{L}_{\mathrm{cls}}^{(0)} + \mathcal{L}_{\mathrm{cls-ign}}^{(0)},
\end{equation}
and the loss for each subsequent HGPS-based IR module is given by:
\begin{equation}
    \mathcal{L}_{\mathrm{HGPS}}^{(k)} = \mathcal{L}_{\mathrm{cls}}^{(k)} + \mathcal{L}_{\mathrm{cls-ign}}^{(k)}.
\end{equation}

Finally, the entire network is trained end-to-end:
\begin{equation}
    \mathcal{L}_{\mathrm{DANCE}} = \mathcal{L}_{\mathrm{WSBDN}} + \sum_{k=1}^{K} \mathcal{L}_{\mathrm{HGPS}}^{(k)}.
\end{equation}

\section{Experiments}

\subsection{Datasets and Evaluation Metrics}

Following standard WSOD protocols, we evaluate our method on three different datasets, Pascal VOC 2007, Pascal VOC 2012 \cite{Everingham2010VOC}, and MS COCO 2014 \cite{Lin2014COCO}. The VOC07 and VOC12 datasets have 9,963 and 22,531 images respectively for 20 object classes, which are both divided into \textit{trainval} (5,011 images for VOC07 and 11,540 images for VOC12) and \textit{test} (4,952 images for VOC07 and 10,991 images for VOC12) sets. During the training phase, the \textit{trainval} set are used to train our model. During the testing phase, two metrics are used for evaluation: the Average Precision (AP) on the \textit{test} set and the Correct Localization (CorLoc) on the \textit{trainval} set. The AP \cite{Everingham2010VOC} is measured following the standard Pascal VOC criterion, where a positive predicted box has at least 50\% overlap with its corresponding ground-truth annotation. The CorLoc \cite{Deselaers2012CorLoc} quantifies localization performance by the percentage of images containing at least one predicted box with at least 50\% overlap to one of the ground-truths. The COCO14 dataset has 164,062 images for 80 classes, which is divided into \textit{train} (82,783 images), \textit{val} (40,504 images) and \textit{test} (40,775 images) sets. Following standard practice, we train our model on the \textit{train} set and evaluate it on the \textit{val} set. We report the mean Average Precision (mAP\textsubscript{[50:95]}) across IoU thresholds ranging from 0.5 to 0.95 with a step size of 0.05, as well as the results at specific IoU threshold of 0.5 (mAP\textsubscript{50}) and 0.75 (mAP\textsubscript{75}) respectively. Furthermore, mAP\textsubscript{\textit{s}}, mAP\textsubscript{\textit{m}} and mAP\textsubscript{\textit{l}} are used to evaluate the performance on small, medium and large objects respectively.

\subsection{Implementation Details} \label{sec:4-2}

Consistent with established practices, we adopt a VGG16 \cite{Simonyan2015VGG} network pre-trained on the ImageNet dataset \cite{Russakovsky2015ImageNet} as our backbone. We replace the last max-pooling layer with an RoI Pooling layer followed by two FC layers. Each proposal is thus mapped to a 4096-length feature vector (i.e., $D=4096$ in Section \ref{sec:3-1}). Following prior works, we utilize MCG \cite{Arbelaez2014MCG} to generate about 2,000 region proposals for each image. The number of IR modules is set to 3 (i.e., $K=3$), a common setting in this field. We adopt all the original hyperparameter settings for our category-specific heatmaps extractor S2C \cite{Kweon2024S2C}. For our proposed HGPS algorithm, we set the high threshold $\tau^{\mathrm{high}} = 0.8$, the low threshold $\tau^{\mathrm{low}} = 0.3$, and the scaling factor $r = 1.2$. During training, the IoU thresholds described in Section \ref{sec:3-3-2} are set to $\tau_1^{\mathrm{IoU}} = 0.5$ and $\tau_2^{\mathrm{IoU}} = 0.1$. During inference, the IoU threshold for NMS is set to 0.3.

We employ data augmentation during both training and testing. For training, we fix the original aspect ratio of each image and resize the shorter side to one of these 22 scales randomly: \{480, 512, 544, 576, 608, 640, 672, 704, 736, 768, 800, 832, 864, 896, 928, 960, 992, 1,024, 1,056, 1,088, 1,120, 1,152\}, while ensuring the longer side not exceeding 4,000. Furthermore, we randomly flip each image in the horizontal direction during training. For testing, we adopt the test time augmentation (TTA) strategy. We resize the shorter side of each image to 8 different scales: {480, 576, 672, 768, 864, 960, 1,056, 1,152}, and test both the original image and its horizontal flip, resulting in 16 augmented views. The final prediction is obtained by averaging the results of all 16 views.

Our model is trained for 40 epochs on both the Pascal VOC 2007 and 2012 datasets, and for 10 epochs on the MS COCO 2014 dataset. The batch size is set to 8. For all the FC layers in MIDN and IR modules, their weights are initialized by Xavier \cite{Glorot2010Xavier}, and biases are initialized to 0. We use the Stochastic Gradient Descent (SGD) \cite{Goodfellow2016DL} optimizer with a weight decay of 0.0005. The initial learning rate is set to 0.002 and is decayed by 0.1 at the end of the 24th epoch.

Our experiments are implemented based on the Detectron2 \cite{Wu2019Detectron2} framework. We run all the experiments on 4 NVIDIA A100-SXM4 GPUs, each with 40GB of VRAM.

\subsection{Comparison with State-of-the-arts}

\begin{table*}[!t]
\caption{Comparison with the state-of-the-art methods on Pascal VOC 2007 test set in terms of AP (\%)}
\label{tab:mAP-VOC07}
\centering
\newcolumntype{P}{>{\centering\arraybackslash}p{21pt}}
\newcolumntype{M}{>{\centering\arraybackslash}p{31pt}}
\resizebox{\textwidth}{!}{
\begin{tabular}{l | P P P P P P P P P P P P P P P P P P P P | M}
\toprule[0.15em]
Method & aero & bike & bird & boat & bottle & bus & car & cat & chair & cow & table & dog & horse & mbike & person & plant & sheep & sofa & train & tv & mAP \\
\midrule[0.05em]
WSDDN \cite{Bilen2016WSDDN} & 39.4 & 50.1 & 31.5 & 16.3 & 12.6 & 64.5 & 42.8 & 42.6 & 10.1 & 35.7 & 24.9 & 38.2 & 34.4 & 55.6 & 9.4  & 14.7 & 30.2 & 40.7 & 54.7 & 46.9 & 34.8 \\
ContextLocNet \cite{Kantorov2016ContextLocNet} & 57.1 & 52.0 & 31.5 & 7.6 & 11.5 & 55.0 & 53.1 & 34.1 & 1.7 & 33.1 & 49.2 & 42.0 & 47.3 & 56.6 & 15.3 & 12.8 & 24.8 & 48.9 & 44.4 & 47.8 & 36.3 \\
OICR \cite{Tang2017OICR} & 58.0 & 62.4 & 31.1 & 19.4 & 13.0 & 65.1 & 62.2 & 28.4 & 24.8 & 44.7 & 30.6 & 25.3 & 37.8 & 65.5 & 15.7 & 24.1 & 41.7 & 46.9 & 64.3 & 62.6 & 41.2 \\
WCCN \cite{Diba2017WCCN} & 49.5 & 60.6 & 38.6 & 29.2 & 16.2 & 70.8 & 56.9 & 42.5 & 10.9 & 44.1 & 29.9 & 42.2 & 47.9 & 64.1 & 13.8 & 23.5 & 45.9 & 54.1 & 60.8 & 54.5 & 42.8 \\
TS\textsuperscript{2}C \cite{Wei2018TS2C} & 59.3 & 57.5 & 43.7 & 27.3 & 13.5 & 63.9 & 61.7 & 59.9 & 24.1 & 46.9 & 36.7 & 45.6 & 39.9 & 62.6 & 10.3 & 23.6 & 41.7 & 52.4 & 58.7 & 56.6 & 44.3 \\
PCL \cite{Tang2018PCL} & 54.4 & 69.0 & 39.3 & 19.2 & 15.7 & 62.9 & 64.4 & 30.0 & 25.1 & 52.5 & 44.4 & 19.6 & 39.3 & 67.7 & 17.8 & 22.9 & 46.6 & 57.5 & 58.6 & 63.0 & 43.5 \\
WS-JDS \cite{Shen2019WS-JDS} & 52.0 & 64.5 & 45.5 & 26.7 & 27.9 & 60.5 & 47.8 & 59.7 & 13.0 & 50.4 & 46.4 & 56.3 & 49.6 & 60.7 & 25.4 & 28.2 & 50.0 & 51.4 & 66.5 & 29.7 & 45.6 \\
OAIL \cite{Kosugi2019OAIL} & 61.5 & 64.8 & 43.7 & 26.4 & 17.1 & 67.4 & 62.4 & 67.8 & 25.4 & 51.0 & 33.7 & 47.6 & 51.2 & 65.2 & 19.3 & 24.4 & 44.6 & 54.1 & 65.6 & 59.5 & 47.6 \\
SDCN \cite{Li2019SDCN} & 59.4 & 71.5 & 38.9 & 32.2 & 21.5 & 67.7 & 64.5 & 68.9 & 20.4 & 49.2 & 47.6 & 60.9 & 55.9 & 67.4 & 31.2 & 22.9 & 45.0 & 53.2 & 60.9 & 64.4 & 50.2 \\
C-MIDN \cite{Gao2019C-MIDN} & 53.3 & 71.5 & 49.8 & 26.1 & 20.3 & 70.3 & 69.9 & 68.3 & 28.7 & 65.3 & 45.1 & 64.6 & 58.0 & 71.2 & 20.0 & 27.5 & 54.9 & 54.9 & 69.4 & 63.5 & 52.6 \\
CSC \cite{Shen2019CSC} & 51.4 & 62.0 & 35.2 & 18.7 & 27.9 & 66.7 & 53.5 & 51.4 & 16.2 & 43.6 & 43.0 & 46.7 & 20.0 & 58.4 & 31.1 & 23.8 & 43.6 & 48.8 & 65.4 & 53.5 & 43.0 \\
PSLR \cite{Zhang2020PSLR} & 62.2 & 61.1 & 51.1 & 33.8 & 18.0 & 66.7 & 66.5 & 65.0 & 18.5 & 59.4 & 44.8 & 60.9 & 65.6 & 66.9 & 24.7 & 26.0 & 51.0 & 53.2 & 66.0 & 62.2 & 51.2 \\
P-MIDN+MGSC \cite{Xu2021P-MIDN+MGSC} & - & - & - & - & - & - & - & - & - & - & - & - & - & - & - & - & - & - & - & - & 53.9 \\
IM-CFB \cite{Yin2021IM-CFB} & 64.1 & 74.6 & 44.7 & 29.4 & 26.9 & 73.3 & 72.0 & 71.2 & 28.1 & \textbf{66.7} & 48.1 & 63.8 & 55.5 & 68.3 & 17.8 & 27.7 & 54.4 & 62.7 & 70.5 & 66.6 & 54.3 \\
D-MIL \cite{Gao2022D-MIL} & 60.4 & 71.3 & 51.1 & 25.4 & 23.8 & 70.4 & 70.3 & 71.9 & 25.2 & 63.4 & 42.6 & 67.1 & 57.7 & 70.1 & 15.5 & 26.6 & \textbf{58.7} & 63.3 & 66.9 & 67.6 & 53.5 \\
BUAA-PAL \cite{Wu2022BUAA-PAL} & \textbf{67.3} & \textbf{78.2} & \textbf{55.5} & 31.0 & 22.0 & 72.9 & \textbf{74.0} & 74.3 & 29.8 & 64.6 & 51.3 & 65.4 & 60.3 & 72.1 & 16.8 & 27.3 & 54.1 & \textbf{64.4} & 69.9 & 34.7 & 54.3 \\
MCC-MCT \cite{Wu2024MCC-MCT} & 59.2 & 69.5 & 54.8 & 27.9 & 28.8 & \textbf{73.4} & 73.1 & 60.4 & \textbf{31.1} & 63.0 & 51.1 & 50.5 & 52.6 & \textbf{72.3} & 17.0 & 29.3 & 49.4 & 56.1 & 69.1 & 65.8 & 52.7 \\
DANCE & 63.2 & 66.8 & 54.8 & \textbf{43.5} & \textbf{36.5} & 71.2 & 68.5 & \textbf{76.8} & 26.5 & 62.1 & \textbf{53.6} & \textbf{72.5} & \textbf{70.6} & 70.3 & \textbf{52.9} & \textbf{30.4} & 53.4 & 53.9 & \textbf{73.0} & \textbf{69.2} & \textbf{58.5} \\
\midrule[0.05em]
WeakSAM$^{\ddag}$ \cite{Zhu2024WeakSAM} & - & - & - & - & - & - & - & - & - & - & - & - & - & - & - & - & - & - & - & - & 58.9 \\
WS-SAM$^{\ddag}$ \cite{Wang2025WS-SAM} & 68.4 & \textbf{78.9} & 62.7 & 35.9 & 36.8 & \textbf{78.1} & \textbf{75.4} & 75.9 & \textbf{35.8} & \textbf{73.8} & 48.3 & 66.5 & 66.8 & \textbf{76.8} & 27.8 & 30.7 & \textbf{65.4} & \textbf{69.4} & 72.5 & 70.1 & 60.8 \\
DANCE (WeakSAM)$^{\ddag}$ & \textbf{72.0} & 71.8 & \textbf{64.4} & \textbf{51.1} & \textbf{47.9} & 77.1 & 73.8 & \textbf{77.4} & 28.3 & 70.7 & \textbf{55.2} & \textbf{76.8} & \textbf{75.7} & 76.6 & \textbf{63.2} & \textbf{31.9} & 59.3 & 51.9 & \textbf{73.9} & \textbf{70.8} & \textbf{63.5} \\
\midrule[0.05em]
\midrule[0.05em]
WSDDN-Ens. \cite{Bilen2016WSDDN} & 46.4 & 58.3 & 35.5 & 25.9 & 14.0 & 66.7 & 53.0 & 39.2 & 8.9 & 41.8 & 26.6 & 38.6 & 44.7 & 59.0 & 10.8 & 17.3 & 40.7 & 49.6 & 56.9 & 50.8 & 39.3 \\
OICR-Ens.+FRCNN \cite{Tang2017OICR} & 65.5 & 67.2 & 47.2 & 21.6 & 22.1 & 68.0 & 68.5 & 35.9 & 5.7 & 63.1 & 49.5 & 30.3 & 64.7 & 66.1 & 13.0 & 25.6 & 50.0 & 57.1 & 60.2 & 59.0 & 47.0 \\
W2F \cite{Zhang2018W2F} & 63.5 & 70.1 & 50.5 & 31.9 & 14.4 & 72.0 & 67.8 & 73.7 & 23.3 & 53.4 & 49.4 & 65.9 & 57.2 & 67.2 & 27.6 & 23.8 & 51.8 & 58.7 & 64.0 & 62.3 & 52.4 \\
PCL-Ens.+FRCNN \cite{Tang2018PCL} & 63.2 & 69.9 & 47.9 & 22.6 & 27.3 & 71.0 & 69.1 & 49.6 & 12.0 & 60.1 & 51.5 & 37.3 & 63.3 & 63.9 & 15.8 & 23.6 & 48.8 & 55.3 & 61.2 & 62.1 & 48.8 \\
WS-JDS+FRCNN \cite{Shen2019WS-JDS} & 64.8 & 70.7 & 51.5 & 25.1 & 29.0 & 74.1 & 69.7 & 69.6 & 12.7 & 69.5 & 43.9 & 54.9 & 39.3 & 71.3 & 32.6 & 29.8 & 57.0 & 61.0 & 66.6 & 57.4 & 52.5 \\
WSOD\textsuperscript{2} \cite{Zeng2019WSOD2} & 65.1 & 64.8 & 57.2 & 39.2 & 24.3 & 69.8 & 66.2 & 61.0 & 29.8 & 64.6 & 42.5 & 60.1 & 71.2 & 70.7 & 21.9 & 28.1 & 58.6 & 59.7 & 52.2 & 64.8 & 53.6 \\
TPEE \cite{Yang2019TPEE} & 57.6 & 70.8 & 50.7 & 28.3 & 27.2 & 72.5 & 69.1 & 65.0 & 26.9 & 64.5 & 47.4 & 47.7 & 53.5 & 66.9 & 13.7 & 29.3 & 56.0 & 54.9 & 63.4 & 65.2 & 51.5 \\
SDCN+FRCNN \cite{Li2019SDCN} & 59.8 & 75.1 & 43.3 & 31.7 & 22.8 & 69.1 & 71.0 & 72.9 & 21.0 & 61.1 & 53.9 & 73.1 & 54.1 & 68.3 & 37.6 & 20.1 & 48.2 & 62.3 & 67.2 & 61.1 & 53.7 \\
C-MIDN+FRCNN \cite{Gao2019C-MIDN} & 54.1 & 74.5 & 56.9 & 26.4 & 22.2 & 68.7 & 68.9 & 74.8 & 25.2 & 64.8 & 46.4 & 70.3 & 66.3 & 67.5 & 21.6 & 24.4 & 53.0 & 59.7 & 68.7 & 58.9 & 53.6 \\
CSC+FRCNN \cite{Shen2019CSC} & 58.4 & 63.3 & 48.1 & 21.7 & 29.6 & 66.7 & 66.3 & 66.1 & 9.3 & 61.1 & 40.5 & 49.5 & 35.9 & 64.9 & 39.2 & 26.4 & 53.2 & 55.6 & 70.2 & 54.0 & 49.0 \\
MIST \cite{Ren2020MIST} & \textbf{68.8} & 77.7 & 57.0 & 27.7 & 28.9 & 69.1 & 74.5 & 67.0 & 32.1 & \textbf{73.2} & 48.1 & 45.2 & 54.4 & 73.7 & 35.0 & 29.3 & \textbf{64.1} & 53.8 & 65.3 & 65.2 & 54.9 \\
SLV \cite{Chen2020SLV} & 65.6 & 71.4 & 49.0 & 37.1 & 24.6 & 69.6 & 70.3 & 70.6 & 30.8 & 63.1 & 36.0 & 61.4 & 65.3 & 68.4 & 12.4 & \textbf{29.9} & 52.4 & 60.0 & 67.6 & 64.5 & 53.5 \\
CASD \cite{Huang2020CASD} & - & - & - & - & - & - & - & - & - & - & - & - & - & - & - & - & - & - & - & - & 56.8 \\
PSLR+FRRCNN \cite{Zhang2020PSLR} & 62.3 & 63.1 & 53.5 & 42.1 & 19.0 & 64.8 & 68.2 & 71.0 & 17.2 & 64.3 & 56.0 & 72.6 & 67.9 & 64.1 & 20.8 & 23.0 & 50.3 & \textbf{69.5} & 65.8 & 59.5 & 53.8 \\
P-MIDN+MGSC+FRCNN \cite{Xu2021P-MIDN+MGSC} & - & - & - & - & - & - & - & - & - & - & - & - & - & - & - & - & - & - & - & - & 55.0 \\
IM-CFB+FRCNN \cite{Yin2021IM-CFB} & 63.3 & 77.5 & 48.3 & 36.0 & 32.6 & 70.8 & 71.9 & 73.1 & 29.1 & 68.7 & 47.1 & 69.4 & 56.6 & 70.9 & 22.8 & 24.8 & 56.0 & 59.8 & 73.2 & 64.6 & 55.8 \\
D-MIL+FRCNN \cite{Gao2022D-MIL} & 58.9 & 71.6 & 54.9 & 24.5 & 26.6 & 70.0 & 67.7 & 74.7 & 25.6 & 62.4 & 50.9 & 69.7 & 50.3 & 66.5 & 24.4 & 23.6 & 50.5 & 65.1 & 69.0 & 66.6 & 53.7 \\
NDI-WSOD \cite{Wang2022NDI-WSOD} & - & - & - & - & - & - & - & - & - & - & - & - & - & - & - & - & - & - & - & - & 56.8 \\
OD-WSCL \cite{Seo2022OD-WSCL} & 65.8 & 79.5 & \textbf{58.1} & 23.7 & 28.6 & 71.2 & \textbf{75.0} & 71.7 & 31.7 & 69.8 & 45.2 & 55.7 & 57.2 & \textbf{75.7} & 29.6 & 24.3 & 61.0 & 55.3 & 71.7 & \textbf{72.0} & 56.1 \\
CPNet \cite{Li2022CPNet} & 66.7 & 75.4 & 54.9 & 31.3 & 25.7 & 74.7 & 74.1 & 69.1 & 28.0 & 66.7 & 46.3 & 45.7 & 55.5 & 71.3 & 19.4 & 26.6 & 55.9 & 58.3 & 61.1 & 66.3 & 53.7 \\
BUAA-PAL+Reg \cite{Wu2022BUAA-PAL} & 66.1 & \textbf{80.1} & 41.3 & 30.1 & 28.5 & \textbf{75.3} & 72.0 & 76.2 & \textbf{33.5} & 69.7 & 48.6 & 62.1 & 60.2 & 73.1 & 16.7 & 26.8 & 54.1 & 60.4 & 70.8 & 63.3 & 55.4 \\
CBL \cite{Yin2023CBL} & - & - & - & - & - & - & - & - & - & - & - & - & - & - & - & - & - & - & - & - & 57.4 \\
MCC-MCT+Reg \cite{Wu2024MCC-MCT} & 59.3 & 73.0 & 46.8 & 41.2 & 25.9 & 74.7 & 76.7 & 79.0 & 33.2 & 74.3 & 51.3 & 60.6 & 62.9 & 72.2 & 12.8 & 26.3 & 56.3 & 61.1 & 73.2 & 37.0 & 54.9 \\
ICBC \cite{Yin2025ICBC} & - & - & - & - & - & - & - & - & - & - & - & - & - & - & - & - & - & - & - & - & 58.2 \\
DANCE+FRCNN & 66.3 & 68.7 & 54.3 & \textbf{46.0} & \textbf{37.5} & 73.3 & 69.0 & \textbf{78.1} & 25.8 & 67.1 & \textbf{56.1} & \textbf{77.2} & \textbf{73.0} & 71.1 & \textbf{54.7} & 28.2 & 52.2 & 54.9 & \textbf{74.8} & 69.3 & \textbf{59.9} \\
\bottomrule[0.15em]
\end{tabular}
}
\end{table*}

\begin{table*}[!t]
\caption{Comparison with the state-of-the-art methods on Pascal VOC 2007 trainval set in terms of CorLoc (\%)}
\label{tab:mCorLoc-VOC07}
\centering
\newcolumntype{P}{>{\centering\arraybackslash}p{21pt}}
\newcolumntype{M}{>{\centering\arraybackslash}p{31pt}}
\resizebox{\textwidth}{!}{
\begin{tabular}{l | P P P P P P P P P P P P P P P P P P P P | M}
\toprule[0.15em]
Method & aero & bike & bird & boat & bottle & bus & car & cat & chair & cow & table & dog & horse & mbike & person & plant & sheep & sofa & train & tv & mCorLoc \\
\midrule[0.05em]
WSDDN \cite{Bilen2016WSDDN} & 65.1 & 58.8 & 58.5 & 33.1 & 39.8 & 68.3 & 60.2 & 59.6 & 34.8 & 64.5 & 30.5 & 43.0 & 56.8 & 82.4 & 25.5 & 41.6 & 61.5 & 55.9 & 65.9 & 63.7 & 53.5 \\
ContextLocNet \cite{Kantorov2016ContextLocNet} & 83.3 & 68.6 & 54.7 & 23.4 & 18.3 & 73.6 & 74.1 & 54.1 & 8.6 & 65.1 & 47.1 & 59.5 & 67.0 & 83.5 & 35.3 & 39.9 & 67.0 & 49.7 & 63.5 & 65.2 & 55.1 \\
OICR \cite{Tang2017OICR} & 81.7 & 80.4 & 48.7 & 49.5 & 32.8 & 81.7 & 85.4 & 40.1 & 40.6 & 79.5 & 35.7 & 33.7 & 60.5 & 88.8 & 21.8 & 57.9 & 76.3 & 59.9 & 75.3 & 81.4 & 60.6 \\
WCCN \cite{Diba2017WCCN} & 83.9 & 72.8 & 64.5 & 44.1 & 40.1 & 65.7 & 82.5 & 58.9 & 33.7 & 72.5 & 25.6 & 53.7 & 67.4 & 77.4 & 26.8 & 49.1 & 68.1 & 27.9 & 64.5 & 55.7 & 56.7 \\
TS\textsuperscript{2}C \cite{Wei2018TS2C} & 84.2 & 74.1 & 61.3 & 52.1 & 32.1 & 76.7 & 82.9 & 66.6 & 42.3 & 70.6 & 39.5 & 57.0 & 61.2 & 88.4 & 9.3 & 54.6 & 72.2 & 60.0 & 65.0 & 70.3 & 61.0 \\
PCL \cite{Tang2018PCL} & 79.6 & 85.5 & 62.2 & 47.9 & 37.0 & 83.8 & 83.4 & 43.0 & 38.3 & 80.1 & 50.6 & 30.9 & 57.8 & 90.8 & 27.0 & 58.2 & 75.3 & 68.5 & 75.7 & 78.9 & 62.7 \\
WS-JDS \cite{Shen2019WS-JDS} & 82.9 & 74.0 & 73.4 & 47.1 & 60.9 & 80.4 & 77.5 & 78.8 & 18.6 & 70.0 & 56.7 & 67.0 & 64.5 & 84.0 & 47.0 & 50.1 & 71.9 & 57.6 & 83.3 & 43.5 & 64.5 \\
OAIL \cite{Kosugi2019OAIL} & 85.5 & 79.6 & 68.1 & 55.1 & 33.6 & 83.5 & 83.1 & 78.5 & 42.7 & 79.8 & 37.8 & 61.5 & 74.4 & 88.6 & 32.6 & 55.7 & \textbf{77.9} & 63.7 & 78.4 & 74.1 & 66.7 \\
SDCN \cite{Li2019SDCN} & 85.0 & 83.9 & 58.9 & 59.6 & 43.1 & 79.7 & 85.2 & 77.9 & 31.3 & 78.1 & 50.6 & 75.6 & 76.2 & 88.4 & 49.7 & 56.4 & 73.2 & 62.6 & 77.2 & 79.9 & 68.6 \\
C-MIDN \cite{Gao2019C-MIDN} & - & - & - & - & - & - & - & - & - & - & - & - & - & - & - & - & - & - & - & - & 68.7 \\
CSC \cite{Shen2019CSC} & 76.1 & 75.3 & 61.8 & 42.0 & 54.1 & 74.7 & 78.8 & 67.4 & 32.8 & 73.1 & 46.5 & 59.9 & 37.6 & 78.0 & 56.0 & 42.5 & 71.9 & 67.3 & 82.4 & 65.6 & 62.2 \\
PSLR \cite{Zhang2020PSLR} & 86.3 & 72.9 & 71.2 & 59.0 & 36.3 & 80.2 & 84.4 & 75.6 & 30.8 & 83.6 & 53.2 & 75.1 & 82.7 & 87.1 & 37.7 & 54.6 & 74.2 & 59.1 & 79.8 & 78.9 & 68.1 \\
P-MIDN+MGSC \cite{Xu2021P-MIDN+MGSC} & - & - & - & - & - & - & - & - & - & - & - & - & - & - & - & - & - & - & - & - & 69.8 \\
IM-CFB \cite{Yin2021IM-CFB} & - & - & - & - & - & - & - & - & - & - & - & - & - & - & - & - & - & - & - & - & 70.7 \\
D-MIL \cite{Gao2022D-MIL} & 81.3 & 82.0 & 72.7 & 48.9 & 42.0 & 80.2 & 86.1 & 78.5 & 43.9 & 80.2 & 42.2 & 76.5 & 68.7 & 91.2 & 32.7 & 56.0 & \textbf{81.4} & 69.6 & 78.7 & 79.9 & 68.7 \\
BUAA-PAL \cite{Wu2022BUAA-PAL} & 84.2 & \textbf{86.7} & 71.5 & 52.1 & 38.2 & 83.8 & 87.8 & 84.3 & 47.7 & 80.8 & 50.6 & 76.3 & 79.3 & \textbf{94.0} & 29.5 & \textbf{61.2} & 77.3 & 70.7 & 82.5 & 60.9 & 70.0 \\
MCC-MCT \cite{Wu2024MCC-MCT} & 75.3 & 80.0 & 69.0 & 50.4 & 49.3 & 82.2 & 89.8 & 63.0 & 45.5 & \textbf{86.8} & 54.1 & 59.1 & 73.4 & 92.0 & 20.1 & 57.0 & 80.1 & 59.0 & 82.6 & 79.5 & 67.4 \\
DANCE & \textbf{90.3} & 79.8 & \textbf{86.1} & \textbf{77.9} & \textbf{70.5} & \textbf{88.7} & \textbf{90.5} & \textbf{93.5} & \textbf{50.3} & 83.7 & \textbf{73.0} & \textbf{89.8} & \textbf{89.9} & 91.8 & \textbf{81.3} & 57.1 & 77.1 & \textbf{79.5} & \textbf{95.4} & \textbf{89.8} & \textbf{81.8} \\
\midrule[0.05em]
WeakSAM$^{\ddag}$ \cite{Zhu2024WeakSAM} & - & - & - & - & - & - & - & - & - & - & - & - & - & - & - & - & - & - & - & - & 74.5 \\
WS-SAM$^{\ddag}$ \cite{Wang2025WS-SAM} & - & - & - & - & - & - & - & - & - & - & - & - & - & - & - & - & - & - & - & - & 75.1 \\
DANCE (WeakSAM)$^{\ddag}$ & \textbf{92.9} & \textbf{81.1} & \textbf{86.1} & \textbf{82.9} & \textbf{75.0} & \textbf{90.9} & \textbf{89.9} & \textbf{95.0} & \textbf{53.3} & \textbf{87.2} & \textbf{75.0} & \textbf{91.9} & \textbf{92.0} & \textbf{94.3} & \textbf{85.8} & \textbf{58.0} & \textbf{85.4} & \textbf{74.7} & \textbf{96.6} & \textbf{91.0} & \textbf{83.9} \\
\midrule[0.05em]
\midrule[0.05em]
WSDDN-Ens. \cite{Bilen2016WSDDN} & 68.9 & 68.7 & 65.2 & 42.5 & 40.6 & 72.6 & 75.2 & 53.7 & 29.7 & 68.1 & 33.5 & 45.6 & 65.9 & 86.1 & 27.5 & 44.9 & 76.0 & 62.4 & 66.3 & 66.8 & 58.0 \\
OICR-Ens.+FRCNN \cite{Tang2017OICR} & 85.8 & 82.7 & 62.8 & 45.2 & 43.5 & 84.8 & 87.0 & 46.8 & 15.7 & 82.2 & 51.0 & 45.6 & 83.7 & 91.2 & 22.2 & 59.7 & 75.3 & 65.1 & 76.8 & 78.1 & 64.3 \\
W2F \cite{Zhang2018W2F} & 85.4 & 87.5 & 62.5 & 54.3 & 35.5 & 85.3 & 86.6 & 82.3 & 39.7 & 82.9 & 49.4 & 76.5 & 74.8 & 90.0 & 46.8 & 53.9 & \textbf{84.5} & 68.3 & 79.1 & 79.9 & 70.3 \\
PCL-Ens.+FRCNN \cite{Tang2018PCL} & 83.8 & 85.1 & 65.5 & 43.1 & 50.8 & 83.2 & 85.3 & 59.3 & 28.5 & 82.2 & 57.4 & 50.7 & 85.0 & 92.0 & 27.9 & 54.2 & 72.2 & 65.9 & 77.6 & 82.1 & 66.6 \\
WS-JDS+FRCNN \cite{Shen2019WS-JDS} & 79.8 & 84.0 & 68.3 & 40.2 & 61.5 & 80.5 & 85.8 & 75.8 & 29.7 & 77.7 & 49.5 & 67.4 & 58.6 & 87.4 & 66.2 & 46.6 & 78.5 & 73.7 & 84.5 & 72.8 & 68.6 \\
WSOD\textsuperscript{2} \cite{Zeng2019WSOD2} & 87.1 & 80.0 & 74.8 & 60.1 & 36.6 & 79.2 & 83.8 & 70.6 & 43.5 & \textbf{88.4} & 46.0 & 74.7 & 87.4 & 90.8 & 44.2 & 52.4 & 81.4 & 61.8 & 67.7 & 79.9 & 69.5 \\
TPEE \cite{Yang2019TPEE} & 80.0 & 83.9 & 74.2 & 53.2 & 48.5 & 82.7 & 86.2 & 69.5 & 39.3 & 82.9 & 53.6 & 61.4 & 72.4 & 91.2 & 22.4 & 57.5 & 83.5 & 64.8 & 75.7 & 77.1 & 68.0 \\
SDCN+FRCNN \cite{Li2019SDCN} & 85.0 & 86.7 & 60.7 & 62.8 & 46.6 & 83.2 & 87.8 & 81.7 & 35.8 & 80.8 & 57.4 & 81.6 & 79.9 & 92.4 & 59.3 & 57.5 & 79.4 & 68.5 & 81.7 & 81.4 & 72.5 \\
C-MIDN+FRCNN \cite{Gao2019C-MIDN} & - & - & - & - & - & - & - & - & - & - & - & - & - & - & - & - & - & - & - & - & 71.9 \\
CSC+FRCNN \cite{Shen2019CSC} & 77.3 & 81.5 & 65.8 & 38.7 & 59.0 & 78.0 & 83.3 & 73.3 & 27.2 & 75.2 & 47.0 & 64.9 & 56.1 & 86.9 & 63.7 & 44.1 & 76.0 & 71.2 & 82.0 & 70.3 & 66.1 \\
MIST \cite{Ren2020MIST} & 87.5 & 82.4 & 76.0 & 58.0 & 44.7 & 82.2 & 87.5 & 71.2 & 49.1 & 81.5 & 51.7 & 53.3 & 71.4 & 92.8 & 38.2 & 52.8 & 79.4 & 61.0 & 78.3 & 76.0 & 68.8 \\
SLV \cite{Chen2020SLV} & 84.6 & 84.3 & 73.3 & 58.5 & 49.2 & 80.2 & 87.0 & 79.4 & 46.8 & 83.6 & 41.8 & 79.3 & 88.8 & 90.4 & 19.5 & 59.7 & 79.4 & 67.7 & 82.9 & 83.2 & 71.0 \\
CASD \cite{Huang2020CASD} & - & - & - & - & - & - & - & - & - & - & - & - & - & - & - & - & - & - & - & - & 70.4 \\
PSLR+FRRCNN \cite{Zhang2020PSLR} & 87.9 & 75.7 & 72.7 & 63.3 & 47.7 & 86.3 & 88.2 & 79.4 & \textbf{50.4} & 84.9 & 67.7 & 80.2 & 86.1 & 92.4 & 40.8 & \textbf{64.5} & 80.4 & \textbf{83.1} & 79.9 & 84.6 & 74.8 \\
P-MIDN+MGSC+FRCNN \cite{Xu2021P-MIDN+MGSC} & - & - & - & - & - & - & - & - & - & - & - & - & - & - & - & - & - & - & - & - & 72.4 \\
IM-CFB+FRCNN \cite{Yin2021IM-CFB} & - & - & - & - & - & - & - & - & - & - & - & - & - & - & - & - & - & - & - & - & 72.2 \\
D-MIL+FRCNN \cite{Gao2022D-MIL} & 83.3 & 83.9 & 74.8 & 50.0 & 46.6 & 82.7 & 86.5 & 83.7 & 44.4 & 83.6 & 48.3 & 79.5 & 71.4 & 93.2 & 39.0 & 59.0 & 80.4 & 70.7 & 81.4 & 81.0 & 71.2 \\
NDI-WSOD \cite{Wang2022NDI-WSOD} & - & - & - & - & - & - & - & - & - & - & - & - & - & - & - & - & - & - & - & - & 71.0 \\
OD-WSCL \cite{Seo2022OD-WSCL} & 86.3 & 87.8 & 74.5 & 47.3 & 43.9 & 85.8 & 84.6 & 78.2 & 49.1 & 83.6 & 49.4 & 61.6 & 74.5 & 92.4 & 42.2 & 46.9 & 80.4 & 62.1 & 82.9 & 82.8 & 69.8 \\
CPNet \cite{Li2022CPNet} & 87.8 & 87.7 & 74.4 & 50.3 & 54.6 & 81.6 & \textbf{91.2} & 79.9 & 47.4 & 80.8 & 51.0 & 56.4 & 71.6 & 90.2 & 30.2 & 57.5 & 77.1 & 71.6 & 77.0 & 85.2 & 70.2 \\
BUAA-PAL+Reg \cite{Wu2022BUAA-PAL} & 84.8 & \textbf{90.4} & 61.7 & 54.2 & 50.2 & 87.8 & 86.0 & 84.1 & 50.0 & 85.9 & 49.7 & 73.1 & 80.6 & \textbf{95.0} & 28.1 & 62.9 & 76.3 & 60.7 & 83.9 & 79.1 & 71.2 \\
CBL \cite{Yin2023CBL} & - & - & - & - & - & - & - & - & - & - & - & - & - & - & - & - & - & - & - & - & 71.8 \\
MCC-MCT+Reg \cite{Wu2024MCC-MCT} & 82.5 & 89.8 & 65.5 & 58.5 & 39.3 & 84.8 & 85.0 & 84.6 & 48.6 & 85.9 & 52.0 & 79.8 & 84.1 & 94.0 & 20.0 & 60.1 & 82.5 & 69.9 & 82.5 & 56.6 & 70.3 \\
ICBC \cite{Yin2025ICBC} & - & - & - & - & - & - & - & - & - & - & - & - & - & - & - & - & - & - & - & - & 71.9 \\
DANCE+FRCNN & \textbf{91.6} & 81.1 & \textbf{82.4} & \textbf{77.9} & \textbf{72.5} & \textbf{89.2} & 90.0 & \textbf{93.8} & 49.4 & 86.5 & \textbf{69.0} & \textbf{90.7} & \textbf{91.6} & 92.2 & \textbf{83.2} & 56.3 & 78.1 & 78.2 & \textbf{95.0} & \textbf{89.8} & \textbf{81.9} \\
\bottomrule[0.15em]
\end{tabular}
}
\end{table*}

\begin{figure*}[!t]
\centering
\includegraphics[width=\textwidth]{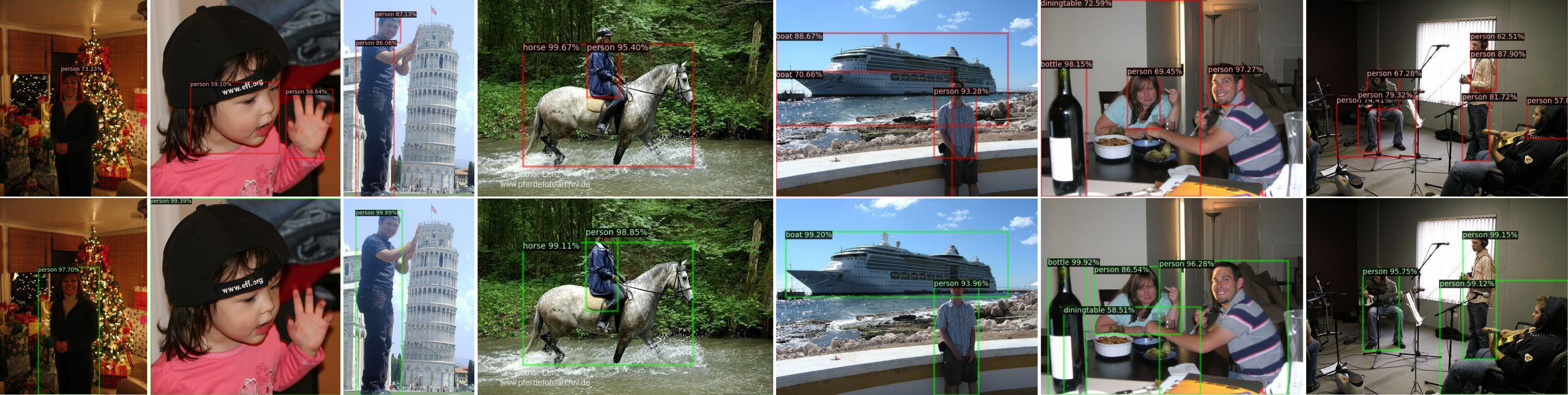}
\caption{Visual comparison results of OICR (top row) and DANCE (bottom row) on ``person'' category. Due to the complex shapes and visual patterns associated with the ``person'' category, OICR tends to detect only parts of the person, such as the head, the hand, the upper body, or the torso. In contrast, our method detects more accurate bounding boxes because the heatmaps provide better coverage of the entire person.}
\label{fig:DANCE-vs-OICR}
\end{figure*}

\begin{table}[!t]
\caption{Comparison with the state-of-the-art methods on Pascal VOC 2012 dataset}
\label{tab:VOC12}
\centering
\newcolumntype{M}{>{\centering\arraybackslash}p{31pt}}
\begin{threeparttable}
\begin{tabular}{l | M | M}
\toprule[0.15em]
Method & mAP & mCorLoc \\
\midrule[0.05em]
ContextLocNet \cite{Kantorov2016ContextLocNet} & 35.3 & 54.8 \\
OICR \cite{Tang2017OICR} & 37.9 & 62.1 \\
WCCN \cite{Diba2017WCCN} & 37.9 & - \\
TS\textsuperscript{2}C \cite{Wei2018TS2C} & 40.0 & 64.4 \\
PCL \cite{Tang2018PCL} & 40.6 & 63.2 \\
WS-JDS \cite{Shen2019WS-JDS} & 39.1 & 63.5 \\
OAIL \cite{Kosugi2019OAIL} & 43.4 & 66.7 \\
SDCN \cite{Li2019SDCN} & 43.5 & 67.9 \\
C-MIDN \cite{Gao2019C-MIDN} & 50.2 & 71.2 \\
CSC \cite{Shen2019CSC} & 37.1 & 61.4 \\
PSLR \cite{Zhang2020PSLR} & 46.3 & 68.7 \\
P-MIDN+MGSC \cite{Xu2021P-MIDN+MGSC} & 52.8 & 73.3 \\
IM-CFB \cite{Yin2021IM-CFB} & 49.4 & 69.6 \\
D-MIL \cite{Gao2022D-MIL} & 49.6 & 70.1 \\
BUAA-PAL \cite{Wu2022BUAA-PAL} & 51.2 & 72.4 \\
MCC-MCT \cite{Wu2024MCC-MCT} & 49.3 & 69.3 \\
DANCE & \textbf{55.6}\tnote{1} & \textbf{80.5} \\
\midrule[0.05em]
\midrule[0.05em]
OICR-Ens.+FRCNN \cite{Tang2017OICR} & 42.5 & 65.6 \\
W2F \cite{Zhang2018W2F} & 47.8 & 69.4 \\
PCL-Ens.+FRCNN \cite{Tang2018PCL} & 44.2 & 68.0 \\
WS-JDS+FRCNN \cite{Shen2019WS-JDS} & 46.1 & 69.5 \\
WSOD\textsuperscript{2} \cite{Zeng2019WSOD2} & 47.2 & 71.9 \\
TPEE \cite{Yang2019TPEE} & 45.6 & 68.7 \\
SDCN+FRCNN \cite{Li2019SDCN} & 46.7 & 69.5 \\
C-MIDN+FRCNN \cite{Gao2019C-MIDN} & 50.3 & 73.3 \\
CSC+FRCNN \cite{Shen2019CSC} & 44.1 & 67.0 \\
MIST \cite{Ren2020MIST} & 52.1 & 70.9 \\
SLV \cite{Chen2020SLV} & 49.2 & 69.2 \\
CASD \cite{Huang2020CASD} & 53.6 & 72.3 \\
PSLR+FRRCNN \cite{Zhang2020PSLR} & 49.7 & 74.5 \\
P-MIDN+MGSC+FRCNN \cite{Xu2021P-MIDN+MGSC} & 53.4 & 76.0 \\
D-MIL+FRCNN \cite{Gao2022D-MIL} & 49.8 & 71.9 \\
NDI-WSOD \cite{Wang2022NDI-WSOD} & 53.9 & 72.2 \\
OD-WSCL \cite{Seo2022OD-WSCL} & 54.6 & 71.2 \\
CPNet \cite{Li2022CPNet} & 50.2 & - \\
BUAA-PAL+Reg \cite{Wu2022BUAA-PAL} & 53.0 & 74.0 \\
CBL \cite{Yin2023CBL} & 53.5 & 72.6 \\
MCC-MCT+Reg \cite{Wu2024MCC-MCT} & 51.1 & 72.3 \\
ICBC \cite{Yin2025ICBC} & 55.5 & 73.4 \\
DANCE+FRCNN & \textbf{56.8}\tnote{2} & \textbf{81.4} \\
\bottomrule[0.15em]
\end{tabular}
\begin{tablenotes}
    \footnotesize
    \item[1] http://host.robots.ox.ac.uk:8080/anonymous/CS4U99.html
    \item[2] http://host.robots.ox.ac.uk:8080/anonymous/VGH65Q.html
\end{tablenotes}
\end{threeparttable}
\end{table}

\begin{table}[!t]
\caption{Comparison with the state-of-the-art methods on MS COCO 2014 val set in terms of mAP (\%)}
\label{tab:AP-COCO14}
\centering
\newcolumntype{M}{>{\centering\arraybackslash}p{20pt}}
\resizebox{\columnwidth}{!}{
\begin{tabular}{l | c M M | M M M}
\toprule[0.15em]
Method & mAP\textsubscript{[50:95]} & mAP\textsubscript{50} & mAP\textsubscript{75} & mAP\textsubscript{\textit{s}} & mAP\textsubscript{\textit{m}} & mAP\textsubscript{\textit{l}} \\
\midrule[0.05em]
PCL \cite{Tang2018PCL} & 8.5 & 19.4 & - & - & - & - \\
C-MIDN \cite{Gao2019C-MIDN} & 9.6 & 21.4 & - & - & - & - \\
CSC \cite{Shen2019CSC} & 10.2 & 20.3 & 8.9 & 2.3 & 10.9 & 18.5 \\
PSLR \cite{Zhang2020PSLR} & 11.1 & 23.6 & - & - & - & - \\
P-MIDN+MGSC \cite{Xu2021P-MIDN+MGSC} & 13.1 & 27.4 & - & - & - & - \\
D-MIL \cite{Gao2022D-MIL} & 11.3 & 24.7 & - & - & - & - \\
BUAA-PAL \cite{Wu2022BUAA-PAL} & 11.9 & 25.6 & - & - & - & - \\
MCC-MCT \cite{Wu2024MCC-MCT} & 11.2 & 24.2 & - & - & - & - \\
DANCE & \textbf{14.6} & \textbf{28.7} & \textbf{13.1} & \textbf{4.5} & \textbf{14.6} & \textbf{24.9} \\
\midrule[0.05em]
\midrule[0.05em]
PCL-Ens.+FRCNN \cite{Tang2018PCL} & 9.2 & 19.6 & - & - & - & - \\
WSOD\textsuperscript{2} \cite{Zeng2019WSOD2} & 10.8 & 22.7 & - & - & - & - \\
MIST \cite{Ren2020MIST} & 11.4 & 24.3 & 9.4 & 3.6 & 12.2 & 17.6 \\
CASD \cite{Huang2020CASD} & 12.8 & 26.4 & - & - & - & - \\
P-MIDN+MGSC+FRCNN \cite{Xu2021P-MIDN+MGSC} & 13.2 & 28.5 & - & - & - & - \\
NDI-WSOD \cite{Wang2022NDI-WSOD} & 12.1 & 26.2 & - & 3.7 & 13.2 & 19.3 \\
OD-WSCL \cite{Seo2022OD-WSCL} & 13.7 & 27.7 & 11.9 & 4.4 & 14.5 & 21.2 \\
CPNet \cite{Li2022CPNet} & 11.4 & 24.2 & - & - & - & - \\
BUAA-PAL+Reg \cite{Wu2022BUAA-PAL} & 12.4 & 26.3 & - & - & - & - \\
CBL \cite{Yin2023CBL} & 13.6 & 27.6 & - & - & - & - \\
MCC-MCT+Reg \cite{Wu2024MCC-MCT} & 12.1 & 25.5 & - & - & - & - \\
DANCE+FRCNN & 15.4 & 30.1 & \textbf{14.7} & \textbf{5.2} & \textbf{16.4} & \textbf{25.8} \\
\bottomrule[0.15em]
\end{tabular}
}
\end{table}

We compare our method with a selection of the most representative and influential WSOD approaches. The results on Pascal VOC are presented in Tables \ref{tab:mAP-VOC07}, \ref{tab:mCorLoc-VOC07} and \ref{tab:VOC12} (with detailed per-class results for VOC 2012 provided in Tables \ref{tab:mAP-VOC12}, and \ref{tab:mCorLoc-VOC12} of Appendix \ref{app:D}), while the results on MS COCO are shown in Table \ref{tab:AP-COCO14}. In these tables, $^\ddag$ means using SAM \cite{Kirillov2023SAM} instead of traditional ways \cite{Uijlings2013SS, Zitnick2014EB, Arbelaez2014MCG} for proposal pre-generation. The upper half of these tables all represent methods with only classification capability, which means models having only bounding box classification branch, while the lower half all represent methods possessing regression capability, which means models either natively equipping with bounding box regression branch or adding a complete Fast R-CNN \cite{Girshick2015FRCNN} or Faster R-CNN \cite{Ren2015FRRCNN} at the end.

As observed in Tables \ref{tab:mAP-VOC07}-\ref{tab:VOC12}, our method achieves the highest performance in fair comparisons on Pascal VOC, whether incorporating Fast R-CNN or not. When using traditional proposals \cite{Uijlings2013SS, Zitnick2014EB, Arbelaez2014MCG}, DANCE outperforms SOTA by 4.2\% in mAP and 11.1\% in mCorLoc, and DANCE+FRCNN outperforms SOTA by 2.5\% in mAP and 7.1\% in mCorLoc on the Pascal VOC 2007 dataset. Performances on Pascal VOC 2012 also surpass all SOTA methods. We also conduct a fair comparison with the recent work WeakSAM \cite{Zhu2024WeakSAM} by using the same region proposals pre-generated by SAM \cite{Kirillov2023SAM} as introduced in their work. DANCE also outperforms WeakSAM by 4.6\% in mAP and 9.4\% in mCorLoc.

Moreover, our method represents highly competitive results in the ``person'' category, where prior approaches have historically performed poorly. Our framework achieves 52.9\%/54.7\% mAP upon person class without/with Fast R-CNN on Pascal VOC 2007 dataset, outperforming prior arts by +21.7\%/+15.5\%, which marks a significant leap forward. As shown in Figure \ref{fig:DANCE-vs-OICR}, divergent upper/lower body apparel colors induce textural partitions that segment humans into disconnected local regions. This phenomenon causes previous methods to predominantly detect isolated body parts (e.g., head, torso, hand, etc.) while missing full-body extents. Our approach overcomes this limitation through category-specific heatmaps that preserve global anatomical context, producing significantly more complete bounding boxes that contain the whole person --- consequently boosting detection precision.

Our method also achieves the highest performance on the MS COCO dataset. As shown in Table \ref{tab:AP-COCO14}, DANCE outperforms the previous SOTA by 1.5\% and 1.3\% in terms of the comprehensive mAP\textsubscript{[50:95]} and the single-threshold mAP\textsubscript{50} respectively. Furthermore, when cascaded with a Fast R-CNN at the end of DANCE to equip the model with bounding box regression capabilities, our approach surpasses the SOTA by 1.7\% and 1.6\% under a fair comparison.

Additionally, we report the detection results for small, medium and large objects respectively in Table \ref{tab:AP-COCO14}. Under a fair comparison, DANCE outperforms CSC\cite{Shen2019CSC} by 2.2\%, 3.7\% and 6.4\% in mAP\textsubscript{\textit{s}}, mAP\textsubscript{\textit{m}} and mAP\textsubscript{\textit{l}} respectively; DANCE+FRCNN also surpasses OD-WSCL\cite{Seo2022OD-WSCL} by 0.8\%, 1.9\% and 4.6\% respectively. These results show that DANCE consistently exceeds the SOTA methods across these fine-grained evaluation metrics.

\subsection{Ablation Studies}

We conduct a series of ablation studies on the Pascal VOC 2007 dataset to validate the effectiveness of our proposed method. First, we demonstrate the significance of the dual-threshold design and show the robustness of the key hyperparameters in HGPS. Then, we evaluate the performance of HGPS across various heatmap generation methods and provide both qualitative and quantitative analyzes. Subsequently, we discuss the effect of each component within WSBDN. We next compare the performance of the two base models WSBDN and WSDDN \cite{Bilen2016WSDDN}, and compare our HGPS algorithm with other selection algorithms such as OICR \cite{Tang2017OICR} to verify each design's efficacy. Furthermore, we report the acceleration of the classification-ignored loss on network convergence. Lastly, we show the efficacy of each innovation of DANCE by quantifying their respective contributions to the performance gains. More ablation studies are stated in Appendix \ref{app:E}.

\subsubsection{Hyperparameters in HGPS}

\begin{figure}[!t]
\centering
\includegraphics[width=0.85\columnwidth]{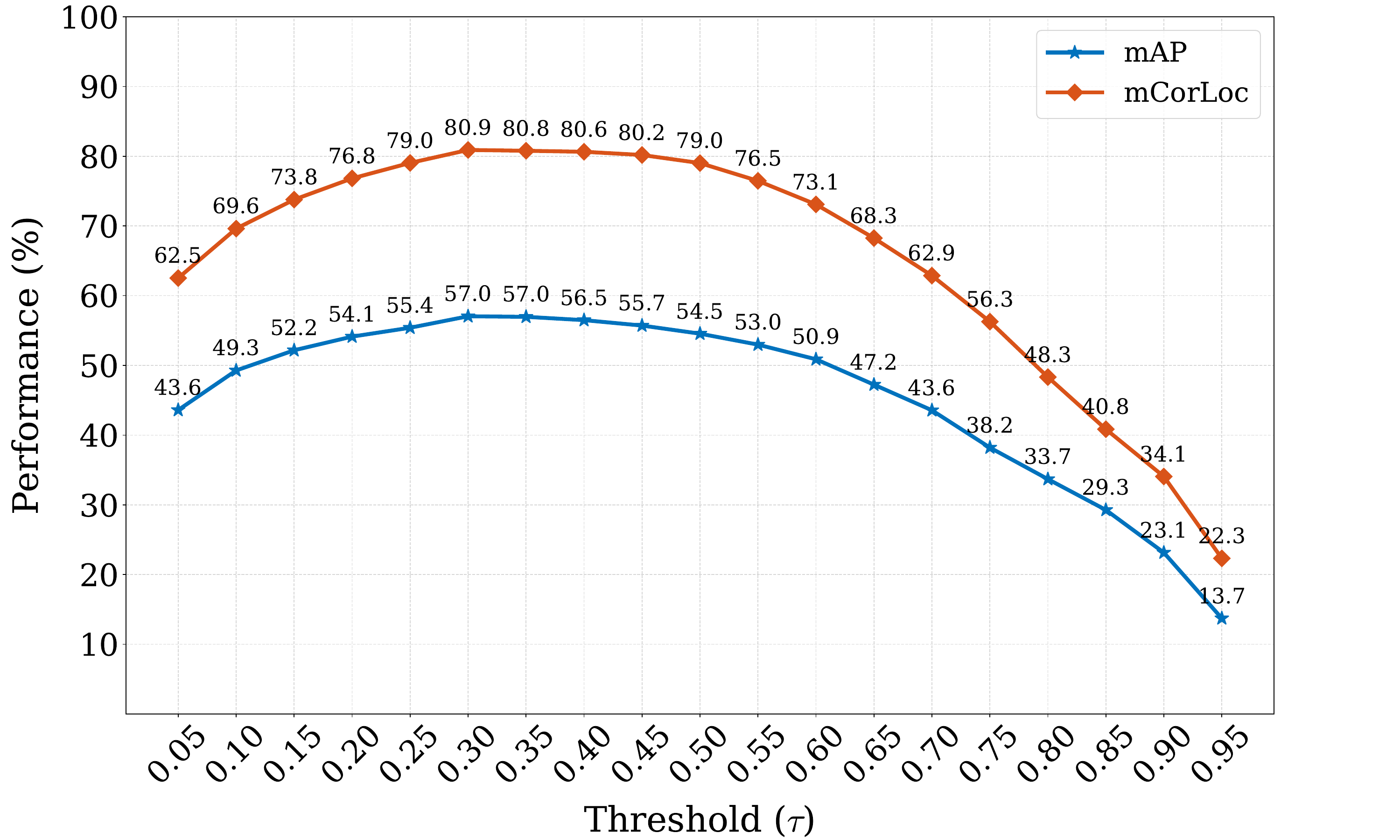}
\vspace{-0.5em}
\caption{Performance of single threshold on Pascal VOC 2007. It demonstrates effectiveness of our dual-threshold design.}
\label{fig:single-threshold}
\end{figure}

Figure \ref{fig:single-threshold} shows the performance of using only a single threshold and directly treating each threshold box as a pseudo GT box. The results exhibit a distinct peak-like curve, and the performance peaks at 0.3, achieving 57.0\% mAP and 80.9\% mCorLoc. Notably, the peak performance is still lower than that of our dual-threshold method. When adding another threshold, the results can outperform by +1.5\% mAP and +0.8\% mCorLoc. This demonstrates that our dual-threshold strategy is a helpful and well-motivated approach that addresses a fundamental challenge in weakly supervised object detection.

\begin{figure}[!t]
\centering
\includegraphics[width=\columnwidth]{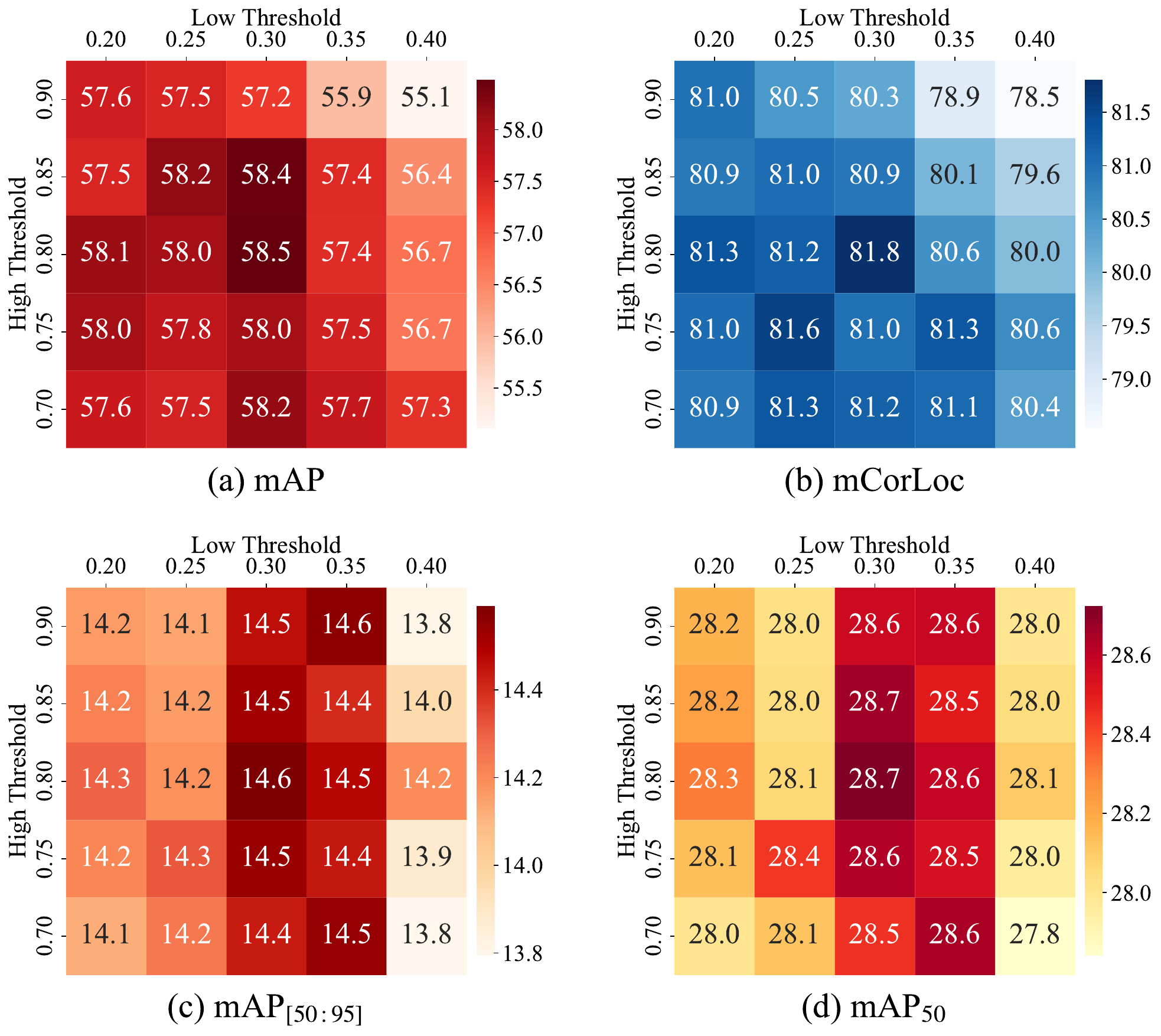}
\vspace{-1.8em}
\caption{Ablation results for different high and low thresholds. Panels (a) and (b) show the experimental results on VOC07, while panels (c) and (d) show the experimental results on COCO14. The results show the robustness of our method to variations in the two thresholds.}
\label{fig:dual-thresholds}
\end{figure}

Next, the impact of the high and low thresholds is illustrated in Figure \ref{fig:dual-thresholds}. It is observed that our method achieves its peak performance when $\tau^{\mathrm{low}} = 0.3$ and $\tau^{\mathrm{high}} = 0.8$ on both the VOC and COCO datasets. This indicates that the high- and low-threshold hyperparameters used in our method generalize well across domains. As the thresholds deviate from this central point, the overall performance tends to decrease, further demonstrating the optimality of our hyperparameter choices. It is also worth noting that even when the optimal hyperparameters are not selected, as $\tau^{\mathrm{low}}$ varies within the range of 0.2--0.4 and $\tau^{\mathrm{high}}$ varies within the range of 0.7--0.9, the worst performance can still reach 55.1\% mAP and 78.5\% mCorLoc on VOC 2007, as well as 13.8\% mAP\textsubscript{[50:95]} and 27.8\% mAP\textsubscript{50} on COCO 2014. This sufficiently indicates that our method is not overly sensitive to the choice of thresholds, thereby demonstrating its strong robustness.

\begin{figure}[!t]
\centering
\includegraphics[width=0.75\columnwidth]{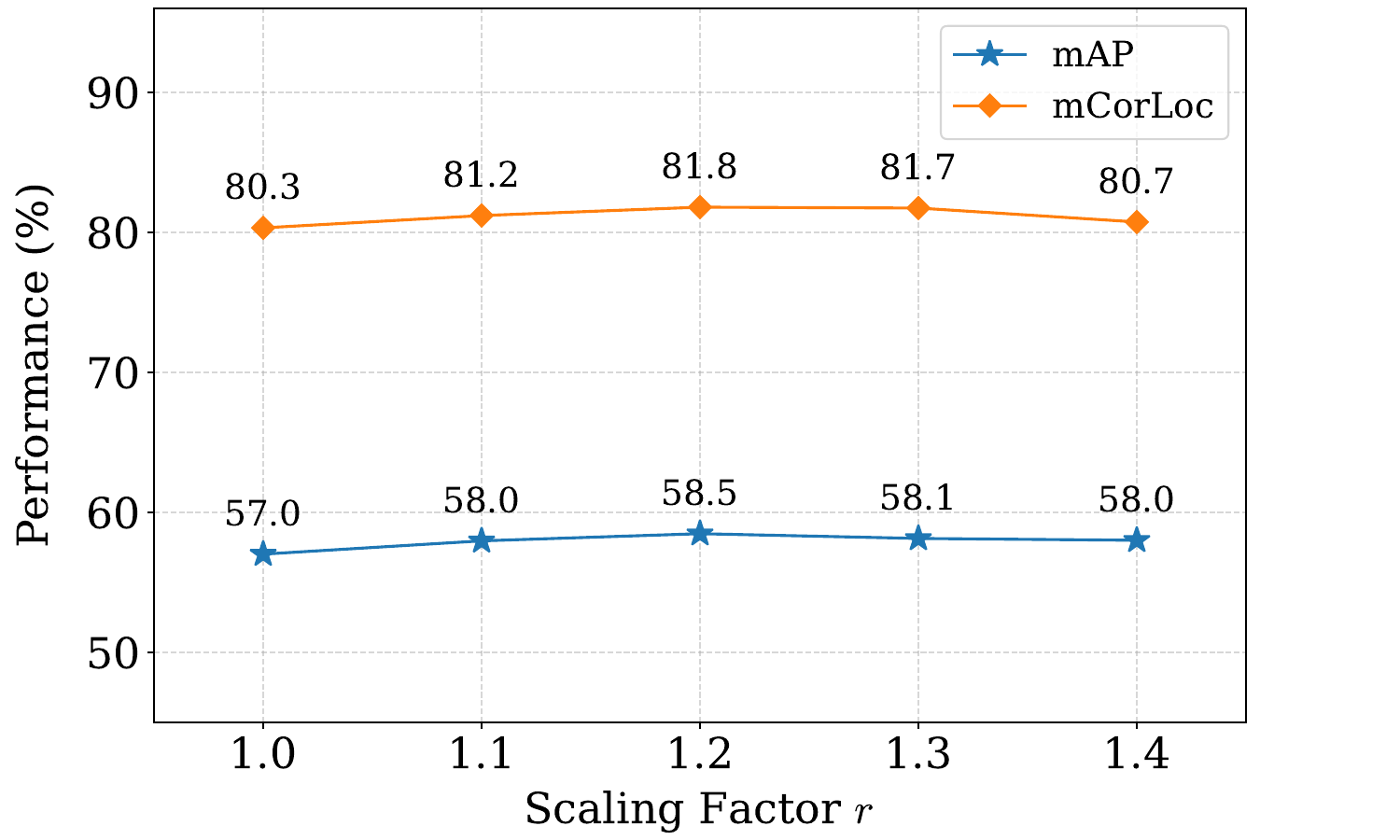}
\vspace{-0.5em}
\caption{Effect of different box scaling factors on Pascal VOC 2007. It performs robustness to variations in the box scaling factor.}
\label{fig:scale-factor}
\end{figure}

We then present the impact of the scaling factor $r$ in Figure \ref{fig:scale-factor}. The performance peaks at $r=1.2$, and any deviation from this value, whether an increase or decrease, results in a performance drop. At $r = 1.0$, the model achieves its lowest performance with 57.0\% mAP and 80.3\% mCorLoc. This result underscores the necessity of the box ``relaxation'' strategy within our proposed HGPS algorithm. However, when $r \geq 1.1$, the mAP consistently remains over 58.0\% and fluctuates by no more than 0.5 percentage points from the peak value. Our method is not highly sensitive to the scaling factor, showcasing its strong robustness.

\subsubsection{Effect of Different Heatmap Generation Methods}

\begin{table}[!t]
\caption{Comparative results of various heatmap generation methods on Pascal VOC 2007}
\label{tab:heatmap-methods}
\centering
\newcolumntype{M}{>{\centering\arraybackslash}p{31pt}}
\begin{tabular}{l | l | M | M}
\toprule[0.15em]
Method & Heatmap & mAP & mCorLoc \\
\midrule[0.05em]
\multirow{8}{*}{DANCE} & - & 46.4 & 65.3 \\
                       & CAM \cite{Zhou2016CAM} \textit{CVPR'16} & 49.6 & 69.9 \\
                       & Grad-CAM \cite{Selvaraju2017Grad-CAM} \textit{ICCV'17} & 49.1 & 69.2 \\
                       & AffinityNet \cite{Ahn2018AffinityNet} \textit{CVPR'18} & 51.2 & 72.3 \\
                       & IRNet \cite{Ahn2019IRNet} \textit{CVPR'19} & 52.0 & 73.4 \\
                       & W-OoD \cite{Lee2022W-OoD} \textit{CVPR'22} & 55.2 & 77.5 \\
                       & ACR \cite{Kweon2023ACR} \textit{CVPR'23} & 55.9 & 78.1 \\
                       & S2C \cite{Kweon2024S2C} \textit{CVPR'24} & \textbf{58.5} & \textbf{81.8} \\
\bottomrule[0.15em]
\end{tabular}
\end{table}

We present the performance of DANCE using various heatmap generation methods in Table \ref{tab:heatmap-methods}. The first row represents the baseline setting without heatmaps, where the top-scoring proposal for each category existing in the image is directly assigned as a pseudo GT box. As observed, even with the most fundamental heatmap extraction methods, CAM \cite{Zhou2016CAM} and Grad-CAM \cite{Selvaraju2017Grad-CAM}, DANCE outperforms the baseline by 3.2\% and 2.7\% in mAP respectively. This demonstrates the inherent effectiveness of our proposed HGPS selection strategy. As long as the provided heatmap can roughly localize the object, applying dual thresholds on the heatmap in conjunction with the score matrix to filter pseudo GT boxes would be better than the global top-scoring selection strategy. Naturally, as the continuous improvement of heatmaps over the years, the performance of our method increases accordingly. When employing S2C \cite{Kweon2024S2C}, our method ultimately achieves 58.5\% mAP and 81.8\% mCorLoc.

\subsubsection{Each Component in WSBDN}

\begin{table}[!t]
\caption{Effect of each component in WSBDN on Pascal VOC 2007}
\label{tab:in-WSBDN}
\centering
\newcolumntype{P}{>{\centering\arraybackslash}p{38pt}}
\newcolumntype{M}{>{\centering\arraybackslash}p{31pt}}
\resizebox{\columnwidth}{!}{
\begin{tabular}{l | P P P P | M | M}
\toprule[0.15em]
Method & $\mathcal{L}_{\mathrm{img}}$ & $C \rightarrow C+1$ & $\mathcal{L}_{\mathrm{cls}}^{(0)}$ & $\mathcal{L}_{\mathrm{cls-ign}}^{(0)}$ & mAP & mCorLoc \\
\midrule[0.05em]
\multirow{6}{*}{WSBDN} & $\surd$ &  &  &  & 34.0 & 57.5 \\
                       & $\surd$ & $\surd$ &  &  & 34.2 & 57.6 \\
                       &  & $\surd$ & $\surd$ &  & 41.4 & 77.1 \\
                       &  & $\surd$ & $\surd$ & $\surd$ & 41.5 & 77.6 \\
                       & $\surd$ & $\surd$ & $\surd$ &  & 52.0 & 80.4 \\
                       & $\surd$ & $\surd$ & $\surd$ & $\surd$ & \textbf{52.2} & \textbf{80.6} \\
\bottomrule[0.15em]
\end{tabular}
}
\end{table}

Table \ref{tab:in-WSBDN} shows the effect of each component in WSBDN, of which the first row corresponds to baseline WSDDN. The first two rows and last two rows denote using $\bm{ws}^{(0)}$ for training and testing, while the middle two rows indicate training and testing with only the class-wise softmax branch. As shown in the first two rows, simply expanding the output dimension of $\bm{s}^{(0)}$ and $\bm{w}^{(0)}$ from $\mathbb{R}^{R \times C}$ to $\mathbb{R}^{R \times (C + 1)}$ yields a modest improvement of +0.2\% in mAP and +0.1\% in mCorLoc. Although this gain is minor, it provides a form in which the class-wise softmax branch can be directly supervised. This new supervision leads to a significant performance leap, boosting mAP and mCorLoc to 41.5\% and 77.6\% --- a massive improvement of +7.5\%/+20.1\%, proving that it is a very effective measure. Additionally, the negative certainty supervision can bring about +0.1\% $\sim$ 0.2\% improvement on performance. When adding softmax over proposals branch and $\mathcal{L}_{\mathrm{img}}$ back, our designed WSBDN base model can ultimately achieve 52.2\% (+18.2\%) mAP and 80.6\% (+23.1\%) mCorLoc, representing tremendous improvement over its baseline WSDDN.

\subsubsection{Influence of WSBDN and HGPS}

\begin{table}[!t]
\caption{Performance comparison of standalone WSBDN and WSDDN on Pascal VOC 2007}
\label{tab:WSBDN-WSDDN}
\centering
\newcolumntype{L}{>{\raggedright\arraybackslash}p{40pt}}
\newcolumntype{M}{>{\centering\arraybackslash}p{31pt}}
\begin{tabular}{L | M M | M M}
\toprule[0.15em]
\multirow{2}{*}{Method} & \multicolumn{2}{c|}{$\bm{s}^{(0)}$} & \multicolumn{2}{c}{$\bm{ws}^{(0)}$} \\
\cmidrule[0.05em]{2-5}
                        & \multicolumn{1}{M|}{mAP} & mCorLoc & \multicolumn{1}{M|}{mAP} & mCorLoc \\
\midrule[0.05em]
WSDDN & \multicolumn{1}{M|}{5.0} & 24.2 & \multicolumn{1}{M|}{34.0} & 57.5 \\
\midrule[0.05em]
WSBDN & \multicolumn{1}{M|}{43.0} & 78.3 & \multicolumn{1}{M|}{52.2} & 80.6 \\
\bottomrule[0.15em]
\end{tabular}
\end{table}

\begin{table}[t]
\caption{Effect of WSBDN over different benchmarks on Pascal VOC 2007}
\label{tab:WSBDN+}
\centering
\newcolumntype{L}{>{\raggedright\arraybackslash}p{66pt}}
\newcolumntype{M}{>{\centering\arraybackslash}p{31pt}}
\begin{tabular}{L | M | M}
\toprule[0.15em]
Method & mAP & mCorLoc \\
\midrule[0.05em]
WSDDN+OICR & 46.2 & 65.2 \\
WSBDN+OICR & \textbf{54.8} & \textbf{76.1} \\
\midrule[0.05em]
WSDDN+PCL & 48.1 & 68.0 \\
WSBDN+PCL & \textbf{56.5} & \textbf{79.2} \\
\midrule[0.05em]
WSDDN+HGPS & 57.3 & 80.1 \\
WSBDN+HGPS & \textbf{58.5} & \textbf{81.8} \\
\bottomrule[0.15em]
\end{tabular}
\end{table}

We first compare the performance of WSBDN with WSDDN in Table \ref{tab:WSBDN-WSDDN} to verify the efficacy of WSBDN. It is evident that WSBDN significantly narrows the performance gap between the $\bm{s}^{(0)}$ and $\bm{ws}^{(0)}$. For WSDDN, the performance of $\bm{s}^{(0)}$ lags behind that of $\bm{ws}^{(0)}$ by a substantial 85.3\% in mAP and 57.9\% in mCorLoc. In contrast, WSBDN narrows this gap to just 17.3\% and 2.9\% on the respective metrics. This reduction in the performance gap also leads to a significant improvement in overall performance: WSBDN can achieve 52.0\% mAP and 80.6\% mCorLoc as a standalone detector, outperforming WSDDN by +18.1\% and +23.1\% on these two metrics respectively.

Subsequently, we respectively adopt WSDDN and WSBDN as the basic MIDN module on several benchmarks and our HGPS to evaluate the performance gains. The results are presented in Table \ref{tab:WSBDN+}. As shown in the first four rows, when combined with the OICR and PCL refinement algorithms, our WSBDN module achieves significant performance gains over the baseline WSDDN. The mAP and the mCorLoc increased by tremendous +8.6\%/+10.9\% and +8.4\%/+11.2\% respectively, validating the effectiveness and versatility of our proposed WSBDN base module.

The last two rows of Table \ref{tab:WSBDN+} demonstrate that our designed HGPS algorithm is highly effective on its own. Even when paired with the WSDDN module, HGPS alone achieves 57.3\% mAP and 80.1\% mCorLoc, confirming its effectiveness.

\subsubsection{Convergence Acceleration of Negative Certainty Supervision}

\begin{figure}[t]
\centering
\includegraphics[width=0.92\columnwidth]{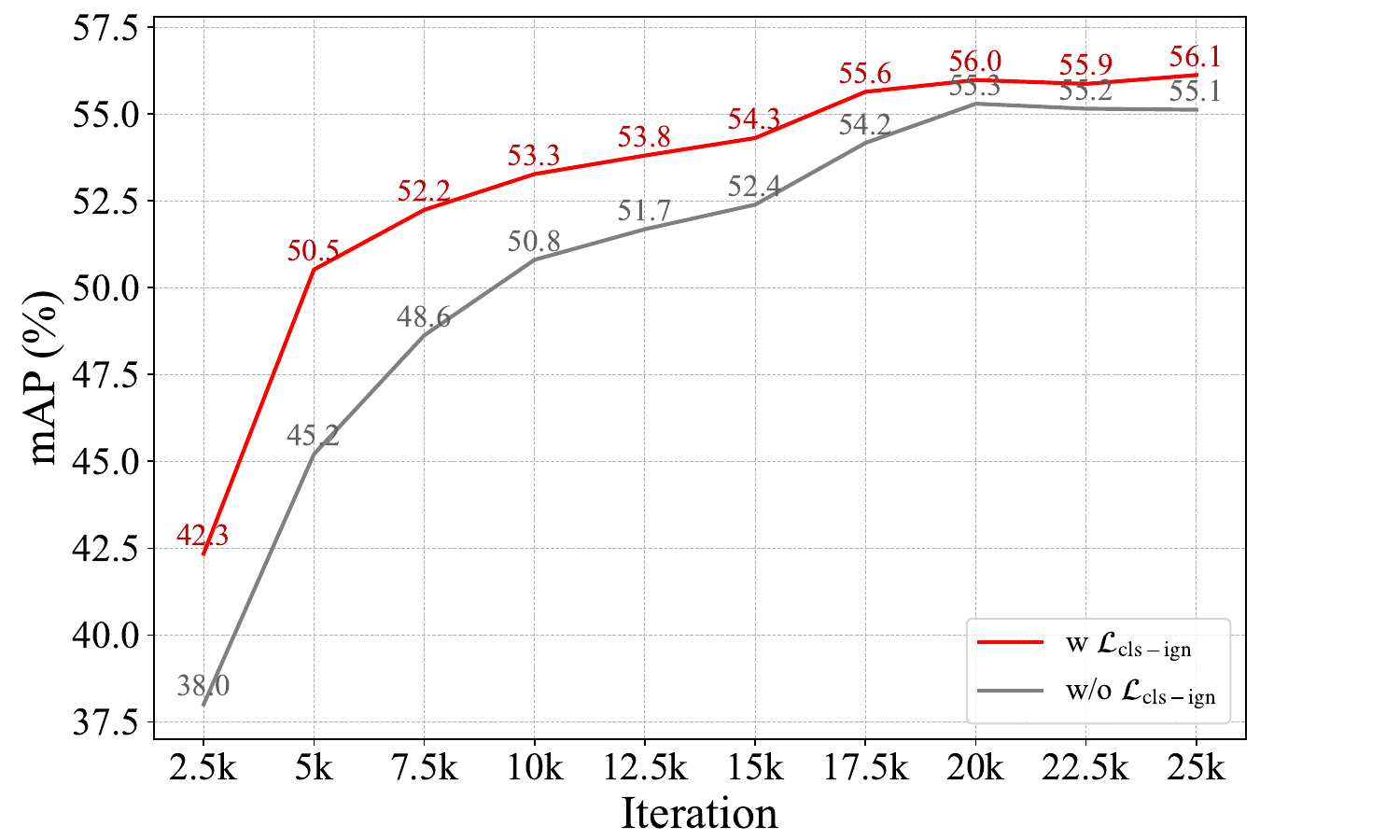}
\caption{Convergence curves with and without the classification-ignored loss on Pascal VOC 2007 (without TTA). It demonstrates acceleration of our negative certainty supervision.}
\label{fig:accelerate}
\end{figure}

We investigated the impact of the classification-ignored loss on the network's convergence speed. Due to time consumption for testing, we only present the results without TTA. As shown in Figure \ref{fig:accelerate}, after 2,500 training iterations, the network with $\mathcal{L}_{\mathrm{cls-ign}}$ reaches an mAP of 42.3\%, while the one without it only achieves 38.0\%. By 5,000 iterations, the network with $\mathcal{L}_{\mathrm{cls-ign}}$ has already surpassed 50\% mAP, whereas its counterpart without the loss is still around 45\%. Starting from the 20,000-th iteration, the network with $\mathcal{L}_{\mathrm{cls-ign}}$ converges and fluctuates around 56\% mAP. In contrast, the baseline eventually stabilizes at approximately 55\% mAP. This demonstrates that supervising ignored proposals with negative certainty information can improve the convergence rate of the whole network.

\subsubsection{Each Innovation of DANCE}

\begin{table}[t]
\caption{Effect of each innovation of DANCE on Pascal VOC 2007}
\label{tab:of-DANCE}
\centering
\newcolumntype{P}{>{\centering\arraybackslash}p{36pt}}
\newcolumntype{M}{>{\centering\arraybackslash}p{38pt}}
\resizebox{\columnwidth}{!}{
\begin{tabular}{l | P P P | M | M}
\toprule[0.15em]
Method & HGPS & WSBDN & NCS & mAP & mCorLoc \\
\midrule[0.05em]
\multirow{7}{*}{DANCE} &  &  &  & 46.2 & 65.2 \\
                            &  &  & $\surd$ & 46.8 & 66.0 \\
                            & $\surd$ &  &  & 57.2 & 79.9 \\
                            & $\surd$ &  & $\surd$ & 57.3 & 80.1 \\
                            &  & $\surd$ &  & 54.4 & 75.8 \\
                            &  & $\surd$ & $\surd$ & 54.8 & 76.1 \\
                            & $\surd$ & $\surd$ & $\surd$ & \textbf{58.5} & \textbf{81.8} \\
\bottomrule[0.15em]
\end{tabular}
}
\end{table}

Table \ref{tab:of-DANCE} summarizes the performance improvements brought by our three proposed components: HGPS, WSBDN, and NCS. Notably, when conducting the ablation study for WSBDN, we must maintain the structural integrity of the overall model. This means subsequent IR modules are still required, but the HGPS selection strategy is excluded. Therefore, we adopt the vanilla OICR selection method as our baseline. Similarly, evaluating HGPS independently also requires a base module, for which we naturally adopt WSDDN to derive the standalone HGPS results. Consequently, rows 1, 4, and 6 in Table \ref{tab:of-DANCE} are identical to rows 1, 5, and 2 in Table \ref{tab:WSBDN+}.

Comparing rows 1 with 3, and 1 with 5 in Table \ref{tab:of-DANCE}, it is evident that using the HGPS selection strategy or the WSBDN module independently yields substantial performance gains of +11.0\%/+14.7\% and +8.2\%/+10.6\%, respectively. Notably, HGPS alone achieves 57.2\% mAP and 79.9\% mCorLoc, which is remarkably close to the final performance (58.5\%/81.8\%), demonstrating its significant effectiveness. Through pairwise comparisons (rows 1 vs. 2, 3 vs. 4, and 5 vs. 6), we observe that the NCS loss alone provides a +0.6\%/+0.8\% boost. However, when added to HGPS or WSBDN, the incremental gain from NCS diminishes to +0.1\%/+0.2\% or +0.4\%/+0.3\%. This suggests that the efficacy of NCS is highly sensitive to the capability of the initial baseline. When the baseline is weak, the noise from ignored proposals is more disruptive, making the classification-ignored loss more impactful. Conversely, with a sufficiently strong baseline, the inherent noise level of these proposals is already low even without explicit constraints, leading to less pronounced improvements from NCS. This observation indirectly underscores the superiority of HGPS and WSBDN. Finally, when all three innovations are integrated, the complete DANCE framework achieves 58.5\%/81.8\%, outperforming each individual component and surpassing the previous SOTA.

\section{Conclusion}

In this paper, we propose a DANCE framework. First, we devise the HGPS algorithm, which utilizes high and low thresholds to pre-filter proposals and dynamically selects pseudo GT boxes in conjunction with the classification score matrix. This enables the pseudo GT boxes to not only avoid being confined solely to the discriminative parts of objects but also successfully distinguish between adjacent intra-class instances. Second, we construct the WSBDN network to reintroduce the background class confidence for proposals and bridge the semantic gap between different matrices. Finally, we introduce the NCS loss, which imposes deterministic negative supervision signals on proposals ignored during training, thereby speeding up convergence. Extensive experiments on PASCAL VOC and COCO datasets demonstrate the superiority of our method.

\section*{Acknowledgments}

This work was supported in part by the Joint Funds of the National Natural Science Foundation of China (Grant No. U22A2036), in part by the National Natural Science Foundation of China (NSFC) / Research Grants Council (RGC) Collaborative Research Scheme (Grant No. 62461160332 \& CRS\_HKUST602/24), in part by the National Natural Science Foundation of China (NSFC) (Grant No.62472122), in part by the Shenzhen Stable Supporting Program (Grant No. GXWD20231130110352002), in part by the Shenzhen Science and Technology Program (Grant No. ZDCY20250901101035012), in part by the Guangdong Basic and Applied Basic Research Foundation (Grant No. 2023A1515110271), and in part by the 2025 Research and Development of Security Detection and Defense Software for AI Models project from China Unicom Intelligent Manufacturing Technology \& Industry (Guangdong) Co., Ltd. (No. YGS25AE1000001). \textit{(Corresponding author: Weizhe Zhang.)}

\appendices
\section{Category-Specific Heatmaps Generator}

We employ From SAM to CAMs (S2C) \cite{Kweon2024S2C} to generate category-specific heatmaps for each image.

Given an image $\bm{I} \in \mathbb{R}^{H \times W \times 3}$, we first build a standard CAM \cite{Zhou2016CAM} network as our heatmap generator. Concretely, we employ an image encoder to extract the feature map $\bm{F} \in \mathbb{R}^{H' \times W' \times D'}$, followed by applying $C$ convolutional kernels of size $1 \times 1 \times D'$ to $\bm{F}$ to produce class activation maps $\bm{A} \in \mathbb{R}^{H' \times W' \times C}$. By using a Global Average Pooling (GAP) layer to $\bm{A}$ along its spatial dimensions in the end, we obtain an image-level class prediction score vector whose dimension is $\mathbb{R}^C$:
\begin{equation}
    \bm{\varphi}^{\mathrm{CAM}} = \mathrm{GAP} \left(\bm{A}\right).
\end{equation}
For multi-label classification, we train the CAM module through the binary cross-entropy loss function:
\begin{equation} \label{equ:L_S2C-CLS}
\resizebox{0.9\columnwidth}{!}{$\displaystyle
    \mathcal{L}_{\mathrm{S2C-CLS}} = -\sum_{c=1}^{C} \left[y_c \log \varphi_c^{\mathrm{CAM}} + \left(1 - y_c\right) \log \left(1 - \varphi_c^{\mathrm{CAM}}\right)\right].
$}
\end{equation}

Additionally, we follow S2C by using SAM-Segment Contrasting (SSC) to obtain reliable segmentation masks for each image. Specifically, we send each image into SAM \cite{Kirillov2023SAM} and utilize the segment-everything option to generate segments. Noticing that the predicted segments may overlap, we assign a pixel to the segment with the smallest region that includes it, thus resulting in a single segmentation map. We define the segmentation map $\bm{SE}$ as a partitioned space where:
\begin{enumerate}[(i)]
    \item $\bm{SE} = \bigcup_{g=1}^{G} \bm{SE}_g$,
    \item $\bm{SE}_i \cap \bm{SE}_j = \varnothing$, $\forall~1 \leq i, j \leq G$, s.t. $i \neq j$,
\end{enumerate}
with $G = \left\vert \bm{SE} \right\vert$ being segment cardinality. We then construct a prototype $\bm{pt}_g$ for each segment $\bm{SE}_g$ as:
\begin{equation}
    \bm{pt}_g = \frac{1}{\left\vert \bm{SE}_g \right\vert} \sum_{\left(h, w\right) \in \bm{SE}_g} \bm{F}_{h', w'},
\end{equation}
where $\left(h, w\right)$ denotes the coordinate position in the original image $\bm{I}$ corresponding to the point $\left(h', w'\right)$ in the feature map $\bm{F}$, and $\bm{pt}_g$ represents the average feature vector of all pixels within $\bm{SE}_g$. In this way, we expect all the pixels' features to have similar representations within a single segment, so we optimize feature representations to converge toward their respective prototypes and build the loss function as follows:
\begin{equation} \label{equ:L_S2C-SSC}
    \mathcal{L}_{\mathrm{S2C-SSC}} = -\sum_{g=1}^{G} \sum_{\left(h, w\right) \in \bm{SE}_g} \frac{\bm{F}_{h', w'} \cdot \bm{pt}_g / T}{\sum_{i=1}^{G} \bm{F}_{h', w'} \cdot \bm{pt}_i / T},
\end{equation}
where $T$ is the temperature coefficient.

At last, we follow S2C by building the CAM-based Prompting Module (CPM) to generate pseudo pixel-level segmentation labels for each input image $\bm{I}$. Specifically, we use a local maximum filter LMF to extract multiple peaks from CAMs as follows: $\bm{p}_c = \mathrm{LMF} \left(\bm{A}_c\right)$, where $\bm{A}_c$ is the CAM of class $c$ and $\bm{p}_c = \left\{\bm{p}_{c, 1}, \bm{p}_{c, 2}, \cdots, \bm{p}_{c, k_c}\right\}$ is the set of the obtained peak points. We then utilize the peak points as point prompts for the SAM and obtain:
\begin{equation}
    \bm{M}_c^{\mathrm{SAM}}, \bm{S}_c^{\mathrm{SAM}} = \mathrm{SAM} \left(\bm{I}; \bm{p}_c\right),
\end{equation}
where $\bm{M}_c^{\mathrm{SAM}}, \bm{S}_c^{\mathrm{SAM}} \in \mathbb{R}^{H' \times W'}$, and the former means the refined category-specific mask, which is composed of binary elements $\left\{0, 1\right\}$, while the later indicates the reliability of each $\left\{0, 1\right\}$-segmented pixel. Here, the pixels with higher stability scores are more likely to be segmented along the given prompt. In the end, we gather the results of SAM and CAMs together to obtain pseudo pixel-level segmentation labels. We first average the activation of CAM for each class based on the SAM mask as follows:
\begin{equation}
    \alpha_c^{\mathrm{CAM}} = \frac{1}{\left\vert \bm{M}_c^{\mathrm{SAM}} \right\vert} \sum_{ \left(h', w'\right) \in \bm{M}_c^{\mathrm{SAM}}} \bm{A}_{h', w', c},
\end{equation}
and then consider it as the reliability of the CAM, therefore the proposed confidence map of each class $c$ is defined as $\bm{S}_c = \alpha_c^{\mathrm{CAM}} \cdot \bm{S}_c^{\mathrm{SAM}} \in \mathbb{R}^{H' \times W'}$, so that the pseudo segmentation map $\hat{\bm{S}} \in \mathbb{R}^{H' \times W'}$ is acquired by:
\begin{equation}
\resizebox{0.88\columnwidth}{!}{$\displaystyle
    \hat{S}_{h', w'} =
    \begin{cases}
        \arg \max\limits_{c} S_{h', w', c}, & \mbox{if}~\max\limits_{c} S_{h', w', c} \geq \tau_{\mathrm{S2C}} \\
        C+1, & \mbox{otherwise}
    \end{cases}
$}
\end{equation}
where $\tau_{\mathrm{S2C}}$ denotes the pseudo confidence threshold separating foreground and background pixels. Subsequently, we define the background activation map $\bm{A}_{C+1} \in \mathbb{R}^{H' \times W'}$ using CAMs as follows: $A_{h', w', C+1} = 1 - \max\limits_{c} A_{h', w', c}$ and concatenate it with the origin CAMs together to generate $\bm{A}^{+} \in \mathbb{R}^{H' \times W' \times \left(C+1\right)}$ for building the cross-entropy loss function:
\begin{equation} \label{equ:L_S2C-CPM}
\resizebox{0.88\columnwidth}{!}{$\displaystyle
    \mathcal{L}_{\mathrm{S2C-CPM}} = -\sum_{h'=1}^{H'}\sum_{w'=1}^{W'}\sum_{c=1}^{C+1} \mathbbm{1}\left[\hat{S}_{h', w'}=c\right] \log A_{h', w', c}^{+}.
$}
\end{equation}

By jointly optimizing the three losses defined in Equations \ref{equ:L_S2C-CLS}, \ref{equ:L_S2C-SSC}, and \ref{equ:L_S2C-CPM}, the entire CAM network is trained to get high-quality heatmaps:
\begin{equation}
    \mathcal{L}_{\mathrm{S2C}} = \mathcal{L}_{\mathrm{S2C-CLS}} + \mathcal{L}_{\mathrm{S2C-SSC}} + \mathcal{L}_{\mathrm{S2C-CPM}}.
\end{equation}

\section{Detailed Motivation of HGPS} \label{app:B}

It is a widely recognized consensus in the field of WSOD that if pseudo GT boxes are generated by simply selecting high-scoring proposals from the classification score matrix, these proposals will eventually converge to the discriminative part of an object. Consequently, the pseudo GT boxes fail to capture the full extent of an object, leading to suboptimal model performance. We note that several existing methods \cite{Diba2017WCCN, Zhang2018ZLDN, Zhang2018ML-LocNet, Chen2020SLV, Liao2022SPE} have been dedicated to solving this problem by using a heatmap thresholding approach, which enables pseudo GT boxes to capture the full extent of objects. However, this series of methods suffers from another inherent problem: when adjacent intra-class instances are present in the image, the areas within these objects would like to exhibit a certain amount of heat. As a result, the pseudo GT boxes tend to merge them into a single, large box, failing to distinguish individual instances, a phenomenon also demonstrated in Figure \ref{fig:hgps-vis}\subref{fig:hgps-vis-d}.

This leads us to a crucial question: why do all these methods choose a low threshold to generate pseudo GT boxes? The reason is that a higher threshold would result in heatmaps failing to capture the full extent of the object, omitting parts of its contours. This observation, however, provided us with significant inspiration. By observing the heatmap in Figure \ref{fig:hgps-vis}\subref{fig:hgps-vis-b}, we noticed that the heat is highest at the center of an individual object and gradually decreases towards its edges. Therefore, setting a higher threshold, while not encompassing the entire object, could effectively distinguish each individual instance.

However, the preceding phenomena and analysis also reveal that when thresholding heatmaps, neither a low nor a high threshold can precisely capture the bounding box of a single object. Therefore, relying solely on thresholding heatmaps will inevitably hit a performance bottleneck. This leads us to consider: how can we obtain pseudo GT boxes that better align with the real annotations? Considering that region proposals are generated from intrinsic image properties like edges and textures, they offer a rich set of candidates that often tightly bound the actual objects. For this reason, we still select pseudo GT boxes from the pool of proposals rather than generating them directly via heatmap thresholding. This way, we at least guarantee that boxes closely fitting the real annotations have the potential to become pseudo GT boxes.

So, the problem now becomes: how can we prevent the selected pseudo GT boxes from concentrating solely on the discriminative part of an object? A natural idea is to use the positional information provided by heatmaps to anchor the proposals. High-threshold boxes, while not capturing the full object, at least indicate its general location. Low-threshold boxes, while unable to separate adjacent intra-class instances, can at least cover the complete object. Therefore, we only need to perform a pre-selection step: picking out proposals that lie between the high- and low-threshold boxes. Because the high-threshold boxes are not only concentrated on the discriminative part of objects, the proposals we pre-selected are not those that only contain the discriminative part, and because the low-threshold boxes cover the complete object, some of the selected proposals are certain to encompass the objects' complete contour. As shown in Figure \ref{fig:hgps-vis}\subref{fig:hgps-vis-e}, the experimental results also confirm the aforementioned conjecture. Therefore, we only choose the final pseudo GT boxes from these pre-selected sets, which means that we use heatmaps for pre-processing to discard a portion of low-quality proposals first, completely eliminating their possibility of becoming pseudo GT boxes. Now, even if a proposal on a discriminative part receives a very high score, it will not become a final pseudo GT box because it is not in our pre-selected sets.

Now, there leaves one final question: if adjacent intra-class instances are present in an image, some proposals might span several of them. These proposals are also located between the high- and low-threshold boxes, meaning they would be included in our pre-selected set. So, how can we avoid selecting these boxes as pseudo GT boxes? Our solution is to incorporate the classification score matrix once again. This is because high scores tend to favor each object's discriminative part, exhibiting a tendency to contract inwards. As the area of the proposal expands, the score decreases. Therefore, a proposal containing only a single object will have a higher score than a proposal spanning multiple instances. Consequently, by selecting the highest-scoring proposal from each pre-selected set to serve as the pseudo GT box, we form the complete HGPS algorithm. This approach circumvents the various defects of previous methods, obtaining higher-quality pseudo GT boxes, and achieving state-of-the-art performance.

\section{Detailed Motivation of WSBDN} \label{app:C}

We think the fundamental reason for the dimensional deficiency problem in WSDDN's \cite{Bilen2016WSDDN} class representation first. The issue arises because the concept of ``background category object'' does not exist at the image level. Consequently, the image-level label has only a dimension of $\mathbb{R}^C$, not $\mathbb{R}^{C+1}$, which, in turn, makes WSDDN impossible to construct an $\mathbb{R}^{R \times (C+1)}$ score matrix.

Delving deeper, this problem reveals an inherent gap between multi-label classification and object detection tasks. In object detection, some bounding boxes are located far from any foreground objects and thus must be assigned a ``background'' label. Therefore, the score vector for each box must have a shape to specially represent background, resulting in a dimension of $\mathbb{R}^{C+1}$. In contrast, multi-label classification is an image-level task where the notion of ``background category object'' is nonsensical. As such, the image-level label vector only has a dimension of $\mathbb{R}^C$.

In the WSOD task setting, supervision is restricted to image-level signals, which necessitates the use of a basic MIDN module. It is this very combination --- the reliance on an MIL module under the constraint of incomplete image-level supervision --- that ultimately causes the dimensional deficiency in WSDDN's class representation.

\begin{figure}[!t]
\centering
\includegraphics[width=\columnwidth]{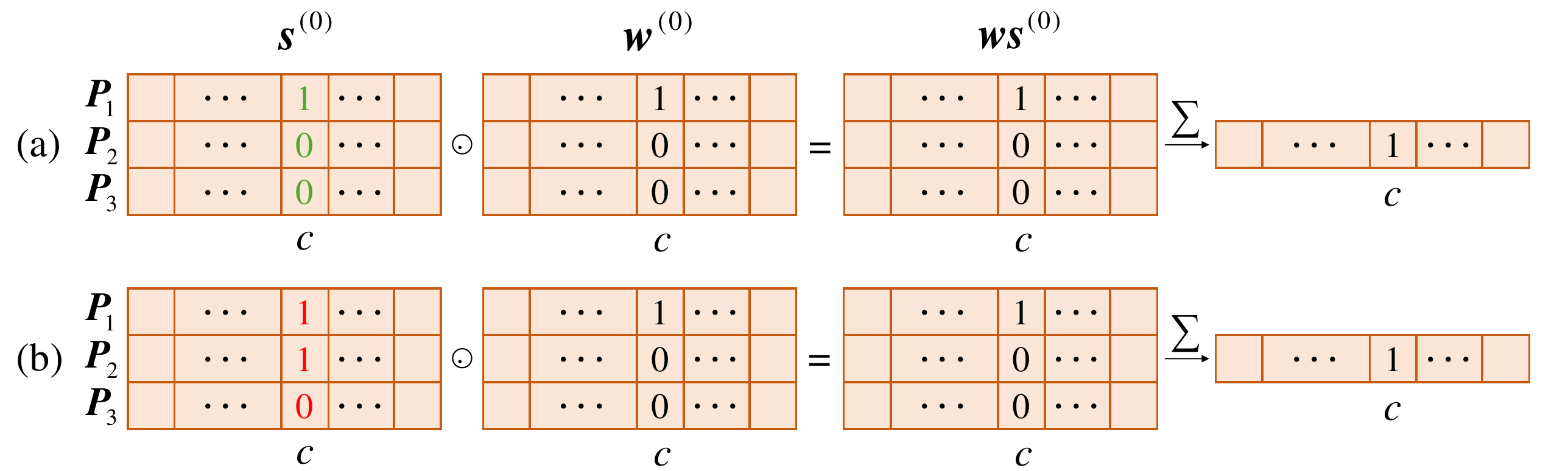}
\caption{Results of the matrices in the process of WSDDN. (a) Desired results. (b) Possible results.}
\label{fig:wsddn-inconsistency}
\end{figure}

Therefore, we redefine the setting of the WSOD task to align it with the paradigm of object detection rather than multi-label classification. Specifically, we change the shape of the WSOD task's label vector $\bm{y}$ from $\mathbb{R}^C$ to $\mathbb{R}^{C+1}$. This change is rooted in a conceptual shift: given that the object detection task evaluates a detector's classification and regression capabilities on a specific distribution of boxes (i.e., proposals), we conceptually push down the label definition from image level to box level to align with the task's scope. We argue that $y_c=1$ or $0$ does not represent the presence or absence of an object of category $c$ in the image, but rather the presence or absence of a box representing an object of category $c$. Clearly, there must be a subset of proposals in any given image representing the background class. Therefore, under this new definition, we stipulate that $y_{C+1}=1$ always holds. Through this mechanism, we align the labels of WSOD with FSOD. We term this redefined label $\bm{y} \in \mathbb{R}^{C+1}$ the ``box-level image label''.

Next, we explore the fundamental reason why WSDDN requires an additional proposal-wise softmax branch, which is absent in FSOD models. This necessity arises because the WSOD label only tells which classes are present, but without any ground-truth annotations. Consequently, with only the class-wise branch, it is impossible to formulate a loss function and train the model, as there is no way to determine the target label for each proposal's score vector. Therefore, the box-level classification score matrix $\bm{s}^{(0)}$ must undergo an additional linear or non-linear transformation to be reduced to the image level, thereby establishing a connection with the image label. This reduction process must also consider the score information of each proposal, which necessitates a weighted sum. The crux is designing these weights to produce a final score that behaves like a probability --- ranging from 0 to 1 and correlating with the confidence of the class's presence. As a result, an additional weight branch is introduced to produce $\bm{w}^{(0)}$. Its purpose is to ``compress'' the scores by assigning a weight to each proposal's prediction score. For example, consider an image with the ``person'' category and 2,000 proposals. If 100 of these proposals are close to the people's ground-truth boxes, their score vectors should ideally have a value of 1 in the shape of person.  A direct summation along the proposal axis would yield a total score of 100 for this class, making it impossible to establish a loss against the binary image-level label (0 or 1). The weight matrix resolves this by effectively normalizing the contributions. Through the Hadamard product, if each of the 100 positive proposals is assigned a weight of approximately 1/100, their weighted scores can be summed to 1, bridging the gap to the image-level label. This ensures that the final image-level score $s_c^{\mathrm{img}}$ is bounded between 0 and 1:
\begin{equation}
\begin{aligned}
s_c^{\mathrm{img}} &= \sum_{r=1}^{R} ws_{r, c}^{(0)} = \sum_{r=1}^{R} s_{r, c}^{(0)} \times w_{r, c}^{(0)} \\
&\geq \sum_{r=1}^{R} 0 \times 0 = 0,
\end{aligned}
\end{equation}
while
\begin{equation}
\begin{aligned}
s_c^{\mathrm{img}} &= \sum_{r=1}^{R} ws_{r, c}^{(0)} = \sum_{r=1}^{R} s_{r, c}^{(0)} \times w_{r, c}^{(0)} \\
&\leq \sum_{r=1}^{R} w_{r, c}^{(0)} = 1,
\end{aligned}
\end{equation}
which allows for the construction of a loss function based on image-level labels.

However, as shown in Table \ref{tab:WSDDN}, although the performance of $\bm{ws}^{(0)}$ is reasonably good, the performance of $\bm{s}^{(0)}$ is catastrophic. To illustrate with the previous example, this means that while the down-weighted scores of the 100 proposals representing the ``person'' class perform decently, their original, unweighted scores for this class do not converge towards 1 at all.

This result is, in fact, predictable. The training objective lacks an explicit constraint on $\bm{s}^{(0)}$, leading to ambiguity in its learned values. For example, as shown in Figure \ref{fig:wsddn-inconsistency}\textcolor{red}{(a)}, consider an image containing class $c$ and three region proposals: $\bm{P}_1, \bm{P}_2,$ and $\bm{P}_3$. Suppose $\bm{P}_1$ closely overlaps with a ground-truth object, while $\bm{P}_2$ and $\bm{P}_3$ are distant. Ideally, $\bm{P}_1$ should be identified as a positive instance for class $c$, while $\bm{P}_2$ and $\bm{P}_3$ should be negatives. So we would naturally expect the learned score matrix $\bm{s}^{(0)}$ to satisfy $s_{1, c}^{(0)} = 1$ and $s_{2, c}^{(0)} = s_{3, c}^{(0)} = 0$. However, an alternative solution where $s_{1, c}^{(0)} = s_{2, c}^{(0)} = 1$ and $s_{3, c}^{(0)} = 0$, combined with weights $w_{1, c}^{(0)} = 1$ and $w_{2, c}^{(0)} = w_{3, c}^{(0)} = 0$, produces the exact same final matrix $\bm{ws}^{(0)}$ as depicted in Figure \ref{fig:wsddn-inconsistency}\textcolor{red}{(b)}. Therefore, without additional supervision, the model has no incentive to prefer the clean solution in (a). As a result, $\bm{s}^{(0)}$ can become noisy, which causes information inconsistency between $\bm{s}^{(0)}$ and $\bm{ws}^{(0)}$.

Such a performance discrepancy severely impacts the final outcome. To address this, we use the pre-obtained proposal clusters to supervise $\bm{s}^{(0)}$. This supervision aims to bridge the performance gap between $\bm{s}^{(0)}$ and the final weighted scores $\bm{ws}^{(0)}$, ultimately leading to an improvement in the model's overall efficacy.

\section{Detailed Comparisons} \label{app:D}

\begin{table*}[!t]
\caption{Comparison with the state-of-the-art methods on PASCAL VOC 2012 test set in terms of AP (\%).}
\label{tab:mAP-VOC12}
\centering
\newcolumntype{P}{>{\centering\arraybackslash}p{21pt}}
\newcolumntype{M}{>{\centering\arraybackslash}p{31pt}}
\begin{threeparttable}
\resizebox{\textwidth}{!}{
\begin{tabular}{l | P P P P P P P P P P P P P P P P P P P P | M}
\toprule[0.15em]
Method & aero & bike & bird & boat & bottle & bus & car & cat & chair & cow & table & dog & horse & mbike & person & plant & sheep & sofa & train & tv & mAP \\
\midrule[0.05em]
ContextLocNet \cite{Kantorov2016ContextLocNet} & 64.0 & 54.9 & 36.4 & 8.1 & 12.6 & 53.1 & 40.5 & 28.4 & 6.6 & 35.3 & 34.4 & 49.1 & 42.6 & 62.4 & 19.8 & 15.2 & 27.0 & 33.1 & 33.0 & 50.0 & 35.3 \\
OICR \cite{Tang2017OICR} & 67.7 & 61.2 & 41.5 & 25.6 & 22.2 & 54.6 & 49.7 & 25.4 & 19.9 & 47.0 & 18.1 & 26.0 & 38.9 & 67.7 & 2.0 & 22.6 & 41.1 & 34.3 & 37.9 & 55.3 & 37.9 \\
WCCN \cite{Diba2017WCCN} & - & - & - & - & - & - & - & - & - & - & - & - & - & - & - & - & - & - & - & - & 37.9 \\
TS$^{2}$C \cite{Wei2018TS2C} & 67.4 & 57.0 & 37.7 & 23.7 & 15.2 & 56.9 & 49.1 & 64.8 & 15.1 & 39.4 & 19.3 & 48.4 & 44.5 & 67.2 & 2.1 & 23.3 & 35.1 & 40.2 & 46.6 & 45.8 & 40.0 \\
PCL \cite{Tang2018PCL} & 58.2 & 66.0 & 41.8 & 24.8 & 27.2 & 55.7 & 55.2 & 28.5 & 16.6 & 51.0 & 17.5 & 28.6 & 49.7 & 70.5 & 7.1 & 25.7 & 47.5 & 36.6 & 44.1 & 59.2 & 40.6 \\
WS-JDS \cite{Shen2019WS-JDS} & - & - & - & - & - & - & - & - & - & - & - & - & - & - & - & - & - & - & - & - & 39.1 \\
OAIL \cite{Kosugi2019OAIL} & 70.2 & 61.3 & 43.8 & 28.9 & 23.5 & 54.0 & 52.1 & 55.2 & 19.1 & 51.0 & 15.6 & 52.6 & 56.6 & 68.9 & 22.0 & 21.7 & 43.6 & 37.0 & 34.8 & 56.3 & 43.4 \\
SDCN \cite{Li2019SDCN} & - & - & - & - & - & - & - & - & - & - & - & - & - & - & - & - & - & - & - & - & 43.5 \\
C-MIDN \cite{Gao2019C-MIDN} & 72.9 & 68.9 & 53.9 & 25.3 & 29.7 & 60.9 & 56.0 & 78.3 & 23.0 & 57.8 & 25.7 & 73.0 & 63.5 & 73.7 & 13.1 & 28.7 & 51.5 & 35.0 & 56.1 & 57.5 & 50.2 \\
CSC \cite{Shen2019CSC} & 54.8 & 52.2 & 36.5 & 18.1 & 25.4 & 55.7 & 39.1 & 47.2 & 16.1 & 39.2 & 17.9 & 39.9 & 34.2 & 56.1 & 25.2 & 20.1 & 34.6 & 30.9 & 56.4 & 41.4 & 37.1 \\
PSLR \cite{Zhang2020PSLR} & 70.6 & 63.2 & 49.1 & 31.7 & 22.1 & 59.4 & 54.4 & 53.4 & 14.0 & 55.0 & 32.7 & 64.3 & 58.3 & 69.2 & 12.8 & 23.3 & 47.2 & \textbf{40.6} & 46.7 & 58.3 & 46.3 \\
P-MIDN+MGSC \cite{Xu2021P-MIDN+MGSC} & \textbf{75.1} & 72.4 & 54.2 & 34.6 & 33.7 & 60.9 & 58.3 & 79.3 & 23.3 & 61.7 & 30.7 & 64.3 & \textbf{69.3} & 73.6 & 31.3 & 25.9 & \textbf{55.6} & 39.6 & 50.5 & 61.1 & 52.8 \\
IM-CFB \cite{Yin2021IM-CFB} & - & - & - & - & - & - & - & - & - & - & - & - & - & - & - & - & - & - & - & - & 49.4 \\
D-MIL \cite{Gao2022D-MIL} & 69.5 & 69.5 & 53.6 & 23.9 & 29.2 & 60.0 & 58.1 & 75.0 & 22.4 & 60.5 & 27.4 & 75.8 & 64.2 & 73.0 & 6.3 & 23.8 & 52.7 & 36.6 & 51.4 & 59.1 & 49.6 \\
BUAA-PAL \cite{Wu2022BUAA-PAL} & 74.4 & 74.5 & \textbf{58.1} & 29.5 & 33.5 & 58.5 & 58.5 & 70.5 & \textbf{25.9} & 64.9 & 30.8 & 59.5 & 68.8 & 74.0 & 18.2 & \textbf{29.6} & 54.0 & 34.9 & 51.8 & 54.7 & 51.2 \\
MCC-MCT \cite{Wu2024MCC-MCT} & 74.8 & \textbf{74.6} & 55.6 & 31.2 & 33.2 & 62.1 & \textbf{62.1} & 54.6 & 24.2 & \textbf{66.4} & 26.6 & 40.3 & 64.0 & \textbf{77.0} & 8.6 & 27.2 & 54.8 & 33.0 & 55.2 & 59.5 & 49.3 \\
DANCE & \textbf{75.1} & 62.0 & 57.1 & \textbf{41.7} & \textbf{36.0} & \textbf{63.8} & 57.0 & \textbf{81.1} & 19.9 & 64.2 & \textbf{37.9} & \textbf{78.3} & 68.8 & 71.3 & \textbf{53.5} & 22.6 & 50.7 & 37.8 & \textbf{69.9} & \textbf{62.8} & \textbf{55.6}\tnote{1} \\
\midrule[0.05em]
\midrule[0.05em]
OICR-Ens.+FRCNN \cite{Tang2017OICR} & 71.4 & 69.4 & 55.1 & 29.8 & 28.1 & 55.0 & 57.9 & 24.4 & 17.2 & 59.1 & 21.8 & 26.6 & 57.8 & 71.3 & 1.0 & 23.1 & 52.7 & 37.5 & 33.5 & 56.6 & 42.5 \\
W2F \cite{Zhang2018W2F} & 73.0 & 69.4 & 45.8 & 30.0 & 28.7 & 58.8 & 58.6 & 56.7 & 20.5 & 58.9 & 10.0 & 69.5 & 67.0 & 73.4 & 7.4 & 24.6 & 48.2 & 46.8 & 50.7 & 58.0 & 47.8 \\
PCL-Ens.+FRCNN \cite{Tang2018PCL} & 69.0 & 71.3 & 56.1 & 30.3 & 27.3 & 55.2 & 57.6 & 30.1 & 8.6 & 56.6 & 18.4 & 43.9 & 64.6 & 71.8 & 7.5 & 23.0 & 46.0 & 44.1 & 42.6 & 58.8 & 44.2 \\
WS-JDS+FRCNN \cite{Shen2019WS-JDS} & - & - & - & - & - & - & - & - & - & - & - & - & - & - & - & - & - & - & - & - & 46.1 \\
WSOD$^{2}$ \cite{Zeng2019WSOD2} & - & - & - & - & - & - & - & - & - & - & - & - & - & - & - & - & - & - & - & - & 47.2 \\
TPEE \cite{Yang2019TPEE} & 60.4 & 68.6 & 51.4 & 22.0 & 25.9 & 49.4 & 58.4 & 62.1 & 14.5 & 58.8 & 24.6 & 60.4 & 64.3 & 70.3 & 9.4 & 26.0 & 47.7 & 45.5 & 36.7 & 55.8 & 45.6 \\
SDCN+FRCNN \cite{Li2019SDCN} & - & - & - & - & - & - & - & - & - & - & - & - & - & - & - & - & - & - & - & - & 46.7 \\
C-MIDN+FRCNN \cite{Gao2019C-MIDN} & 72.0 & 70.7 & 58.7 & 27.2 & 26.0 & 59.0 & 54.3 & \textbf{82.6} & 21.5 & 55.7 & 26.0 & 78.3 & 66.2 & 72.8 & 16.7 & 20.4 & 44.8 & 37.5 & 61.9 & 54.3 & 50.3 \\
CSC+FRCNN \cite{Shen2019CSC} & 64.3 & 61.4 & 47.2 & 22.5 & 29.3 & 61.9 & 50.3 & 48.6 & 17.7 & 50.5 & 22.6 & 45.7 & 43.4 & 68.8 & 34.8 & 22.2 & 48.2 & 39.9 & 59.1 & 44.6 & 44.1 \\
MIST \cite{Ren2020MIST} & \textbf{78.3} & 73.9 & 56.5 & 30.4 & 37.4 & 64.2 & 59.3 & 60.3 & 26.6 & 66.8 & 25.0 & 55.0 & 61.8 & 79.3 & 14.5 & 30.3 & 61.5 & 40.7 & 56.4 & 63.5 & 52.1 \\
SLV \cite{Chen2020SLV} & - & - & - & - & - & - & - & - & - & - & - & - & - & - & - & - & - & - & - & - & 49.2 \\
CASD \cite{Huang2020CASD} & - & - & - & - & - & - & - & - & - & - & - & - & - & - & - & - & - & - & - & - & 53.6 \\
PSLR+FRRCNN \cite{Zhang2020PSLR} & 71.3 & 67.4 & 55.6 & 31.9 & 30.1 & 59.1 & 56.6 & 55.4 & 16.2 & 56.1 & \textbf{45.0} & 70.5 & 65.4 & 70.8 & 13.5 & 28.1 & 45.6 & \textbf{54.0} & 43.4 & 58.1 & 49.7 \\
P-MIDN+MGSC+FRCNN \cite{Xu2021P-MIDN+MGSC} & 73.5 & 74.4 & 55.9 & 31.9 & 33.9 & 61.4 & 61.2 & 82.1 & 26.6 & 59.1 & 29.3 & 66.5 & 69.5 & 75.4 & 36.1 & 31.0 & 54.1 & 38.9 & 42.4 & 64.3 & 53.4 \\
D-MIL+FRCNN \cite{Gao2022D-MIL} & 69.6 & 70.2 & 53.4 & 23.7 & 33.5 & 61.3 & 58.8 & 80.1 & 22.9 & 56.4 & 27.4 & 76.2 & 64.2 & 73.2 & 6.5 & 25.7 & 47.0 & 36.2 & 47.4 & 61.7 & 49.8 \\
NDI-WSOD \cite{Wang2022NDI-WSOD} & - & - & - & - & - & - & - & - & - & - & - & - & - & - & - & - & - & - & - & - & 53.9 \\
OD-WSCL \cite{Seo2022OD-WSCL} & 73.8 & 74.7 & 61.3 & 32.9 & \textbf{40.0} & 64.6 & 59.8 & 68.1 & 26.3 & \textbf{67.5} & 23.0 & 67.1 & 62.8 & \textbf{80.6} & 17.3 & \textbf{34.1} & \textbf{63.4} & 44.4 & 66.2 & \textbf{64.9} & 54.6 \\
CPNet \cite{Li2022CPNet} & 73.0 & 74.1 & 57.6 & 33.9 & 37.0 & 60.1 & \textbf{63.0} & 36.5 & \textbf{29.8} & 66.3 & 26.0 & 52.2 & \textbf{73.4} & 74.8 & 5.7 & 31.8 & 58.8 & 44.0 & 41.4 & 64.2 & 50.2 \\
BUAA-PAL+Reg \cite{Wu2022BUAA-PAL} & 74.5 & \textbf{75.4} & 60.4 & 29.7 & 31.2 & 64.5 & 57.9 & 77.4 & 25.1 & 64.5 & 25.1 & 72.9 & 68.4 & 74.7 & 25.7 & 28.7 & 51.3 & 36.5 & 59.6 & 56.6 & 53.0 \\
CBL \cite{Yin2023CBL} & - & - & - & - & - & - & - & - & - & - & - & - & - & - & - & - & - & - & - & - & 53.5 \\
MCC-MCT+Reg \cite{Wu2024MCC-MCT} & 74.8 & \textbf{75.4} & \textbf{61.5} & 27.8 & 33.8 & 63.8 & 57.0 & 73.4 & 29.6 & 62.7 & 30.9 & 62.2 & 59.6 & 77.6 & 9.4 & 27.3 & 55.9 & 38.1 & 45.7 & 56.1 & 51.1 \\
ICBC \cite{Yin2025ICBC} & - & - & - & - & - & - & - & - & - & - & - & - & - & - & - & - & - & - & - & - & 55.5 \\
DANCE+FRCNN & 77.2 & 66.7 & 60.6 & \textbf{42.4} & 38.5 & \textbf{64.9} & 59.1 & 81.8 & 19.4 & 62.9 & 39.6 & \textbf{78.8} & 69.7 & 72.0 & \textbf{55.5} & 23.7 & 47.9 & 39.7 & \textbf{73.6} & 62.9 & \textbf{56.8}\tnote{2} \\
\bottomrule[0.15em]
\end{tabular}
}
\begin{tablenotes}
    \footnotesize
    \item[1] http://host.robots.ox.ac.uk:8080/anonymous/CS4U99.html
    \item[2] http://host.robots.ox.ac.uk:8080/anonymous/VGH65Q.html
\end{tablenotes}
\end{threeparttable}
\end{table*}

\begin{table*}[!t]
\caption{Comparison with the state-of-the-art methods on PASCAL VOC 2012 trainval set in terms of CorLoc (\%).}
\label{tab:mCorLoc-VOC12}
\centering
\newcolumntype{P}{>{\centering\arraybackslash}p{21pt}}
\newcolumntype{M}{>{\centering\arraybackslash}p{31pt}}
\resizebox{\textwidth}{!}{
\begin{tabular}{l | P P P P P P P P P P P P P P P P P P P P | M}
\toprule[0.15em]
Method & aero & bike & bird & boat & bottle & bus & car & cat & chair & cow & table & dog & horse & mbike & person & plant & sheep & sofa & train & tv & mCorLoc \\
\midrule[0.05em]
ContextLocNet \cite{Kantorov2016ContextLocNet} & 78.3 & 70.8 & 52.5 & 34.7 & 36.6 & 80.0 & 58.7 & 38.6 & 27.7 & 71.2 & 32.3 & 48.7 & 76.2 & 77.4 & 16.0 & 48.4 & 69.9 & 47.5 & 66.9 & 62.9 & 54.8 \\
OICR \cite{Tang2017OICR} & 86.2 & 84.2 & 68.7 & 55.4 & 46.5 & 82.8 & 74.9 & 32.2 & 46.7 & 82.8 & 42.9 & 41.0 & 68.1 & 89.6 & 9.2 & 53.9 & 81.0 & 52.9 & 59.5 & 83.2 & 62.1 \\
WCCN \cite{Diba2017WCCN} & - & - & - & - & - & - & - & - & - & - & - & - & - & - & - & - & - & - & - & - & - \\
TS$^{2}$C \cite{Wei2018TS2C} & 79.1 & 83.9 & 64.6 & 50.6 & 37.8 & 87.4 & 74.0 & 74.1 & 40.4 & 80.6 & 42.6 & 53.6 & 66.5 & 88.8 & 18.8 & 54.9 & 80.4 & 60.4 & 70.7 & 79.3 & 64.4 \\
PCL \cite{Tang2018PCL} & 77.2 & 83.0 & 62.1 & 55.0 & 49.3 & 83.0 & 75.8 & 37.7 & 43.2 & 81.6 & 46.8 & 42.9 & 73.3 & 90.3 & 21.4 & 56.7 & \textbf{84.4} & 55.0 & 62.9 & 82.5 & 63.2 \\
WS-JDS \cite{Shen2019WS-JDS} & - & - & - & - & - & - & - & - & - & - & - & - & - & - & - & - & - & - & - & - & 63.5 \\
OAIL \cite{Kosugi2019OAIL} & 86.5 & 82.1 & 67.2 & 58.7 & 48.9 & 80.5 & 75.6 & 62.3 & 46.0 & 81.9 & 40.0 & 64.2 & 82.4 & 88.2 & 44.2 & 53.5 & 78.1 & 54.7 & 56.7 & 82.9 & 66.7 \\
SDCN \cite{Li2019SDCN} & - & - & - & - & - & - & - & - & - & - & - & - & - & - & - & - & - & - & - & - & 67.9 \\
C-MIDN \cite{Gao2019C-MIDN} & - & - & - & - & - & - & - & - & - & - & - & - & - & - & - & - & - & - & - & - & 71.2 \\
CSC \cite{Shen2019CSC} & - & - & - & - & - & - & - & - & - & - & - & - & - & - & - & - & - & - & - & - & 61.4 \\
PSLR \cite{Zhang2020PSLR} & - & - & - & - & - & - & - & - & - & - & - & - & - & - & - & - & - & - & - & - & 68.7 \\
P-MIDN+MGSC \cite{Xu2021P-MIDN+MGSC} & - & - & - & - & - & - & - & - & - & - & - & - & - & - & - & - & - & - & - & - & 73.3 \\
IM-CFB \cite{Yin2021IM-CFB} & - & - & - & - & - & - & - & - & - & - & - & - & - & - & - & - & - & - & - & - & 69.6 \\
D-MIL \cite{Gao2022D-MIL} & 84.5 & 83.0 & 71.5 & 51.9 & 52.1 & 89.5 & 76.7 & 83.9 & 51.5 & 87.7 & 52.3 & 82.7 & 84.5 & 91.2 & 19.4 & 53.0 & \textbf{84.4} & 50.8 & 67.8 & 83.0 & 70.1 \\
BUAA-PAL \cite{Wu2022BUAA-PAL} & 90.2 & \textbf{88.0} & 74.0 & 48.5 & 56.7 & 85.7 & 86.7 & 76.0 & 52.0 & 86.6 & 62.6 & 66.8 & 81.2 & \textbf{94.6} & 28.2 & \textbf{66.0} & 82.7 & 65.3 & 76.8 & 78.8 & 72.4 \\
MCC-MCT \cite{Wu2024MCC-MCT} & 90.7 & 85.5 & 70.8 & 51.9 & 57.8 & 87.4 & \textbf{88.1} & 51.4 & 51.1 & \textbf{89.0} & 61.1 & 49.5 & 77.6 & 92.3 & 19.0 & 63.3 & 83.4 & 65.1 & 76.1 & 74.8 & 69.3 \\
DANCE & \textbf{92.5} & 74.3 & \textbf{80.4} & \textbf{75.2} & \textbf{64.9} & \textbf{94.5} & 81.5 & \textbf{93.4} & \textbf{53.1} & 87.5 & \textbf{69.5} & \textbf{92.1} & \textbf{89.6} & 89.7 & \textbf{80.7} & 57.7 & 80.9 & \textbf{71.2} & \textbf{91.2} & \textbf{90.6} & \textbf{80.5} \\
\midrule[0.05em]
\midrule[0.05em]
OICR-Ens.+FRCNN \cite{Tang2017OICR} & 89.3 & 86.3 & 75.2 & 57.9 & 53.5 & 84.0 & 79.5 & 35.2 & 47.2 & 87.4 & 43.4 & 43.8 & 77.0 & 91.0 & 10.4 & 60.7 & 86.8 & 55.7 & 62.0 & 84.7 & 65.6 \\
W2F \cite{Zhang2018W2F} & 88.8 & 85.8 & 64.9 & 56.0 & 54.3 & 88.1 & 79.1 & 67.8 & 46.5 & 86.1 & 26.7 & 77.7 & 87.2 & 89.7 & 28.5 & 56.9 & 85.6 & 63.7 & 71.3 & 83.0 & 69.4 \\
PCL-Ens.+FRCNN \cite{Tang2018PCL} & 86.7 & 86.7 & 74.8 & 56.8 & 53.8 & 84.2 & 80.1 & 42.0 & 36.4 & 86.7 & 46.5 & 54.1 & 87.0 & 92.7 & 24.6 & 62.0 & 86.2 & 63.2 & 70.9 & 84.2 & 68.0 \\
WS-JDS+FRCNN \cite{Shen2019WS-JDS} & - & - & - & - & - & - & - & - & - & - & - & - & - & - & - & - & - & - & - & - & 69.5 \\
WSOD$^{2}$ \cite{Zeng2019WSOD2} & - & - & - & - & - & - & - & - & - & - & - & - & - & - & - & - & - & - & - & - & 71.9 \\
TPEE \cite{Yang2019TPEE} & 80.2 & 83.0 & 73.1 & 51.6 & 48.3 & 79.8 & 76.6 & 70.3 & 44.1 & 87.7 & 50.9 & 70.3 & 84.7 & 92.4 & 28.5 & 59.3 & 83.4 & 64.6 & 63.8 & 81.2 & 68.7 \\
SDCN+FRCNN \cite{Li2019SDCN} & - & - & - & - & - & - & - & - & - & - & - & - & - & - & - & - & - & - & - & - & 69.5 \\
C-MIDN+FRCNN \cite{Gao2019C-MIDN} & - & - & - & - & - & - & - & - & - & - & - & - & - & - & - & - & - & - & - & - & 73.3 \\
CSC+FRCNN \cite{Shen2019CSC} & - & - & - & - & - & - & - & - & - & - & - & - & - & - & - & - & - & - & - & - & 67.0 \\
MIST \cite{Ren2020MIST} & 91.7 & 85.6 & 71.7 & 56.6 & 55.6 & 88.6 & 77.3 & 63.4 & 53.6 & \textbf{90.0} & 51.6 & 62.6 & 79.3 & 94.2 & 32.7 & 58.8 & 90.5 & 57.7 & 70.9 & 85.7 & 70.9 \\
SLV \cite{Chen2020SLV} & - & - & - & - & - & - & - & - & - & - & - & - & - & - & - & - & - & - & - & - & 69.2 \\
CASD \cite{Huang2020CASD} & - & - & - & - & - & - & - & - & - & - & - & - & - & - & - & - & - & - & - & - & 72.3 \\
PSLR+FRRCNN \cite{Zhang2020PSLR} & - & - & - & - & - & - & - & - & - & - & - & - & - & - & - & - & - & - & - & - & 74.5 \\
P-MIDN+MGSC+FRCNN \cite{Xu2021P-MIDN+MGSC} & - & - & - & - & - & - & - & - & - & - & - & - & - & - & - & - & - & - & - & - & 76.0 \\
D-MIL+FRCNN \cite{Gao2022D-MIL} & 87.4 & 83.9 & 73.2 & 55.6 & 57.4 & 90.5 & 78.8 & 84.7 & 54.0 & 87.7 & 54.8 & 84.1 & 87.2 & 92.5 & 20.7 & 55.6 & 86.5 & 49.2 & 70.7 & 83.9 & 71.9 \\
NDI-WSOD \cite{Wang2022NDI-WSOD} & - & - & - & - & - & - & - & - & - & - & - & - & - & - & - & - & - & - & - & - & 72.2 \\
OD-WSCL \cite{Seo2022OD-WSCL} & 88.2 & 88.3 & 75.0 & 59.7 & 58.9 & 89.3 & 73.2 & 57.8 & 53.4 & 88.0 & 48.7 & 67.5 & 78.3 & 94.0 & 34.8 & 61.6 & \textbf{91.7} & 59.4 & 70.9 & 84.4 & 71.2 \\
CPNet \cite{Li2022CPNet} & - & - & - & - & - & - & - & - & - & - & - & - & - & - & - & - & - & - & - & - & - \\
BUAA-PAL+Reg \cite{Wu2022BUAA-PAL} & 90.2 & \textbf{89.9} & 75.8 & 48.5 & 56.2 & 89.6 & 84.5 & 82.9 & 50.9 & 86.5 & 59.6 & 76.8 & 82.6 & \textbf{95.2} & 35.0 & \textbf{64.3} & 80.9 & 66.5 & 83.0 & 81.9 & 74.0 \\
CBL \cite{Yin2023CBL} & - & - & - & - & - & - & - & - & - & - & - & - & - & - & - & - & - & - & - & - & 72.6 \\
MCC-MCT+Reg \cite{Wu2024MCC-MCT} & 90.4 & 89.1 & 75.6 & 47.2 & 58.4 & 90.8 & \textbf{86.7} & 78.7 & \textbf{56.8} & 86.1 & 65.0 & 70.5 & 74.2 & 94.7 & 21.3 & 63.8 & 85.9 & 69.6 & 71.4 & 70.6 & 72.3 \\
ICBC \cite{Yin2025ICBC} & - & - & - & - & - & - & - & - & - & - & - & - & - & - & - & - & - & - & - & - & 73.4 \\
DANCE+FRCNN & \textbf{93.3} & 76.6 & \textbf{83.4} & \textbf{76.0} & \textbf{65.6} & \textbf{93.6} & 83.8 & \textbf{94.2} & 52.9 & 88.4 & \textbf{70.3} & \textbf{92.7} & \textbf{91.1} & 89.5 & \textbf{82.9} & 56.0 & 83.7 & \textbf{71.2} & \textbf{92.6} & \textbf{90.1} & \textbf{81.4} \\
\bottomrule[0.15em]
\end{tabular}
}
\end{table*}

We also provide detailed per-class results on Pascal VOC 2012. As shown in Tables \ref{tab:mAP-VOC12} and \ref{tab:mCorLoc-VOC12}, our method outperforms current SOTA results in most categories. Consistent with our findings on VOC07, our approach achieves particularly significant performance in the ``person'' category.

\section{Computational Cost} \label{app:E}

In this section, we present a comparison between our method and other approaches in terms of computational cost. From a theoretical perspective, since almost all WSOD models follow the same fundamental architecture --- consisting of a base MIDN network and multiple cascaded refinement branches --- the primary difference lies only in the strategies used to select pseudo GT boxes during training. During inference, all methods utilize only the backbone, RoI neck, and the final cascaded refinement branch. Therefore, under the same experimental settings, all methods are expected to have identical computational costs.

To validate this claim, we evaluate three metrics: the number of parameters, GFLOPs, and FPS on the Pascal VOC 2007 \textit{test} set. Specifically, \textbf{GFLOPs} are computed by resizing the shorter edge of each image to 800 pixels while proportionally scaling the longer edge; the total number of floating-point operations for the forward pass over the entire test set is then averaged by the number of images. \textbf{Parameters} refers to the total number of model parameters involved during inference. \textbf{FPS} is defined as the total number of images divided by the total inference time on the test set under the Test-Time Augmentation (TTA) protocol described in Section \ref{sec:4-2} (The inference time per image accounts for the total time required to process 16 augmented views --- derived from 8 different scales combined with horizontal flipping --- and aggregate their predictions).

We evaluate these metrics on OICR \cite{Tang2017OICR}, SLV \cite{Chen2020SLV}, and DANCE. The experimental results confirm our hypothesis: all methods achieve identical computational cost, with an average of 576.0 GFLOPs per image, 134.7M Parameters, and an inference speed of 0.66 FPS. Therefore, our model does not introduce any additional computational overhead compared to other methods.

\section{Failure Case Analysis}

\begin{figure*}[!t]
\centering
\includegraphics[width=\textwidth]{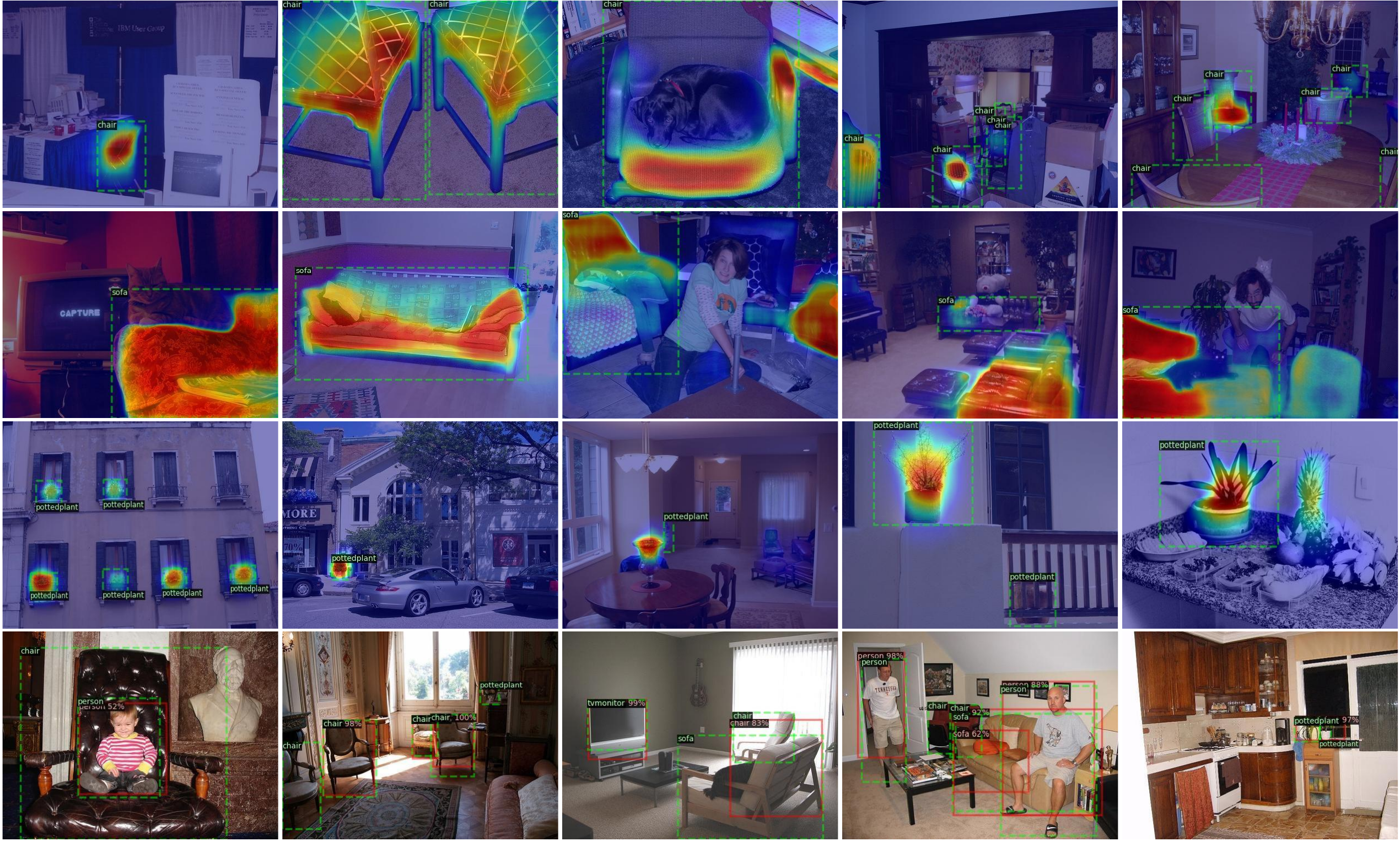}
\caption{Visualization of category-specific heatmaps and detection results on Pascal VOC. The first three rows are heatmap visualizations of images from the \textit{trainval} set, corresponding to the chair, sofa, and potted plant categories from top to bottom respectively. The green dashed boxes represent the GT boxes for the corresponding categories. Among them, the two leftmost columns display high-quality heatmaps, while the three rightmost columns show less satisfactory heatmaps. The last row illustrates the failure detection results for images in the \textit{test} set, where the red solid boxes indicate the detection bounding boxes along with their predicted categories and confidence scores, and the green dashed boxes represent all GT boxes in the image.}
\label{fig:failure-cases}
\end{figure*}

Although our method achieves outstanding average detection metrics across the entire dataset and demonstrates impressive performance on certain specific categories (e.g., \textbf{person}), there are still some categories where it underperforms. As observed in Tables \ref{tab:mAP-VOC07}, \ref{tab:mCorLoc-VOC07}, \ref{tab:mAP-VOC12}, and \ref{tab:mCorLoc-VOC12}, our method yields lower AP and CorLoc on the \textbf{chair}, \textbf{sofa}, and \textbf{potted plant} categories. Therefore, we conduct a qualitative analysis on these categories to investigate the reasons for these failures. We visualize the heatmaps from the \textit{trainval} set and the detection bounding boxes from the \textit{test} set for these three categories. As shown in Figure \ref{fig:failure-cases}, we display both successful and less satisfactory heatmaps. We observe that the quality of the generated heatmaps is sometimes less than ideal, occasionally exhibiting false positives and false negatives, which in turn constrains the performance of our model. We believe that with the continuous efforts of the research community, as heatmap generation methods continue to advance, the performance of DANCE will also reach new heights.

\printbibliography

\begin{figure*}[!p]
\centering
\vfill
\includegraphics[width=\textwidth, height=\textheight, keepaspectratio]{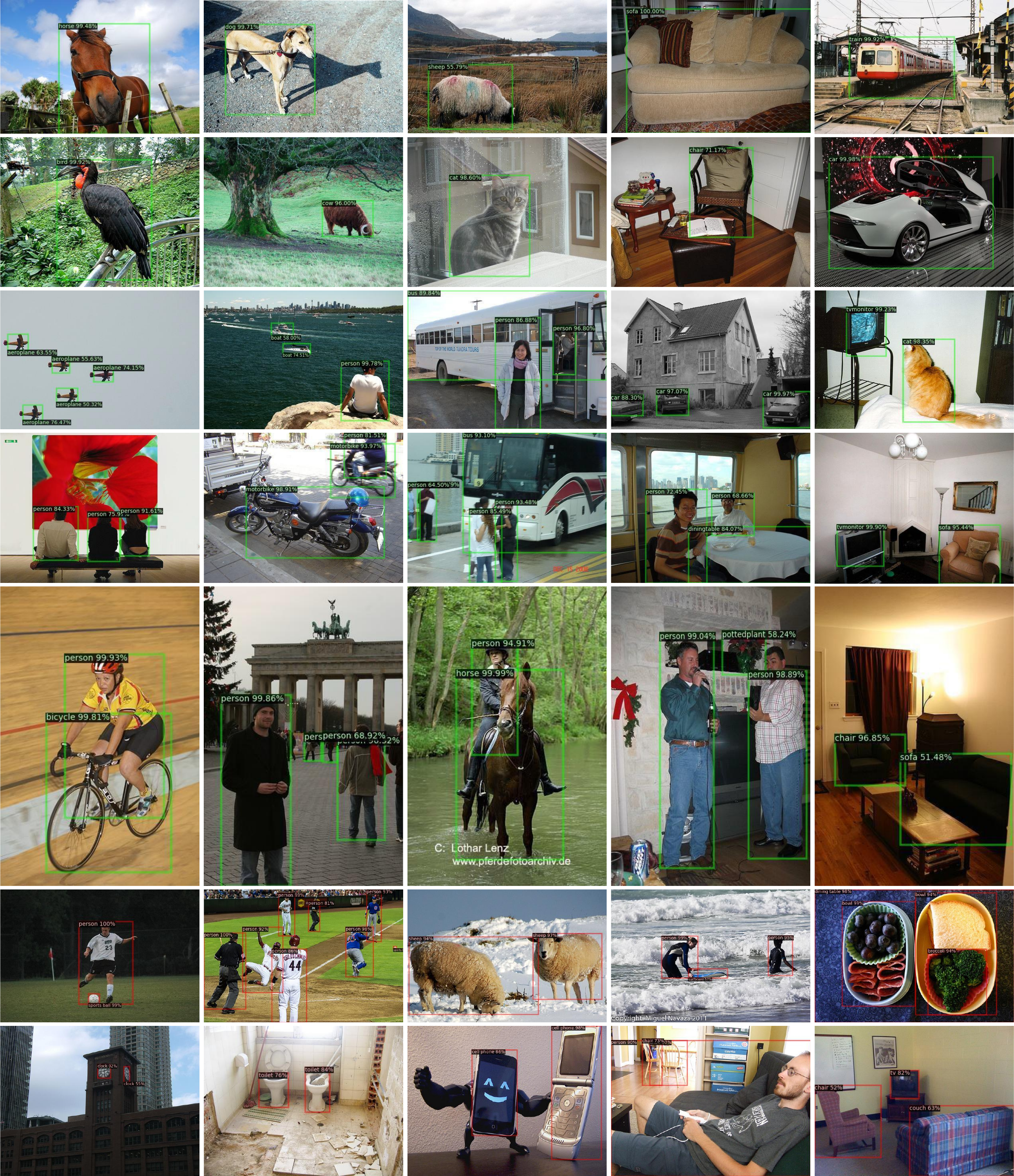}
\caption{More visualization results of DANCE. The first five rows show detection results on the Pascal VOC \textit{test} set, while the last two rows show detection results on the MS COCO \textit{val} set.}
\label{fig:more-vis}
\end{figure*}




\end{document}